\documentclass{article}
\usepackage{iclr2027_conference,times}

\usepackage[T1]{fontenc}
\usepackage[utf8]{inputenc}
\usepackage{amsmath,amssymb}
\usepackage{graphicx}
\usepackage{wrapfig}
\usepackage{booktabs}
\usepackage{multirow}
\usepackage{float}
\usepackage{xcolor}
\usepackage{hyperref}
\usepackage{cleveref}
\usepackage{url}
\usepackage{capt-of}

\newcommand{\method}{TailSteer}

\newif\ifbibliographyready
\bibliographyreadytrue

\graphicspath{{figures/}}
\hypersetup{pdftitle={TailSteer: Trajectory-Tail Steering for Drift-Resistant Multi-turn Image Editing},pdfauthor={}}

\title{TailSteer: Trajectory-Tail Steering for Drift-Resistant Multi-turn Image Editing}

\author{%
\textbf{Ruikang Li$^{1}$ \quad Zixuan Chen$^{1}$ \quad Jiarui Wu$^{1}$ \quad Mingde Yao$^{1}$ \quad Yiyuan Zhang$^{1}$}\\
\textbf{Yujin Wang$^{3}$ \quad Hanming Deng$^{2}$ \quad Lewei Lu$^{2}$ \quad Tianfan Xue$^{1}$}\\[0.5em]
{\normalfont $^{1}$MMLab, CUHK}\\
{\normalfont $^{2}$SenseTime Research}\\
{\normalfont $^{3}$Shanghai Artificial Intelligence Laboratory}%
}

\iclrfinalcopy

\begin{document}
\maketitle
\vspace{-8mm}
\begin{figure}[h]
  \centering
  \includegraphics[width=\textwidth]{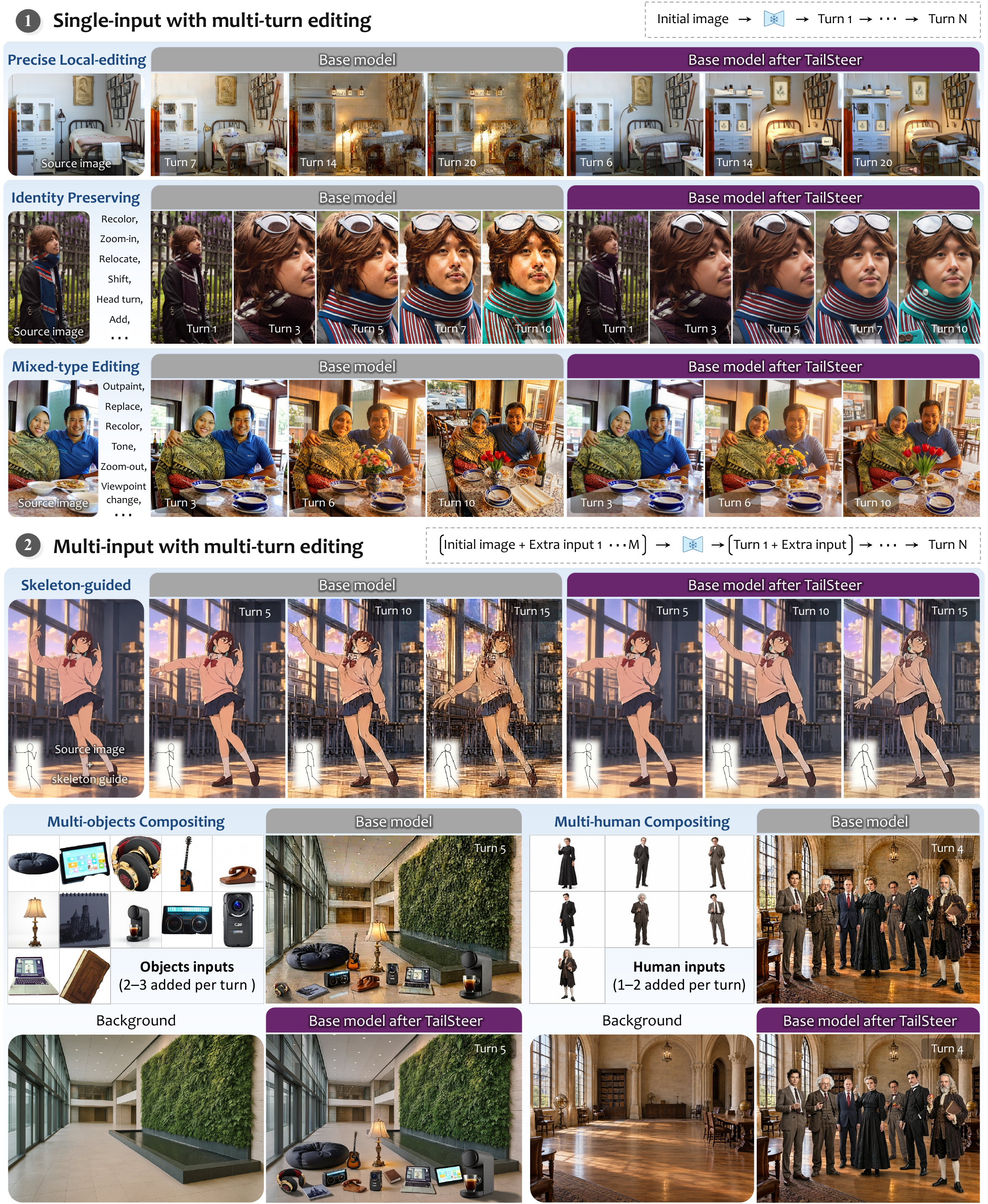}
  \caption{\textbf{Single- and multi-input multi-turn editing.}
    TailSteer substantially reduces cumulative drift over editing
  turns while retaining the base model's (FLUX.2 [klein] Base 9B) editing capability.}
  \label{fig:teaser}
\end{figure}

\begin{abstract}
  Strong single-turn editing performance does not guarantee reliable multi-turn interaction: as edited outputs are repeatedly reused, small unintended changes can accumulate into severe visual drift.
This long-term drift is often mitigated through training on multi-turn data or cross-turn rollouts, potentially altering pretrained priors, or through specialized inference-time mechanisms that are difficult to scale to foundation models.
We introduce \textbf{TailSteer}, a final-stage training framework that improves multi-turn robustness while retaining the generative and editing capabilities acquired through earlier training. 
TailSteer follows a simple principle: keep early denoising dynamics close to those of the base model and concentrate preservation corrections near the trajectory tail, where fine-grained deviations can be suppressed. These corrections are learned from the editor's own on-policy denoising rollouts, using unchanged source content as a preservation reference without requiring curated ground-truth editing pairs.
Experiments across the FLUX, SenseNova, and Qwen families show that TailSteer substantially reduces cumulative drift over 15–20 editing turns while largely preserving base-model performance on extensive text-to-image and single-image editing benchmarks. These results demonstrate that long-horizon editing robustness can be unlocked in foundation models after TailSteer, rather than by relearning their existing capabilities. \textbf{Project page:} \url{https://lyricccco.github.io/tailsteer/}
\end{abstract}

\section{Introduction}
\label{sec_1_introduction}

Large-scale generative models have substantially advanced instruction-based image editing, achieving increasingly strong instruction following, visual fidelity, and content preservation in single-turn editing~\citep{openai2025gptimage,openai2025gptimage15,labs2025flux1kontextflowmatching,flux22025,google2025gemini25flashimage}. At the same time, image generation is increasingly embedded in interactive and conversational creative workflows. Recent systems, such as Nano Banana 2, ChatGPT Images, Reve, or Muse Image, allow users to repeatedly refine an image through natural-language feedback, visual references, and agentic interactions~\citep{google2026nanobanana2,openai2026images25,reve2026reve21,meta2026museimage}. This interaction paradigm better matches how visual content is actually created: rather than specifying a complete target at once, users progressively develop an idea through a sequence of edits.  

However, applying one-shot editors to iterative editing introduces a critical challenge: small errors in each edit can accumulate across successive turns. In a multi-turn workflow, each edited output becomes the next input, causing unintended changes to be repeatedly inherited by subsequent generations. As illustrated in Fig.~\ref{fig:teaser}, repeated editing can wash out fine textures, alter non-target object or facial details, distort background structures, and eventually produce visible artifacts. Thus, strong single-turn editing performance does not necessarily translate into robust long-horizon editing.

Existing approaches to improving multi-turn image editing can be broadly categorized into three families. 
i) \emph{Adding auxiliary modules or constraints} to a pretrained model, by introducing spatial guidance (e.g., masks), specially trained LoRA adapters, or frequency corrections \citep{couairon2023diffedit,dx8152flux2consistency,liao2026freqedit,wang2026ditdrift}. These mechanisms improve preservation by sacrificing or constraining part of the large-scale generative prior learned by the pretrained model, and are therefore often limited to a relatively narrow range of localized edit types.
ii) \emph{Refining the representation space} to better support image editing. As reconstruction fidelity and generative capability are widely recognized as a latent-generative dilemma \citep{yao2025reconstruction,dieleman2025latents}, Black Forest Labs specifically treats the VAE representation to obtain a more suitable editing space \citep{flux2latent2025}, while other approaches bypass the VAE and perform pixel-diffusion \citep{yu2026pixeldit2}. Although these designs effectively alleviate errors within a single editing turn, they still suffer from noticeable degradation across repeated edits.
iii) \emph{Maintaining explicit memory across edits} within creative workflows, by caching latent states or previous visual context \citep{yu2025contextmemory,dalal2025oneminute,kim2025layerwisememory}, or by leveraging reasoning models for agentic planning and iterative refinement in commercial generative systems \citep{meta2026museimage,openai2026images25}. While these mechanisms help retain historical information across turns, they do not directly improve the underlying model itself. Consequently, long-horizon consistency remains bottlenecked by the base model's own robustness under repeated editing, often with additional inference overhead.

\newpage
\vspace*{-6mm}
Although these approaches can improve multi-turn consistency, they often require auxiliary modules, alternative representations, or dedicated models. We instead directly improve the foundation model, so that multi-turn robustness is built into the model. This will not introduce additional inference components and also remain complementary to existing agentic workflow-level solutions.

\begin{wrapfigure}{r}{0.5\textwidth}
  \centering
  \vspace{-5mm}
  \includegraphics[width=\linewidth]{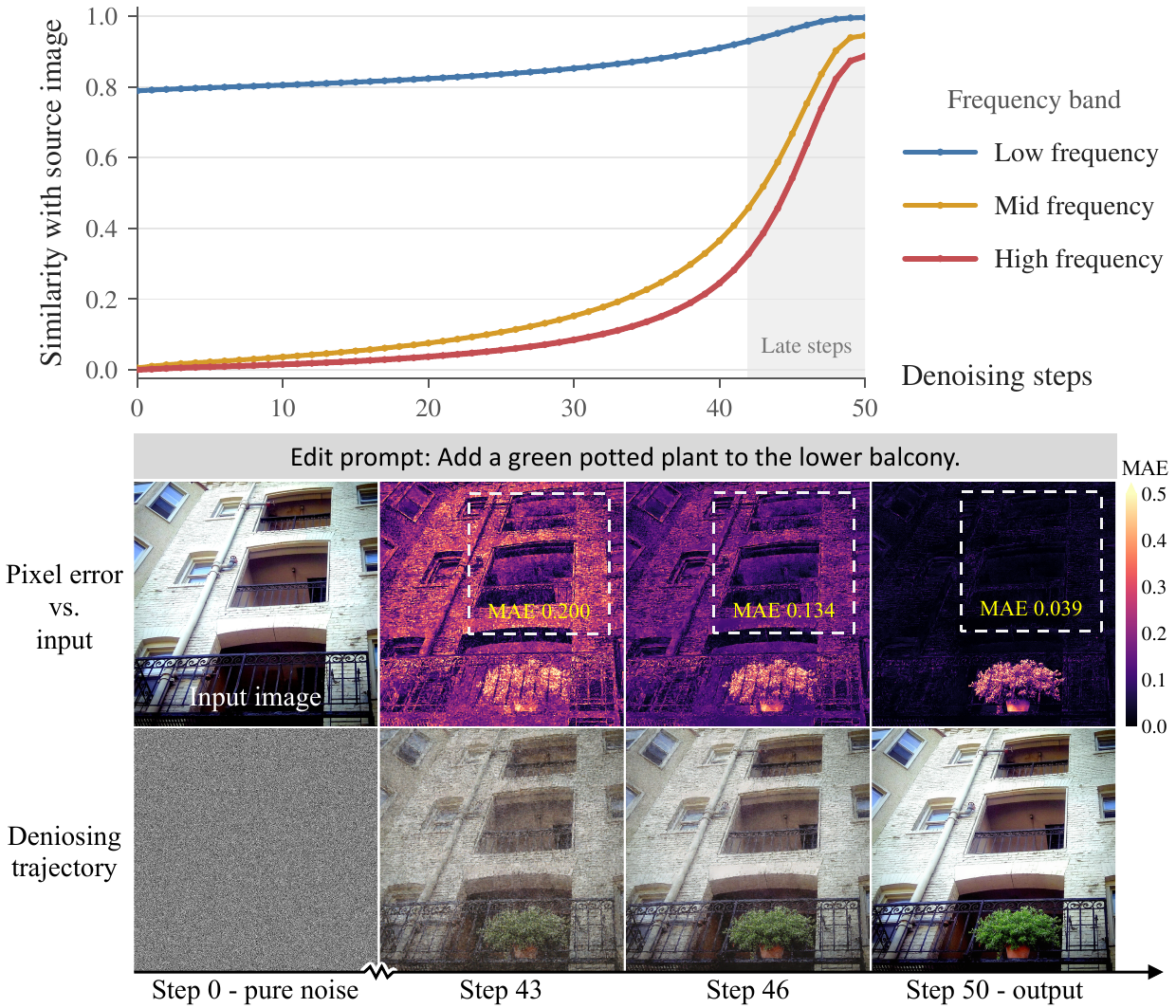}
  \caption{\textbf{Trajectory tail remains correctable.}
    Low-frequency content is recovered early, whereas
    higher-frequency details are primarily recovered near the tail (top),
    while preservation error in unchanged regions continues to decrease but remains non-zero at the endpoint (bottom).
  }
  \vspace{-5mm}
  \label{fig_2_frequency_analysis}
\end{wrapfigure}

To improve the existing model, we first analyze why drifting happens in image editing, by inspecting the denoising trajectory.
Prior studies show that diffusion sampling follows a coarse-to-fine progression: low-frequency structure is established early, whereas high-frequency details are mainly recovered near the trajectory tail ~\citep{qian2024boosting}. This is particularly relevant to multi-turn editing, where repeated edits progressively degrade high-frequency information~\citep{liao2026freqedit}. Moreover, because each inference state depends on preceding model predictions, prediction errors can propagate along the sampling trajectory~\citep{ning2023inputperturbation,ning2024elucidating,li2024timeshift}. Fig.~\ref{fig_2_frequency_analysis} reveals that low-frequency similarity is already high before the final steps, whereas higher-frequency similarity rises sharply near the tail. Meanwhile, preservation error in unchanged regions continues to decrease but remains nonzero at the endpoint. These observations motivate refining the trajectory tail, where fine-grained details are finalized and residual preservation errors remain correctable.

Based on this insight, we introduce \textbf{TailSteer}, a final-stage training framework that improves multi-turn robustness by steering only the trajectory tail. Rather than broadly reshaping the foundation model's learned prior, \textit{TailSteer keeps early denoising dynamics close to those of the base model while progressively steering later denoising states toward a preservation-aware endpoint}. To address the trajectory-level train-test gap, TailSteer learns from states reached through its own on-policy denoising rollouts, using unchanged source content as a preservation reference without curating ground-truth editing pairs. This targeted steering improves long-horizon robustness while retaining the foundation model's learned priors.

TailSteer's tail-focused design offers three practical advantages. 
First, TailSteer improves long-horizon editing robustness while largely preserving the generative and editing capabilities acquired through earlier training. Extensive evaluations on text-to-image generation and single-turn editing benchmarks show performance close to the foundation models, alongside substantial reductions in cumulative drift.
Second, TailSteer generalizes across model families and input settings by operating directly on denoising trajectories, without relying on a particular latent representation, attention mechanism, or model architecture. We validate this generality across FLUX~\citep{flux2klein2026}, SenseNova~\citep{diao2026sensenova}, and Qwen~\citep{qwen2026image21} backbones, including multi-image compositing and pose-guided editing.
Finally, TailSteer constructs preservation supervision from source images and the base model's own rollouts, without requiring curated ground-truth editing pairs or cross-turn training data. Our scaling experiments further show that larger and more diverse rollout datasets can improve drift resistance while retaining the base model's editing capabilities.

Our contributions are three-fold:
(1) We identify the denoising trajectory tail as a key intervention point for multi-turn editing, where fine-grained details are finalized while residual preservation errors remain correctable without broadly altering earlier dynamics.
(2) We introduce \textbf{TailSteer}, a final-stage training framework that learns from the model's own on-policy rollouts, unlocking multi-turn robustness through trajectory-tail steering while largely preserving the base model's learned priors.
(3) We introduce \textbf{Add \& Preserve} to separately evaluate cumulative drift and edit completion, alongside extensive evaluations on text-to-image generation, single-turn editing, and general multi-turn benchmarks. Results across FLUX, SenseNova, and Qwen demonstrate substantially enhanced model-native long-horizon preservation while largely retaining original capabilities, with preservation performance rivaling or even surpassing GPT-Image-2.5 Sunburst (via direct API calls).
\section{Related Work}
\label{sec:related-work}
\vspace{-1mm}
Multi-turn degradation has also been observed in LLMs~\citep{laban2026llms}, despite their growing success in agentic workflows. For image generation, however, repeatedly reusing generated outputs can compound visual errors, making long-horizon editing robustness a critical bottleneck for agentic creation. Existing methods address this limitation from several directions.

\vspace{-3mm}
\paragraph{Representation fidelity.}
Prior work has explored the trade-off between reconstruction fidelity and generative learnability in latent representations, motivating improved representation designs~\citep{dieleman2025latents,yao2025reconstruction,flux2latent2025}. For iterative editing, other approaches mitigate accumulated reconstruction errors by reusing latent states across turns or improving stability under repeated encoding and decoding~\citep{joseph2024iterative,almog2025reedvae}. Alternatively, pixel-space generative models bypass the VAE bottleneck altogether~\citep{li2026back,yu2026pixeldit2,diao2026sensenovau1}. These approaches address representation-level limitations, whereas TailSteer improves long-horizon preservation by refining existing models' denoising dynamics.

\vspace{-3mm}
\paragraph{Inference-time guidance.}
Prior work improves preservation through mask-guided constraints that localize edits~\citep{avrahami2022blended,couairon2023diffedit,yin2025consistedit}, and frequency-based corrections that mitigate content degradation across turns~\citep{liao2026freqedit,wang2026ditdrift}. Beyond these constraints, training-free methods guide inversion and sampling to balance editing and preservation~\citep{zhou2025multiturn,jiang2026flowdc,kim2025reflex}. While effective in specific settings, these specialized inference procedures may not readily generalize across foundation models and do not directly improve the underlying model's robustness.

\vspace{-3mm}
\paragraph{Feature reuse and visual memory.}
Prior work improves editing consistency through feature reuse and cross-turn memory~\citep{zhu2025kvedit,kim2025layerwisememory}. Long-horizon video generation similarly leverages context retrieval and test-time memory~\citep{yu2025contextmemory,dalal2025oneminute}. Yet long-horizon consistency remains bottlenecked by the underlying generator's per-step robustness.

\vspace{-3mm}
\paragraph{Multi-turn-specific training.}
Prior work trains multi-turn editors using video-derived sequences or cross-turn reinforcement learning~\citep{qu2026vincie,huang2026mteditflow}, while others train consistency LoRAs~\citep{dx8152flux2consistency}. In contrast, TailSteer unlocks base model's multi-turn potential through trajectory-tail steering, without relearning editing behavior from cross-turn supervision.
\section{Method}
\label{sec_3_method}
\vspace{-1mm}
TailSteer follows a simple principle: preserve the base model's early denoising dynamics while steering only its tail toward better preservation. We present the framework through three progressively connected parts. First, we construct preservation-aware supervision directly from the frozen editor's own rollouts, forming a ``data-free'' training paradigm without manually curated ground-truth editing pairs. Second, we study a one-step terminal intervention with supervision applied only to the final sampling step, demonstrating that endpoint correction alone can reduce cumulative drift.  Finally, we extend this single-step intervention to \textbf{TailSteer}, distributing preservation-oriented steering over multiple steps in the trajectory tail while retaining the original dynamics of the earlier steps.

\begin{figure}[t]
  \centering
  \includegraphics[width=\linewidth]{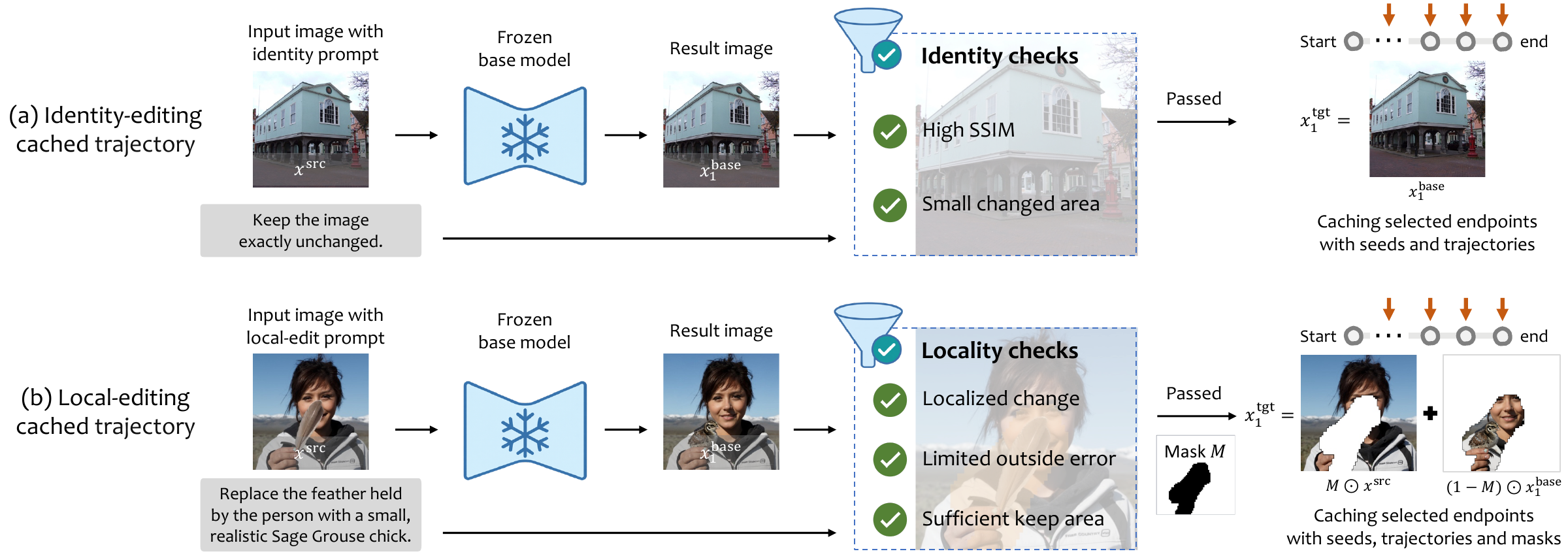}
  \caption{\textbf{Data curation of preservation-aware trajectory endpoints.}
    We construct supervision from rollouts of the frozen base editor using two
    complementary edit types: (a) identity editing provides full-image preservation
    trajectories, while (b) local editing provides partial-preservation trajectories
    with spatially localized changes. Only samples that pass the corresponding
    sanity checks are retained and cached as training trajectories for subsequent
    TailSteer optimization.}
  \label{fig_3_data_curation}
\end{figure}

\subsection{Preservation-Aware Trajectory Endpoint}
\label{sec_3_1_terminal_intervention}

To steer the trajectory tail, the first question is what the trajectory should be steered toward. This target should remain aligned with the model's own editing trajectory while correcting only the unintended deviations accumulated during sampling. However, conventional manually curated source--target editing pairs are poorly suited to this purpose. Given the same source image and instruction, a generative editor may produce different yet valid outputs under different sampling seeds. For example, an instruction to add an apple may yield apples with different sizes, colors, or surface patterns. Our goal is not to force the model toward one particular edit realization, but to correct the rollout-induced errors while preserving the edit already chosen by the model. 

Remarkably, the preservation target is already available in the source image: the content not requested to change. For example, when replacing an object in a local region, the surrounding subject and background should remain unchanged. Although modern editors exhibit strong instruction following to localize the requested edit, their denoising rollouts still introduce small deviations into such non-target content. These model-induced deviations therefore provide exactly the preservation errors that TailSteer aims to correct.

To obtain reliable supervision, we construct two complementary trajectory types in which the resulting edit is expected to preserve part of the source image, \emph{identity editing} and \emph{local editing}, as illustrated in Fig.~\ref{fig_3_data_curation}.  Identity editing requests no visual change and therefore provides full-image preservation supervision, whereas local editing confines the requested modification to a localized region, leaving the remaining content available for partial preservation supervision. We generate rollouts with the frozen base model and filter them using the corresponding identity or locality checks, retaining only trajectories with reliable preservation regions. For accepted edits, these regions are further converted into preservation masks, as described below.

For local editing, we further derive a binary preservation mask from the difference between the source image and the model's generated result. The mask is not specified in advance as we do not require the generated edit to match a ground-truth target; instead, we only require the observed change in the masked regions to be sufficiently localized. Let $x^{\mathrm{src}}$ denote the source state, $x_1^{\mathrm{base}}$ the endpoint of the frozen base editor's rollout, and $M$ the resulting preservation mask, where $M=1$ denotes regions that should remain unchanged. We construct preservation-aware target endpoint as
\begin{equation}
    x_1^{\mathrm{tgt}}
    =
    M \odot x^{\mathrm{src}}
    +
    (1-M) \odot x_1^{\mathrm{base}}.
\end{equation}
For identity editing, we simply set $M=1$. Thus, $x_1^{\mathrm{tgt}}$ restores the source content in preservation regions while retaining the edit realization produced by the base editor elsewhere.

For each accepted rollout, we cache the source state, selected intermediate states, the terminal state, the preservation mask, and the sampling seed for subsequent TailSteer training. Importantly, the two restricted prompt families are used only to obtain reliable preservation supervision; TailSteer itself learns a trajectory-level correction rather than an edit-specific behavior. We later show that this supervision generalizes beyond these training edits, improving multi-turn robustness for subject identity and multi-image-conditioned editing without sacrificing broader editing capabilities.

\subsection{One-step Terminal Intervention}
\label{sec_3_2_terminal_intervention}

\begin{figure}[t]
    \centering
    \includegraphics[width=\linewidth]{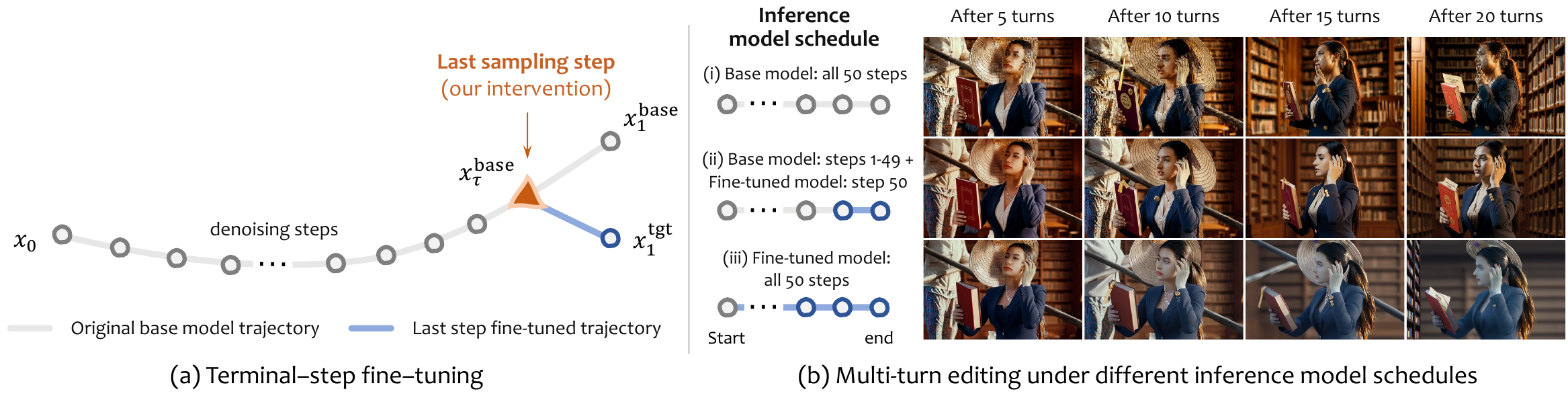}
    \vspace{-6mm}
    \caption{\textbf{One-step terminal intervention: benefits and limitations.}
    (a) We fine-tune FLUX.2-klein-9B with supervision only at the final sampling
    step, steering from the cached pre-terminal state
    $x_{\tau}^{\mathrm{base}}$ toward the preservation-aware target
    $x_1^{\mathrm{tgt}}$.
    (b) The base model accumulates visible identity drift over repeated edits (top row). Applying the fine-tuned model only at the final step markedly improves long-horizon preservation (middle row), whereas using it throughout sampling causes desaturation and detail loss (bottom row). Therefore, it is necessary to preserve the earlier dynamics.}
    \vspace{-4mm}
    \label{fig_4_terminal_intervention}
\end{figure}

We first test whether the curated preservation-aware endpoint can improve multi-turn stability through the smallest possible intervention: correcting only the final sampling step. To test this native solution, we initialize from the original base model and fine-tune it exclusively on the final sampling step: starting from the cached pre-terminal state, the model is supervised toward the preservation-aware endpoint, while the preceding timesteps receive no training objective (Fig.~\ref{fig_4_terminal_intervention}~(a)).

Let $\tau<1$ denote the pre-terminal sampling time, with time increasing from noise at $0$ to output at $1$. Given the cached state $x_\tau^{\mathrm{base}}$, the velocity toward the
preservation-aware endpoint $x_1^{\mathrm{tgt}}$ is
\begin{equation}
    v_{\tau}
    = \frac{x_1^{\mathrm{tgt}}-x_\tau^{\mathrm{base}}}{1-\tau}.
    \label{eq:terminal_velocity}
\end{equation}
We then use $v_{\tau}$ as the supervision target only at the terminal update, encouraging the velocity predicted from $x_{\tau}^{\mathrm{base}}$ to endpoint $x_{1}^{\mathrm{tgt}}$:
\begin{equation}
    \mathcal{L}_{\mathrm{Terminal}}
    =
    \mathbb{E}\!\left[
        \left\|
        v_\theta(x_\tau^{\mathrm{base}},\tau,c)
        - v_\tau
        \right\|_2^2
    \right],
    \label{eq:terminal_loss}
\end{equation}
where $c$ denotes the editing condition. This confines the intervention to the final sampling update, directly examining how much multi-turn stability can be improved by modifying only the last step.

We evaluate this terminal intervention on FLUX.2 [klein] Base 9B with 50 sampling steps per edit (Fig.~\ref{fig_4_terminal_intervention}(b)). When the base model is used for all 50 steps, the base model exhibits visible facial degradation after five turns, which worsens with continued editing. In contrast, using the base model for the first 49 steps and the fine-tuned model only for the final update markedly reduces visible drift and better preserves subject identity, with visually stable results even after 20 turns. This demonstrates that the preservation-aware endpoint provides effective supervision for long-horizon multi-turn stability, even when correction is confined to a single sampling update.

However, applying the fine-tuned model to all 50 steps reveals a different trade-off. Although the outputs exhibit fewer conspicuous artifacts than those of the base model, they become desaturated and overly smooth, losing facial and scene details. Since the model's parameters are shared across timesteps, terminal-only supervision can also disturb the model's learned dynamics at earlier steps; preserving the original model prior therefore requires explicit model switching. Moreover, concentrating all correction in the final update leaves no opportunity for gradual adjustment.  These limitations motivate our full TailSteer framework, which explicitly trains the model to preserve the model's earlier dynamics while distributing the correction over the denoising trajectory tail, allowing a single trained model to perform the entire sampling process with improved multi-turn stability.

\subsection{Tail Steering}
\label{sec_3_3_tail_steering}

\begin{figure}[t]
  \centering
  \includegraphics[width=\linewidth]{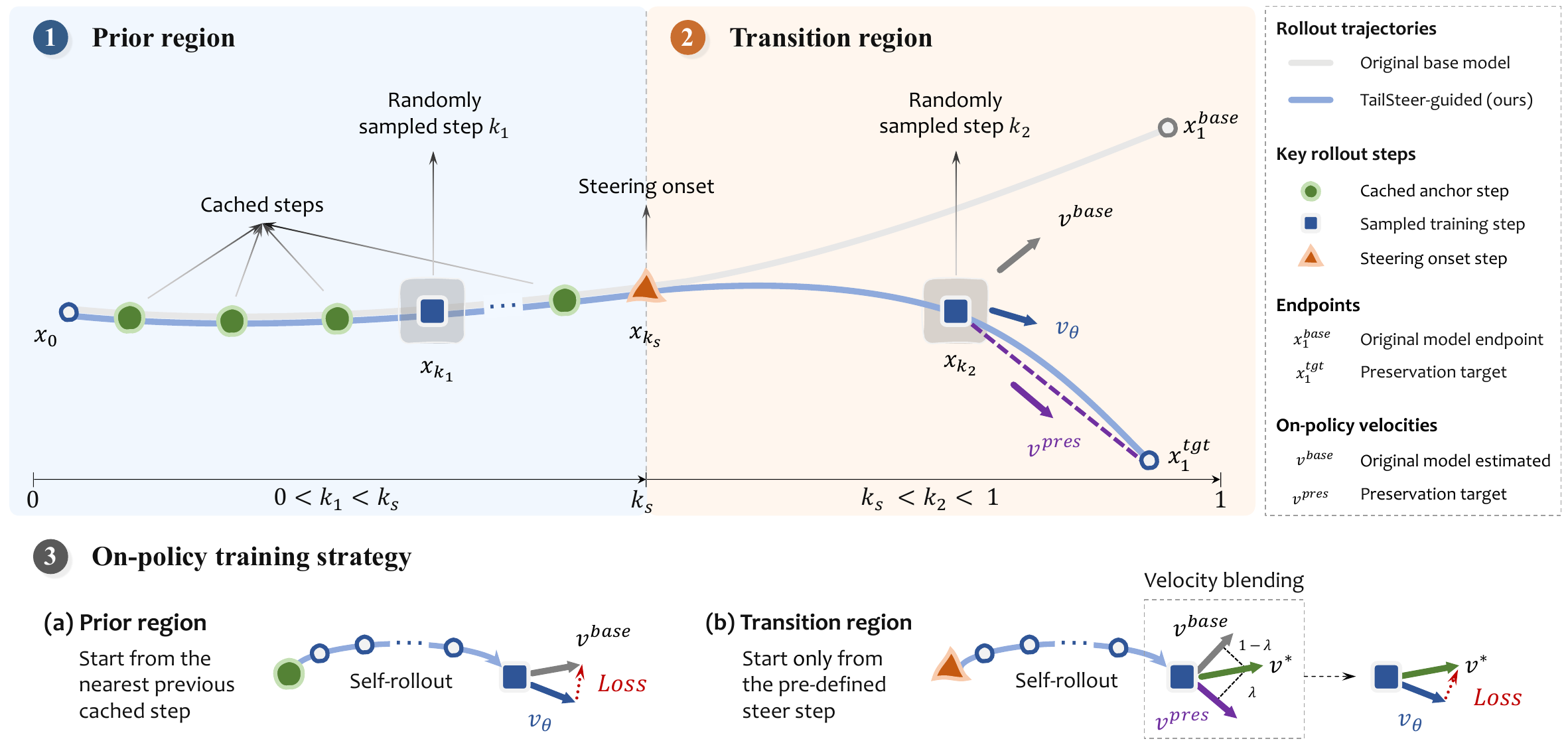}
  \caption{\textbf{On-policy training of TailSteer.}
    At randomly sampled training states $x_{k_1}$ and $x_{k_2}$, the refined
    model receives region-dependent supervision. In the prior region, its
    prediction is matched to the frozen base model's velocity
    $v^{\mathrm{base}}$; in the transition region, the target $v^*$ progressively
    blends $v^{\mathrm{base}}$ with the preservation velocity
    $v^{\mathrm{pres}}$ toward $x_1^{\mathrm{tgt}}$.
    To obtain these states on-policy, the refined model rolls out from the nearest
    preceding cached anchor in prior region, and from the fixed steering-onset
    state in transition region. The velocity target is then recomputed at the
    state actually reached by the refined model, with gradients applied only to
    its prediction at the sampled step.}
  \label{fig_5_framework}
\end{figure}

The terminal intervention above leaves earlier predictions unconstrained and concentrates all correction in a single update. We therefore introduce \textbf{TailSteer}, an on-policy last-stage training framework that preserves the model's learned dynamics over the earlier trajectory while progressively steering its tail toward the preservation-aware endpoint. Starting from the frozen base model, we initialize a trainable copy, denoted by $v_\theta$, as the \emph{refined model}. Importantly, all supervision is constructed on states reached by the refined model's own denoising rollouts, as illustrated in Fig.~\ref{fig_5_framework}.

\paragraph{Steering onset.}
We extend the terminal correction to a trajectory tail by introducing a steering onset $k_s<\tau$. This partitions the trajectory into a \emph{prior region}, $0\leq k\leq k_s$, and a \emph{transition region}, $k_s<k<1$. Before $k_s$, the refined model is trained to reproduce the original base model's dynamics; after $k_s$, the supervision gradually shifts toward the preservation endpoint.

\paragraph{Cached trajectory anchors.}
To obtain rollout states from the refined model without repeatedly replaying the full trajectory, we use cached base-model states as rollout anchors. In the prior region, states are cached at intervals of at most three sampling updates, including the onset state $x_{k_s}^{\mathrm{base}}$. For a sampled prior-region step, the refined model rolls out from the nearest strictly preceding cached state. In the transition region, all rollouts instead start from $x_{k_s}^{\mathrm{base}}$, with no intermediate tail states used as restart anchors. This keeps early rollouts short while allowing the refined model to evolve along its own trajectory throughout the tail. We denote the state reached at the sampled step by $x_k$.

\paragraph{Velocity blending.}
At the state $x_k$ reached by the refined model, we evaluate the frozen base model to obtain $v_k^{\mathrm{base}}=v^{\mathrm{base}}(x_k,k,c)$. We generalize the endpoint-directed velocity in Eq.~\ref{eq:terminal_velocity} to $v_k^{\mathrm{pres}}=(x_1^{\mathrm{tgt}}-x_k)/(1-k)$, where $1-k$ is the remaining sampling time. The supervision target combines these velocities according to the trajectory region:
\begin{equation}
    v_k^* =
    \begin{cases}
        v_k^{\mathrm{base}},
        & 0<k\leq k_s, \\[2pt]
        \lambda(k)v_k^{\mathrm{base}}
        +[1-\lambda(k)]v_k^{\mathrm{pres}},
        & k_s<k\leq\tau.
    \end{cases}
    \label{eq:tail_velocity}
\end{equation}
The weight $\lambda(k)$ decreases from $1$ at $k_s$ to $0$ at $\tau$, gradually shifting supervision from the base dynamics to endpoint correction. At the final update, the target has the same endpoint-directed form as in Sec.~\ref{sec_3_2_terminal_intervention}, now evaluated at the state reached by the refined model. 
% The weight schedule of $\lambda(k)$ is detailed in the Appendix.

\paragraph{On-policy training.}
Crucially, both $v_k^{\mathrm{base}}$ and $v_k^{\mathrm{pres}}$ are recomputed at the state $x_k$ actually reached by the refined model. The supervision therefore adapts to deviations along the current rollout instead of relying on targets precomputed on the frozen base trajectory. We optimize
\begin{equation}
\mathcal{L}_{\mathrm{TailSteer}}
=
\mathbb{E}\!\left[
\operatorname{MSE}\!\left(
v_\theta(x_k,k,c),
\operatorname{sg}[v_k^*]
\right)
\right],
\label{eq:tailsteer_loss}
\end{equation}
where the expectation is over cached examples and sampled training steps, and $\operatorname{sg}$ denotes stop-gradient. The rollout itself is performed without gradient tracking, so gradients are backpropagated only through the refined model's prediction at the sampled step, while all trainable parameters of the refined model are updated by this loss.

At inference, the refined model performs the entire sampling trajectory on its own. All auxiliary components---including the frozen base model, preservation targets, masks, and velocity blending---are used only during training TailSteer, requiring no model switching, external adapters, or additional consistency mechanisms at inference. TailSteer therefore improves multi-turn robustness as an intrinsic capability of the refined foundation model.
\section{Experiments}
\label{sec_4_experiments}

\begin{figure}[t]
  \centering
  \vspace{-12mm}
  \includegraphics[width=\linewidth]{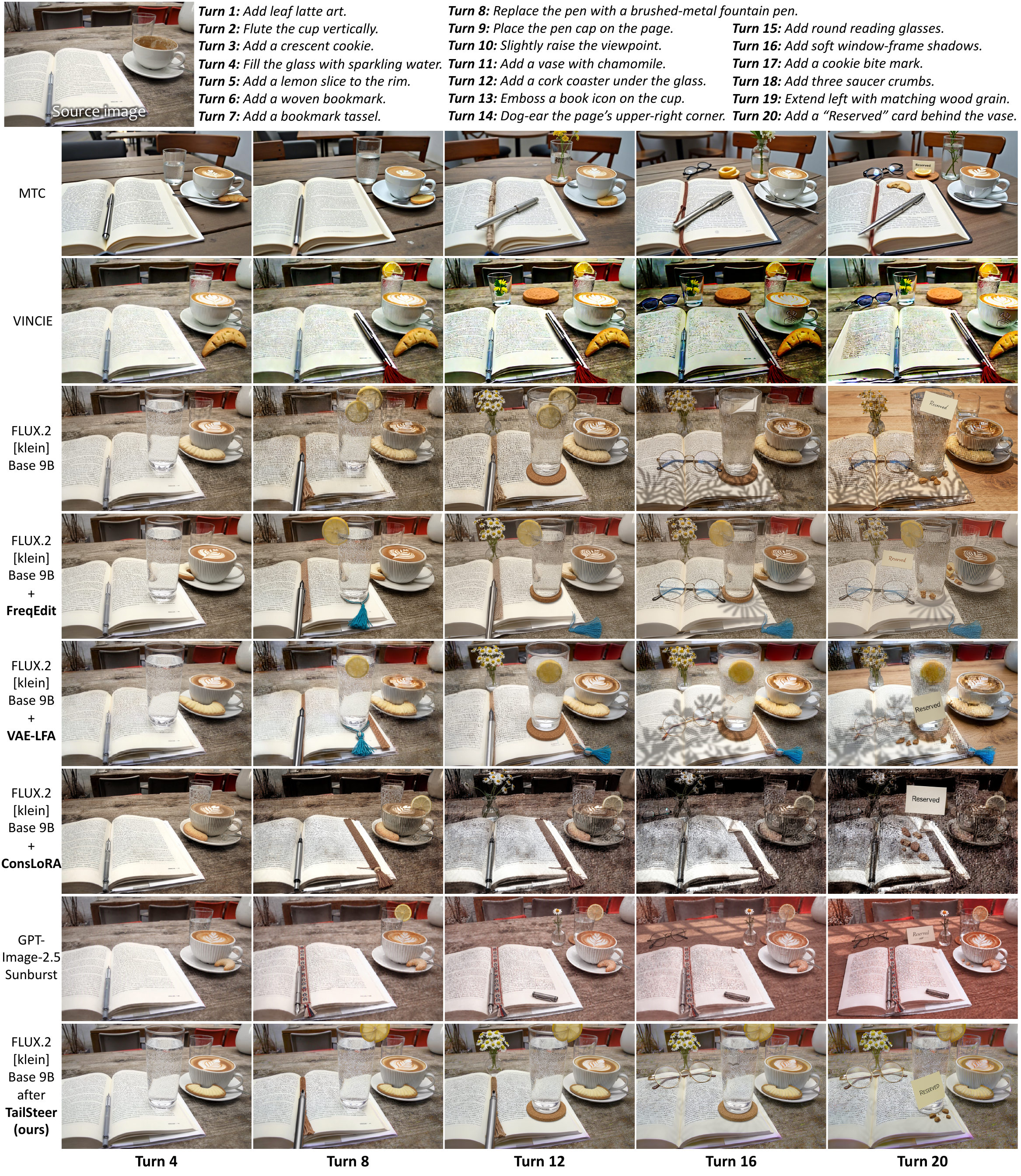}
  \vspace{-8mm}
    \caption{\textbf{20-turn qualitative comparison.}
While competing methods exhibit cumulative visual drift, TailSteer faithfully preserves the original tabletop color and previously established details, including the coffee cup, latte art, and cookie on the right, even after 20 successive editing turns.}
    \vspace{-2mm}
  \label{fig_6_main_experiment}
\end{figure}
\vspace{-1mm}
\subsection{Experimental Details}
\vspace{-2mm}
We evaluate TailSteer on four foundation models across three model families: FLUX.1 Kontext [dev] (30 steps)~\citep{labs2025flux1kontextflowmatching}, FLUX.2 [klein] Base 9B (50 steps)~\citep{flux2klein2026}, SenseNova-U1.5-8B-MoT (50 steps)~\citep{diao2026sensenova}, and Qwen-Image-2.1 (40 steps)~\citep{qwen2026image21}. We collect 11,947 candidate source images from Open Images~\citep{kuznetsova2020openimages} and MONET~\citep{aubin2026monet}, each paired with three independent local-edit instructions and one identity-edit instruction. Training trajectories are generated and filtered independently for each backbone. We train each model for 2,000 updates using BF16 precision and an effective batch size of 16 on eight NVIDIA H100 GPUs. We evaluate three steering onsets per backbone using the final checkpoints, with $k_s=25, 46, 42, 30$ selected for qualitative comparisons on Kontext, Klein, SenseNova, and Qwen, respectively. At inference, the refined model is applied throughout the denoising trajectory under the original sampling settings. The steering onset determines where preservation-oriented steering begins during training, rather than an inference-time model switch. 
% Detailed implementation and effective training strategies provided in the Appendix.
% Appendix~\ref{app}.

\subsection{Qualitative Evaluation}

\begin{figure}[t]
  \centering
  \vspace{-12mm}
  \includegraphics[width=\linewidth]{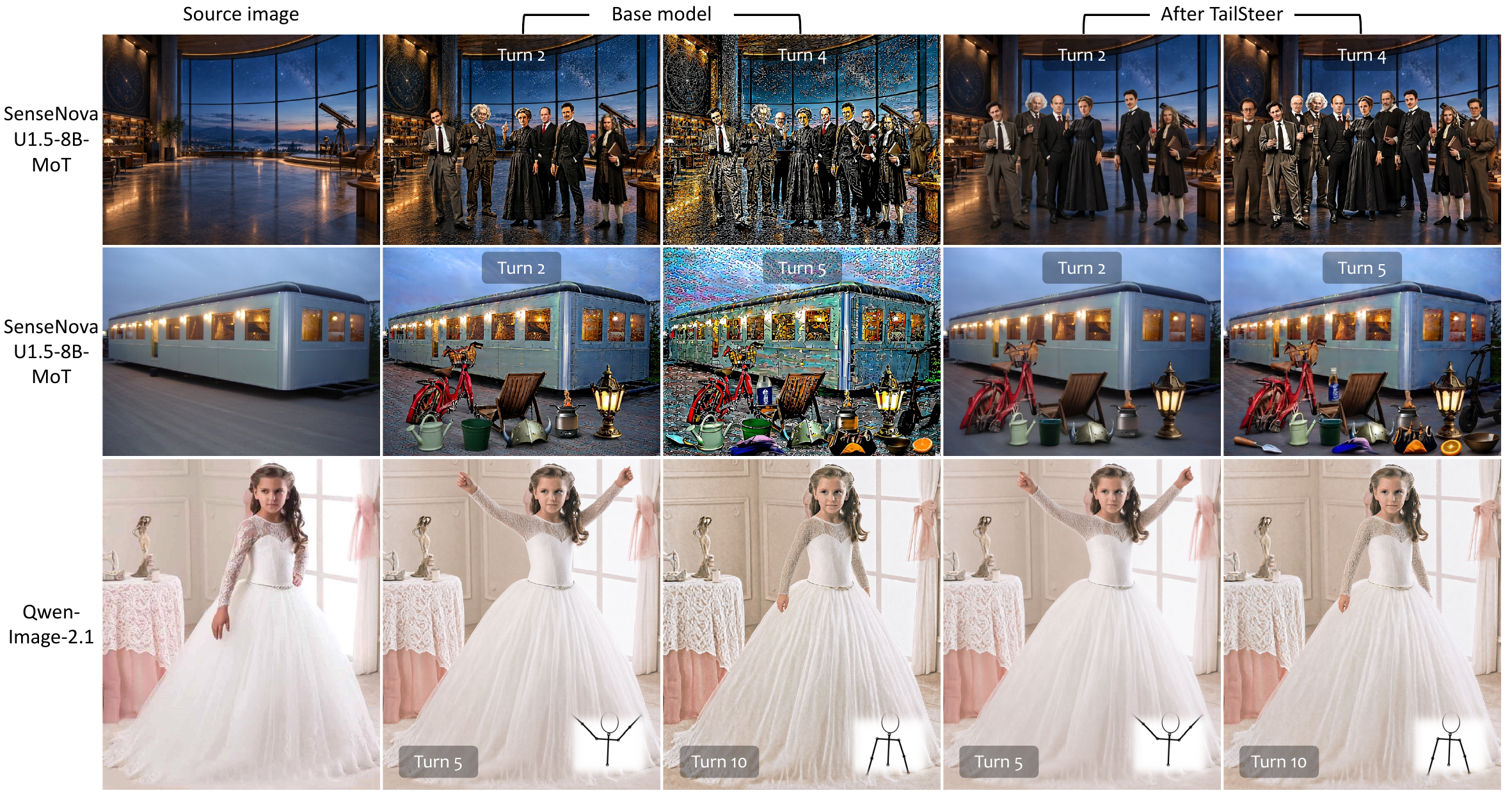}
  \vspace{-6mm}
    \caption{\textbf{Multi-input multi-turn editing.}
    TailSteer retains base models' multi-image editing capabilities while substantially reducing cumulative drift, preserving backgrounds during multi-person and multi-object compositing and the girl's dress details across successive skeleton-guided edits.}
    \vspace{-3mm}
  \label{fig_7_main_experiment_2}
\end{figure}

Fig.~\ref{fig_6_main_experiment} compares TailSteer with training-free MTC~\citep{zhou2025multiturn}, video-data-trained VINCIE~\citep{qu2026vincie}, and inference-time preservation modules, including FreqEdit~\citep{liao2026freqedit}, VAE-LFA~\citep{wang2026ditdrift}, and Consistency LoRA V2~\citep{dx8152flux2consistency}. We also evaluate GPT-Image-2.5 Sunburst~\citep{openai2026images25} as a strong proprietary generator via direct API calls. MTC maintains scene coherence but loses source-specific details, while VINCIE exhibits severe degradation across turns. Although the preservation modules alleviate some drift, they noticeably alter the base models' behavior and substantially degrade its single-turn editing performance, as shown in Table~\ref{tab:single-turn-preservation}. In contrast, TailSteer directly strengthens the base model, preserving the original tabletop color and previously established details, including the coffee cup, latte art, and cookie, even after 20 turns. Notably, TailSteer preserves the tabletop color more faithfully than GPT-Image-2.5 in this example, demonstrating that a open-weight model after TailSteer can rival or even surpass strong proprietary generators in long-horizon tasks. This advantage extends to multi-input editing (as shown in Fig.~\ref{fig_7_main_experiment_2}), where TailSteer retains the base models' compositing and skeleton-guided editing capabilities while better preserving backgrounds and the girl's dress details across pose changes.
% Additional comparisons are provided in the appendix.

\begin{table}[!t]
  \centering
  \definecolor{benchIncrease}{HTML}{147D73}
  \definecolor{benchDecrease}{HTML}{B65F2E}
  \colorlet{benchUp}{benchIncrease!75}
  \colorlet{benchDown}{benchDecrease!75}
  \vspace{-16mm}
  \caption{\textbf{Single-turn capability retention.} Text-to-image generation and single-turn editing performance across base models. \textcolor{benchUp}{Teal $\uparrow$} and \textcolor{benchDown}{orange $\downarrow$} indicate improvement/degradation from each base model. \textbf{Bold number} marks the best non-base score per backbone and metric, including ties.
  }
  \label{tab:single-turn-preservation}
  \begingroup
  \fontsize{6.5}{7.5}\selectfont
  \setlength{\tabcolsep}{0.8pt}
  \renewcommand{\arraystretch}{1}
  \newcommand{\benchdelta}[3][black!60]{\raisebox{-0.45ex}{{\fontsize{5}{6}\selectfont\textcolor{#1}{\makebox[0.65em][c]{$#2$}#3}}}}
  \newcommand{\benchscore}[2]{\mbox{#1\,#2}}
  \newcommand{\benchhead}[1]{{\fontsize{6}{7}\selectfont\mdseries #1}}
  \begin{tabular*}{\linewidth}{@{\extracolsep{\fill}}llc*{7}{c}@{}}
    \toprule
    \multirow{2}{*}{Backbone} & \multirow{2}{*}{Baseline} & \multirow{2}{*}{$k_s$}
      & \multicolumn{3}{c}{\textbf{Text-to-image}}
      & \multicolumn{4}{c}{\textbf{Single-turn image editing}} \\
    \cmidrule(lr){4-6}\cmidrule(l){7-10}
    & & & \benchhead{Qwen-IB EN $\uparrow$} & \benchhead{Qwen-IB ZH $\uparrow$}
      & \benchhead{BizGen ES (\%) $\uparrow$} & \benchhead{GEdit-EN $\uparrow$} & \benchhead{GEdit-ZH $\uparrow$}
      & \benchhead{RISE Acc. (\%) $\uparrow$} & \benchhead{ImgEdit $\uparrow$} \\
    \midrule
    \multirow{5}{*}{\shortstack[l]{FLUX.1\\Kontext [dev]}} & Base model & -- & \benchscore{41.19}{\phantom{\benchdelta{\uparrow}{0.00}}} & \benchscore{19.23}{\phantom{\benchdelta{\uparrow}{0.00}}} & \benchscore{\phantom{0}1.70}{\phantom{\benchdelta{\uparrow}{0.00}}} & \benchscore{5.99}{\phantom{\benchdelta{\uparrow}{0.00}}} & \benchscore{0.89}{\phantom{\benchdelta{\uparrow}{0.00}}} & \benchscore{\phantom{0}8.89}{\phantom{\benchdelta{\uparrow}{0.00}}} & \benchscore{3.77}{\phantom{\benchdelta{\uparrow}{0.00}}} \\
    \cmidrule[0.25pt](l){4-10}
     & W/ FreqEdit & -- & -- & -- & -- & \benchscore{5.20}{\benchdelta[benchDown]{\downarrow}{0.79}} & \benchscore{0.64}{\benchdelta[benchDown]{\downarrow}{0.25}} & \benchscore{\phantom{0}6.94}{\benchdelta[benchDown]{\downarrow}{1.94}} & \benchscore{3.62}{\benchdelta[benchDown]{\downarrow}{0.15}} \\
     & \multirow{3}{*}{After \method{}} & 22 & \benchscore{\textbf{41.63}}{\benchdelta[benchUp]{\uparrow}{0.44}} & \benchscore{19.17}{\benchdelta[benchDown]{\downarrow}{0.06}} & \benchscore{\textbf{\phantom{0}2.17}}{\benchdelta[benchUp]{\uparrow}{0.48}} & \benchscore{5.93}{\benchdelta[benchDown]{\downarrow}{0.06}} & \benchscore{0.81}{\benchdelta[benchDown]{\downarrow}{0.07}} & \benchscore{\phantom{0}6.94}{\benchdelta[benchDown]{\downarrow}{1.94}} & \benchscore{3.80}{\benchdelta[benchUp]{\uparrow}{0.03}} \\
     &  & 25 & \benchscore{41.50}{\benchdelta[benchUp]{\uparrow}{0.32}} & \benchscore{\textbf{19.19}}{\benchdelta[benchDown]{\downarrow}{0.04}} & \benchscore{\phantom{0}2.07}{\benchdelta[benchUp]{\uparrow}{0.38}} & \benchscore{5.95}{\benchdelta[benchDown]{\downarrow}{0.04}} & \benchscore{0.81}{\benchdelta[benchDown]{\downarrow}{0.07}} & \benchscore{\phantom{0}7.22}{\benchdelta[benchDown]{\downarrow}{1.67}} & \benchscore{\textbf{3.83}}{\benchdelta[benchUp]{\uparrow}{0.06}} \\
     &  & 27 & \benchscore{41.30}{\benchdelta[benchUp]{\uparrow}{0.11}} & \benchscore{18.89}{\benchdelta[benchDown]{\downarrow}{0.34}} & \benchscore{\phantom{0}1.70}{\benchdelta[black!60]{=}{0.00}} & \benchscore{\textbf{5.97}}{\benchdelta[benchDown]{\downarrow}{0.02}} & \benchscore{\textbf{0.86}}{\benchdelta[benchDown]{\downarrow}{0.02}} & \benchscore{\textbf{\phantom{0}7.50}}{\benchdelta[benchDown]{\downarrow}{1.39}} & \benchscore{\textbf{3.83}}{\benchdelta[benchUp]{\uparrow}{0.06}} \\
    \midrule
    \multirow{6}{*}{\shortstack[l]{FLUX.2\\{[Klein]}\\Base 9B}} & Base model & -- & \benchscore{49.27}{\phantom{\benchdelta{\uparrow}{0.00}}} & \benchscore{49.16}{\phantom{\benchdelta{\uparrow}{0.00}}} & \benchscore{\phantom{0}9.87}{\phantom{\benchdelta{\uparrow}{0.00}}} & \benchscore{7.47}{\phantom{\benchdelta{\uparrow}{0.00}}} & \benchscore{7.42}{\phantom{\benchdelta{\uparrow}{0.00}}} & \benchscore{14.17}{\phantom{\benchdelta{\uparrow}{0.00}}} & \benchscore{4.20}{\phantom{\benchdelta{\uparrow}{0.00}}} \\
    \cmidrule[0.25pt](l){4-10}
     & W/ FreqEdit & -- & -- & -- & -- & \benchscore{7.20}{\benchdelta[benchDown]{\downarrow}{0.27}} & \benchscore{7.18}{\benchdelta[benchDown]{\downarrow}{0.25}} & \benchscore{11.11}{\benchdelta[benchDown]{\downarrow}{3.06}} & \benchscore{4.09}{\benchdelta[benchDown]{\downarrow}{0.12}} \\
     & W/ ConsLoRA & -- & -- & -- & -- & \benchscore{6.58}{\benchdelta[benchDown]{\downarrow}{0.89}} & \benchscore{6.43}{\benchdelta[benchDown]{\downarrow}{0.99}} & \benchscore{\phantom{0}9.44}{\benchdelta[benchDown]{\downarrow}{4.72}} & \benchscore{3.86}{\benchdelta[benchDown]{\downarrow}{0.34}} \\
     & \multirow{3}{*}{After \method{}} & 43 & \benchscore{49.68}{\benchdelta[benchUp]{\uparrow}{0.41}} & \benchscore{\textbf{49.48}}{\benchdelta[benchUp]{\uparrow}{0.33}} & \benchscore{\phantom{0}8.30}{\benchdelta[benchDown]{\downarrow}{1.57}} & \benchscore{7.43}{\benchdelta[benchDown]{\downarrow}{0.04}} & \benchscore{7.35}{\benchdelta[benchDown]{\downarrow}{0.07}} & \benchscore{13.33}{\benchdelta[benchDown]{\downarrow}{0.83}} & \benchscore{4.17}{\benchdelta[benchDown]{\downarrow}{0.03}} \\
     &  & 46 & \benchscore{49.42}{\benchdelta[benchUp]{\uparrow}{0.15}} & \benchscore{49.37}{\benchdelta[benchUp]{\uparrow}{0.21}} & \benchscore{\textbf{10.97}}{\benchdelta[benchUp]{\uparrow}{1.10}} & \benchscore{\textbf{7.52}}{\benchdelta[benchUp]{\uparrow}{0.05}} & \benchscore{7.39}{\benchdelta[benchDown]{\downarrow}{0.04}} & \benchscore{\textbf{13.61}}{\benchdelta[benchDown]{\downarrow}{0.56}} & \benchscore{4.19}{\benchdelta[benchDown]{\downarrow}{0.02}} \\
     &  & 48 & \benchscore{\textbf{49.74}}{\benchdelta[benchUp]{\uparrow}{0.47}} & \benchscore{49.46}{\benchdelta[benchUp]{\uparrow}{0.31}} & \benchscore{\phantom{0}9.37}{\benchdelta[benchDown]{\downarrow}{0.50}} & \benchscore{7.41}{\benchdelta[benchDown]{\downarrow}{0.06}} & \benchscore{\textbf{7.44}}{\benchdelta[benchUp]{\uparrow}{0.01}} & \benchscore{11.39}{\benchdelta[benchDown]{\downarrow}{2.78}} & \benchscore{\textbf{4.24}}{\benchdelta[benchUp]{\uparrow}{0.03}} \\
    \midrule
    \multirow{4}{*}{\shortstack[l]{SenseNova\\U1.5-8B\\MoT}} & Base model & -- & \benchscore{53.43}{\phantom{\benchdelta{\uparrow}{0.00}}} & \benchscore{53.25}{\phantom{\benchdelta{\uparrow}{0.00}}} & \benchscore{45.85}{\phantom{\benchdelta{\uparrow}{0.00}}} & \benchscore{8.33}{\phantom{\benchdelta{\uparrow}{0.00}}} & \benchscore{8.30}{\phantom{\benchdelta{\uparrow}{0.00}}} & \benchscore{31.39}{\phantom{\benchdelta{\uparrow}{0.00}}} & \benchscore{4.43}{\phantom{\benchdelta{\uparrow}{0.00}}} \\
    \cmidrule[0.25pt](l){4-10}
     & \multirow{3}{*}{After \method{}} & 37 & \benchscore{52.45}{\benchdelta[benchDown]{\downarrow}{0.98}} & \benchscore{52.78}{\benchdelta[benchDown]{\downarrow}{0.46}} & \benchscore{43.75}{\benchdelta[benchDown]{\downarrow}{2.10}} & \benchscore{\textbf{8.30}}{\benchdelta[benchDown]{\downarrow}{0.04}} & \benchscore{8.28}{\benchdelta[benchDown]{\downarrow}{0.02}} & \benchscore{\textbf{31.67}}{\benchdelta[benchUp]{\uparrow}{0.28}} & \benchscore{4.39}{\benchdelta[benchDown]{\downarrow}{0.05}} \\
     &  & 42 & \benchscore{52.24}{\benchdelta[benchDown]{\downarrow}{1.19}} & \benchscore{52.58}{\benchdelta[benchDown]{\downarrow}{0.67}} & \benchscore{45.73}{\benchdelta[benchDown]{\downarrow}{0.12}} & \benchscore{8.27}{\benchdelta[benchDown]{\downarrow}{0.06}} & \benchscore{\textbf{8.31}}{\benchdelta[benchUp]{\uparrow}{0.01}} & \benchscore{29.44}{\benchdelta[benchDown]{\downarrow}{1.94}} & \benchscore{\textbf{4.41}}{\benchdelta[benchDown]{\downarrow}{0.02}} \\
     &  & 47 & \benchscore{\textbf{52.55}}{\benchdelta[benchDown]{\downarrow}{0.89}} & \benchscore{\textbf{53.02}}{\benchdelta[benchDown]{\downarrow}{0.22}} & \benchscore{\textbf{46.17}}{\benchdelta[benchUp]{\uparrow}{0.32}} & \benchscore{\textbf{8.30}}{\benchdelta[benchDown]{\downarrow}{0.04}} & \benchscore{8.25}{\benchdelta[benchDown]{\downarrow}{0.05}} & \benchscore{\textbf{31.67}}{\benchdelta[benchUp]{\uparrow}{0.28}} & \benchscore{4.39}{\benchdelta[benchDown]{\downarrow}{0.05}} \\
    \midrule
    \multirow{4}{*}{\shortstack[l]{Qwen\\Image-2.1}} & Base model & -- & \benchscore{53.39}{\phantom{\benchdelta{\uparrow}{0.00}}} & \benchscore{53.34}{\phantom{\benchdelta{\uparrow}{0.00}}} & \benchscore{57.03}{\phantom{\benchdelta{\uparrow}{0.00}}} & \benchscore{8.11}{\phantom{\benchdelta{\uparrow}{0.00}}} & \benchscore{7.80}{\phantom{\benchdelta{\uparrow}{0.00}}} & \benchscore{20.56}{\phantom{\benchdelta{\uparrow}{0.00}}} & \benchscore{4.40}{\phantom{\benchdelta{\uparrow}{0.00}}} \\
    \cmidrule[0.25pt](l){4-10}
     & \multirow{3}{*}{After \method{}} & 24 & \benchscore{\textbf{53.49}}{\benchdelta[benchUp]{\uparrow}{0.10}} & \benchscore{52.86}{\benchdelta[benchDown]{\downarrow}{0.48}} & \benchscore{56.25}{\benchdelta[benchDown]{\downarrow}{0.78}} & \benchscore{7.96}{\benchdelta[benchDown]{\downarrow}{0.15}} & \benchscore{7.73}{\benchdelta[benchDown]{\downarrow}{0.07}} & \benchscore{18.33}{\benchdelta[benchDown]{\downarrow}{2.22}} & \benchscore{4.36}{\benchdelta[benchDown]{\downarrow}{0.04}} \\
     &  & 30 & \benchscore{53.06}{\benchdelta[benchDown]{\downarrow}{0.33}} & \benchscore{\textbf{53.20}}{\benchdelta[benchDown]{\downarrow}{0.14}} & \benchscore{\textbf{57.38}}{\benchdelta[benchUp]{\uparrow}{0.35}} & \benchscore{\textbf{8.04}}{\benchdelta[benchDown]{\downarrow}{0.06}} & \benchscore{7.72}{\benchdelta[benchDown]{\downarrow}{0.09}} & \benchscore{18.06}{\benchdelta[benchDown]{\downarrow}{2.50}} & \benchscore{\textbf{4.42}}{\benchdelta[benchUp]{\uparrow}{0.02}} \\
     &  & 35 & \benchscore{53.32}{\benchdelta[benchDown]{\downarrow}{0.07}} & \benchscore{53.06}{\benchdelta[benchDown]{\downarrow}{0.28}} & \benchscore{56.67}{\benchdelta[benchDown]{\downarrow}{0.35}} & \benchscore{8.02}{\benchdelta[benchDown]{\downarrow}{0.09}} & \benchscore{\textbf{7.75}}{\benchdelta[benchDown]{\downarrow}{0.05}} & \benchscore{\textbf{18.61}}{\benchdelta[benchDown]{\downarrow}{1.94}} & \benchscore{4.39}{\benchdelta[benchDown]{\downarrow}{0.01}} \\
    \bottomrule
  \end{tabular*}
  \endgroup
  \par
  \vspace{-1mm}
  \caption{\textbf{Multi-turn editing and drift evaluation.} MSE-Bench, EdiVal assess general multi-turn editing, while Add \& Preserve measures Cumulative Object Accuracy (COA) and outside-box drift.}
  \label{tab:multi-turn-preservation}
  \begingroup
  \fontsize{6.5}{7.5}\selectfont
  \setlength{\tabcolsep}{0.8pt}
  \renewcommand{\arraystretch}{1}
  \newcommand{\benchdelta}[3][black!60]{\raisebox{-0.45ex}{{\fontsize{5}{6}\selectfont\textcolor{#1}{\makebox[0.65em][c]{$#2$}#3}}}}
  \newcommand{\benchscore}[2]{\mbox{#1\,#2}}
  \newcommand{\benchhead}[1]{{\fontsize{6}{7}\selectfont\mdseries #1}}
  \begin{tabular*}{\linewidth}{@{\extracolsep{\fill}}llc*{9}{c}@{}}
    \toprule
    \multirow{2}{*}{Backbone} & \multirow{2}{*}{Baseline} & \multirow{2}{*}{$k_s$}
      & \multicolumn{3}{c}{\textbf{MSE-Bench}}
      & \multicolumn{2}{c}{\textbf{EdiVal}} & \multicolumn{4}{c}{\textbf{Add \& Preserve}} \\
    \cmidrule(lr){4-6}\cmidrule(lr){7-8}\cmidrule(l){9-12}
    & & & \benchhead{SR@5 (\%) $\uparrow$} & \benchhead{PF@5 $\uparrow$} & \benchhead{Cons.@5 $\uparrow$}
      & \benchhead{IF@3 (\%) $\uparrow$} & \benchhead{CC@3 (\%) $\uparrow$} & \benchhead{COA (\%) $\uparrow$} & \benchhead{PSNR@10 $\uparrow$} & \benchhead{SSIM@10 $\uparrow$} & \benchhead{LPIPS@10 $\downarrow$} \\
    \midrule
    \multirow{6}{*}{\shortstack[l]{FLUX.1\\Kontext [dev]}} & Base model & -- & \benchscore{16.00}{\phantom{\benchdelta{\uparrow}{0.00}}} & \benchscore{1.69}{\phantom{\benchdelta{\uparrow}{0.00}}} & \benchscore{2.08}{\phantom{\benchdelta{\uparrow}{0.00}}} & \benchscore{17.31}{\phantom{\benchdelta{\uparrow}{0.00}}} & \benchscore{87.46}{\phantom{\benchdelta{\uparrow}{0.00}}} & \benchscore{34.77}{\phantom{\benchdelta{\uparrow}{0.00}}} & \benchscore{18.64}{\phantom{\benchdelta{\uparrow}{0.00}}} & \benchscore{0.585}{\phantom{\benchdelta{\uparrow}{0.000}}} & \benchscore{0.388}{\phantom{\benchdelta{\uparrow}{0.000}}} \\
    \cmidrule[0.25pt](l){4-12}
     & W/ FreqEdit & -- & \benchscore{10.00}{\benchdelta[benchDown]{\downarrow}{6.00}} & \benchscore{1.02}{\benchdelta[benchDown]{\downarrow}{0.67}} & \benchscore{1.09}{\benchdelta[benchDown]{\downarrow}{0.99}} & \benchscore{11.54}{\benchdelta[benchDown]{\downarrow}{5.77}} & \benchscore{\textbf{92.68}}{\benchdelta[benchUp]{\uparrow}{5.22}} & \benchscore{\textbf{36.82}}{\benchdelta[benchUp]{\uparrow}{2.04}} & \benchscore{17.20}{\benchdelta[benchDown]{\downarrow}{1.44}} & \benchscore{0.519}{\benchdelta[benchDown]{\downarrow}{0.066}} & \benchscore{0.278}{\benchdelta[benchUp]{\downarrow}{0.110}} \\
     & W/ VAE-LFA & -- & \benchscore{10.00}{\benchdelta[benchDown]{\downarrow}{6.00}} & \benchscore{1.00}{\benchdelta[benchDown]{\downarrow}{0.69}} & \benchscore{1.10}{\benchdelta[benchDown]{\downarrow}{0.98}} & \benchscore{16.78}{\benchdelta[benchDown]{\downarrow}{0.52}} & \benchscore{86.39}{\benchdelta[benchDown]{\downarrow}{1.07}} & \benchscore{35.07}{\benchdelta[benchUp]{\uparrow}{0.30}} & \benchscore{19.65}{\benchdelta[benchUp]{\uparrow}{1.01}} & \benchscore{0.602}{\benchdelta[benchUp]{\uparrow}{0.017}} & \benchscore{0.363}{\benchdelta[benchUp]{\downarrow}{0.025}} \\
     & \multirow{3}{*}{After \method{}} & 22 & \benchscore{13.00}{\benchdelta[benchDown]{\downarrow}{3.00}} & \benchscore{1.43}{\benchdelta[benchDown]{\downarrow}{0.26}} & \benchscore{1.68}{\benchdelta[benchDown]{\downarrow}{0.40}} & \benchscore{16.26}{\benchdelta[benchDown]{\downarrow}{1.05}} & \benchscore{89.31}{\benchdelta[benchUp]{\uparrow}{1.85}} & \benchscore{35.03}{\benchdelta[benchUp]{\uparrow}{0.26}} & \benchscore{20.60}{\benchdelta[benchUp]{\uparrow}{1.96}} & \benchscore{0.671}{\benchdelta[benchUp]{\uparrow}{0.086}} & \benchscore{0.268}{\benchdelta[benchUp]{\downarrow}{0.121}} \\
     &  & 25 & \benchscore{\textbf{17.00}}{\benchdelta[benchUp]{\uparrow}{1.00}} & \benchscore{\textbf{1.60}}{\benchdelta[benchDown]{\downarrow}{0.09}} & \benchscore{\textbf{1.74}}{\benchdelta[benchDown]{\downarrow}{0.34}} & \benchscore{\textbf{16.96}}{\benchdelta[benchDown]{\downarrow}{0.35}} & \benchscore{89.08}{\benchdelta[benchUp]{\uparrow}{1.62}} & \benchscore{34.84}{\benchdelta[benchUp]{\uparrow}{0.07}} & \benchscore{\textbf{21.02}}{\benchdelta[benchUp]{\uparrow}{2.38}} & \benchscore{\textbf{0.672}}{\benchdelta[benchUp]{\uparrow}{0.087}} & \benchscore{\textbf{0.257}}{\benchdelta[benchUp]{\downarrow}{0.131}} \\
     &  & 27 & \benchscore{16.00}{\benchdelta[black!60]{=}{0.00}} & \benchscore{1.51}{\benchdelta[benchDown]{\downarrow}{0.18}} & \benchscore{1.72}{\benchdelta[benchDown]{\downarrow}{0.36}} & \benchscore{16.78}{\benchdelta[benchDown]{\downarrow}{0.52}} & \benchscore{88.70}{\benchdelta[benchUp]{\uparrow}{1.24}} & \benchscore{34.58}{\benchdelta[benchDown]{\downarrow}{0.19}} & \benchscore{20.10}{\benchdelta[benchUp]{\uparrow}{1.46}} & \benchscore{0.646}{\benchdelta[benchUp]{\uparrow}{0.061}} & \benchscore{0.297}{\benchdelta[benchUp]{\downarrow}{0.092}} \\
    \midrule
    \multirow{7}{*}{\shortstack[l]{FLUX.2\\{[Klein]}\\Base 9B}} & Base model & -- & \benchscore{40.00}{\phantom{\benchdelta{\uparrow}{0.00}}} & \benchscore{3.91}{\phantom{\benchdelta{\uparrow}{0.00}}} & \benchscore{4.03}{\phantom{\benchdelta{\uparrow}{0.00}}} & \benchscore{32.69}{\phantom{\benchdelta{\uparrow}{0.00}}} & \benchscore{85.24}{\phantom{\benchdelta{\uparrow}{0.00}}} & \benchscore{31.02}{\phantom{\benchdelta{\uparrow}{0.00}}} & \benchscore{13.85}{\phantom{\benchdelta{\uparrow}{0.00}}} & \benchscore{0.289}{\phantom{\benchdelta{\uparrow}{0.000}}} & \benchscore{0.524}{\phantom{\benchdelta{\uparrow}{0.000}}} \\
    \cmidrule[0.25pt](l){4-12}
     & W/ FreqEdit & -- & \benchscore{33.00}{\benchdelta[benchDown]{\downarrow}{7.00}} & \benchscore{3.69}{\benchdelta[benchDown]{\downarrow}{0.22}} & \benchscore{4.11}{\benchdelta[benchUp]{\uparrow}{0.08}} & \benchscore{27.97}{\benchdelta[benchDown]{\downarrow}{4.72}} & \benchscore{87.84}{\benchdelta[benchUp]{\uparrow}{2.60}} & \benchscore{\textbf{35.94}}{\benchdelta[benchUp]{\uparrow}{4.91}} & \benchscore{13.80}{\benchdelta[benchDown]{\downarrow}{0.05}} & \benchscore{0.318}{\benchdelta[benchUp]{\uparrow}{0.030}} & \benchscore{0.405}{\benchdelta[benchUp]{\downarrow}{0.119}} \\
     & W/ VAE-LFA & -- & \benchscore{33.00}{\benchdelta[benchDown]{\downarrow}{7.00}} & \benchscore{3.38}{\benchdelta[benchDown]{\downarrow}{0.53}} & \benchscore{3.56}{\benchdelta[benchDown]{\downarrow}{0.47}} & \benchscore{\textbf{31.47}}{\benchdelta[benchDown]{\downarrow}{1.22}} & \benchscore{85.93}{\benchdelta[benchUp]{\uparrow}{0.68}} & \benchscore{31.03}{\benchdelta[benchUp]{\uparrow}{0.01}} & \benchscore{16.47}{\benchdelta[benchUp]{\uparrow}{2.61}} & \benchscore{0.434}{\benchdelta[benchUp]{\uparrow}{0.146}} & \benchscore{0.375}{\benchdelta[benchUp]{\downarrow}{0.149}} \\
     & W/ ConsLoRA & -- & \benchscore{39.00}{\benchdelta[benchDown]{\downarrow}{1.00}} & \benchscore{3.94}{\benchdelta[benchUp]{\uparrow}{0.03}} & \benchscore{\textbf{4.47}}{\benchdelta[benchUp]{\uparrow}{0.44}} & \benchscore{26.05}{\benchdelta[benchDown]{\downarrow}{6.64}} & \benchscore{\textbf{91.19}}{\benchdelta[benchUp]{\uparrow}{5.95}} & \benchscore{32.87}{\benchdelta[benchUp]{\uparrow}{1.85}} & \benchscore{14.63}{\benchdelta[benchUp]{\uparrow}{0.77}} & \benchscore{0.446}{\benchdelta[benchUp]{\uparrow}{0.158}} & \benchscore{0.442}{\benchdelta[benchUp]{\downarrow}{0.082}} \\
     & \multirow{3}{*}{After \method{}} & 43 & \benchscore{32.00}{\benchdelta[benchDown]{\downarrow}{8.00}} & \benchscore{3.26}{\benchdelta[benchDown]{\downarrow}{0.65}} & \benchscore{3.47}{\benchdelta[benchDown]{\downarrow}{0.56}} & \benchscore{30.94}{\benchdelta[benchDown]{\downarrow}{1.75}} & \benchscore{87.38}{\benchdelta[benchUp]{\uparrow}{2.14}} & \benchscore{30.92}{\benchdelta[benchDown]{\downarrow}{0.10}} & \benchscore{19.76}{\benchdelta[benchUp]{\uparrow}{5.91}} & \benchscore{0.616}{\benchdelta[benchUp]{\uparrow}{0.328}} & \benchscore{0.295}{\benchdelta[benchUp]{\downarrow}{0.229}} \\
     &  & 46 & \benchscore{33.00}{\benchdelta[benchDown]{\downarrow}{7.00}} & \benchscore{3.23}{\benchdelta[benchDown]{\downarrow}{0.68}} & \benchscore{3.32}{\benchdelta[benchDown]{\downarrow}{0.71}} & \benchscore{30.24}{\benchdelta[benchDown]{\downarrow}{2.45}} & \benchscore{87.57}{\benchdelta[benchUp]{\uparrow}{2.32}} & \benchscore{31.84}{\benchdelta[benchUp]{\uparrow}{0.81}} & \benchscore{\textbf{19.96}}{\benchdelta[benchUp]{\uparrow}{6.11}} & \benchscore{\textbf{0.621}}{\benchdelta[benchUp]{\uparrow}{0.333}} & \benchscore{\textbf{0.290}}{\benchdelta[benchUp]{\downarrow}{0.235}} \\
     &  & 48 & \benchscore{\textbf{41.00}}{\benchdelta[benchUp]{\uparrow}{1.00}} & \benchscore{\textbf{4.05}}{\benchdelta[benchUp]{\uparrow}{0.14}} & \benchscore{4.32}{\benchdelta[benchUp]{\uparrow}{0.29}} & \benchscore{29.72}{\benchdelta[benchDown]{\downarrow}{2.97}} & \benchscore{86.61}{\benchdelta[benchUp]{\uparrow}{1.37}} & \benchscore{31.61}{\benchdelta[benchUp]{\uparrow}{0.59}} & \benchscore{19.56}{\benchdelta[benchUp]{\uparrow}{5.71}} & \benchscore{0.594}{\benchdelta[benchUp]{\uparrow}{0.305}} & \benchscore{0.317}{\benchdelta[benchUp]{\downarrow}{0.207}} \\
    \midrule
    \multirow{4}{*}{\shortstack[l]{SenseNova\\U1.5-8B\\MoT}} & Base model & -- & \benchscore{50.00}{\phantom{\benchdelta{\uparrow}{0.00}}} & \benchscore{4.88}{\phantom{\benchdelta{\uparrow}{0.00}}} & \benchscore{5.30}{\phantom{\benchdelta{\uparrow}{0.00}}} & \benchscore{40.21}{\phantom{\benchdelta{\uparrow}{0.00}}} & \benchscore{71.59}{\phantom{\benchdelta{\uparrow}{0.00}}} & \benchscore{94.93}{\phantom{\benchdelta{\uparrow}{0.00}}} & \benchscore{11.23}{\phantom{\benchdelta{\uparrow}{0.00}}} & \benchscore{0.166}{\phantom{\benchdelta{\uparrow}{0.000}}} & \benchscore{0.607}{\phantom{\benchdelta{\uparrow}{0.000}}} \\
    \cmidrule[0.25pt](l){4-12}
     & \multirow{3}{*}{After \method{}} & 37 & \benchscore{54.00}{\benchdelta[benchUp]{\uparrow}{4.00}} & \benchscore{\textbf{5.57}}{\benchdelta[benchUp]{\uparrow}{0.69}} & \benchscore{\textbf{5.83}}{\benchdelta[benchUp]{\uparrow}{0.53}} & \benchscore{39.51}{\benchdelta[benchDown]{\downarrow}{0.70}} & \benchscore{\textbf{77.16}}{\benchdelta[benchUp]{\uparrow}{5.57}} & \benchscore{93.67}{\benchdelta[benchDown]{\downarrow}{1.26}} & \benchscore{\textbf{18.22}}{\benchdelta[benchUp]{\uparrow}{6.99}} & \benchscore{\textbf{0.493}}{\benchdelta[benchUp]{\uparrow}{0.327}} & \benchscore{\textbf{0.415}}{\benchdelta[benchUp]{\downarrow}{0.192}} \\
     &  & 42 & \benchscore{\textbf{55.00}}{\benchdelta[benchUp]{\uparrow}{5.00}} & \benchscore{5.40}{\benchdelta[benchUp]{\uparrow}{0.52}} & \benchscore{5.64}{\benchdelta[benchUp]{\uparrow}{0.34}} & \benchscore{\textbf{42.31}}{\benchdelta[benchUp]{\uparrow}{2.10}} & \benchscore{77.03}{\benchdelta[benchUp]{\uparrow}{5.44}} & \benchscore{94.78}{\benchdelta[benchDown]{\downarrow}{0.15}} & \benchscore{\textbf{18.22}}{\benchdelta[benchUp]{\uparrow}{6.99}} & \benchscore{0.489}{\benchdelta[benchUp]{\uparrow}{0.323}} & \benchscore{0.417}{\benchdelta[benchUp]{\downarrow}{0.190}} \\
     &  & 47 & \benchscore{\textbf{55.00}}{\benchdelta[benchUp]{\uparrow}{5.00}} & \benchscore{5.34}{\benchdelta[benchUp]{\uparrow}{0.46}} & \benchscore{5.63}{\benchdelta[benchUp]{\uparrow}{0.33}} & \benchscore{39.16}{\benchdelta[benchDown]{\downarrow}{1.05}} & \benchscore{76.36}{\benchdelta[benchUp]{\uparrow}{4.77}} & \benchscore{\textbf{95.39}}{\benchdelta[benchUp]{\uparrow}{0.46}} & \benchscore{17.47}{\benchdelta[benchUp]{\uparrow}{6.24}} & \benchscore{0.452}{\benchdelta[benchUp]{\uparrow}{0.286}} & \benchscore{0.444}{\benchdelta[benchUp]{\downarrow}{0.163}} \\
    \midrule
    \multirow{4}{*}{\shortstack[l]{Qwen\\Image-2.1}} & Base model & -- & \benchscore{42.00}{\phantom{\benchdelta{\uparrow}{0.00}}} & \benchscore{4.41}{\phantom{\benchdelta{\uparrow}{0.00}}} & \benchscore{4.53}{\phantom{\benchdelta{\uparrow}{0.00}}} & \benchscore{26.22}{\phantom{\benchdelta{\uparrow}{0.00}}} & \benchscore{89.02}{\phantom{\benchdelta{\uparrow}{0.00}}} & \benchscore{67.36}{\phantom{\benchdelta{\uparrow}{0.00}}} & \benchscore{16.93}{\phantom{\benchdelta{\uparrow}{0.00}}} & \benchscore{0.438}{\phantom{\benchdelta{\uparrow}{0.000}}} & \benchscore{0.270}{\phantom{\benchdelta{\uparrow}{0.000}}} \\
    \cmidrule[0.25pt](l){4-12}
     & \multirow{3}{*}{After \method{}} & 24 & \benchscore{38.00}{\benchdelta[benchDown]{\downarrow}{4.00}} & \benchscore{3.93}{\benchdelta[benchDown]{\downarrow}{0.48}} & \benchscore{4.19}{\benchdelta[benchDown]{\downarrow}{0.34}} & \benchscore{24.13}{\benchdelta[benchDown]{\downarrow}{2.10}} & \benchscore{89.66}{\benchdelta[benchUp]{\uparrow}{0.64}} & \benchscore{66.31}{\benchdelta[benchDown]{\downarrow}{1.05}} & \benchscore{\textbf{18.38}}{\benchdelta[benchUp]{\uparrow}{1.45}} & \benchscore{\textbf{0.524}}{\benchdelta[benchUp]{\uparrow}{0.086}} & \benchscore{\textbf{0.233}}{\benchdelta[benchUp]{\downarrow}{0.037}} \\
     &  & 30 & \benchscore{\textbf{44.00}}{\benchdelta[benchUp]{\uparrow}{2.00}} & \benchscore{\textbf{4.67}}{\benchdelta[benchUp]{\uparrow}{0.26}} & \benchscore{\textbf{5.04}}{\benchdelta[benchUp]{\uparrow}{0.51}} & \benchscore{\textbf{24.83}}{\benchdelta[benchDown]{\downarrow}{1.40}} & \benchscore{\textbf{89.70}}{\benchdelta[benchUp]{\uparrow}{0.68}} & \benchscore{\textbf{66.94}}{\benchdelta[benchDown]{\downarrow}{0.43}} & \benchscore{18.15}{\benchdelta[benchUp]{\uparrow}{1.22}} & \benchscore{0.513}{\benchdelta[benchUp]{\uparrow}{0.075}} & \benchscore{0.239}{\benchdelta[benchUp]{\downarrow}{0.031}} \\
     &  & 35 & \benchscore{38.00}{\benchdelta[benchDown]{\downarrow}{4.00}} & \benchscore{3.92}{\benchdelta[benchDown]{\downarrow}{0.49}} & \benchscore{4.39}{\benchdelta[benchDown]{\downarrow}{0.14}} & \benchscore{23.95}{\benchdelta[benchDown]{\downarrow}{2.27}} & \benchscore{89.31}{\benchdelta[benchUp]{\uparrow}{0.29}} & \benchscore{65.68}{\benchdelta[benchDown]{\downarrow}{1.68}} & \benchscore{17.74}{\benchdelta[benchUp]{\uparrow}{0.82}} & \benchscore{0.493}{\benchdelta[benchUp]{\uparrow}{0.056}} & \benchscore{0.249}{\benchdelta[benchUp]{\downarrow}{0.021}} \\
    \midrule
    \multicolumn{2}{c}{{\fontsize{5.5}{6.5}\selectfont GPT-Image-2.5-Sunburst (API)}} & -- & \benchscore{48.45}{\phantom{\benchdelta{\uparrow}{0.00}}} & \benchscore{4.52}{\phantom{\benchdelta{\uparrow}{0.00}}} & \benchscore{4.67}{\phantom{\benchdelta{\uparrow}{0.00}}} & \benchscore{41.10}{\phantom{\benchdelta{\uparrow}{0.00}}} & \benchscore{88.80}{\phantom{\benchdelta{\uparrow}{0.00}}} & \benchscore{98.29}{\phantom{\benchdelta{\uparrow}{0.00}}} & \benchscore{18.55}{\phantom{\benchdelta{\uparrow}{0.00}}} & \benchscore{0.540}{\phantom{\benchdelta{\uparrow}{0.000}}} & \benchscore{0.378}{\phantom{\benchdelta{\uparrow}{0.000}}} \\
    \bottomrule
    \vspace{-8mm}
  \end{tabular*}
  \endgroup
\end{table}

\subsection{Quantitative Evaluation}
\paragraph{Single-turn capability retention.}
We evaluate text-to-image generation on Qwen-Image-Bench~\citep{li2026qwenimagebench} and BizGenEval~\citep{li2026bizgeneval}, and single-turn editing on GEdit-Bench~\citep{liu2025step1xedit}, RISEBench~\citep{zhao2025envisioning}, and ImgEdit-Bench~\citep{ye2025imgedit}. Table~\ref{tab:single-turn-preservation} shows that TailSteer largely retains pretrained capabilities across all four backbones and reported steering onsets. Text-to-image performance remains close to base model, as do GEdit and ImgEdit scores. In contrast, models with FreqEdit or ConsLoRA exhibit substantially larger declines in GEdit performance. These results support TailSteer's ability to improve multi-turn robustness without broadly compromising the base model's capabilities acquired through earlier training.

\vspace{-2mm}
\paragraph{General multi-turn editing.}
While the preceding single-turn benchmarks assess the retention of pretrained generative and editing capabilities, we further evaluate their reliability across successive edits. MSE-Bench~\citep{qu2026vincie} measures instruction following and consistency over five editing turns, while EdiVal-Bench~\citep{chen2026edival} employs agentic, object-centric evaluation to assess editing reliability as visual context evolves. As shown in Table~\ref{tab:multi-turn-preservation}, TailSteer consistently improves EdiVal content consistency across all settings and achieves higher MSE-Bench success rates at selected steering onsets, although instruction-following gains remain less consistent. However, aggregate multi-turn success cannot fully distinguish inherent editing limitations from accumulated drift: for example, a failed edit in a later turn may reflect either an instruction the base model already struggles with or visual degradation inherited from earlier turns. This motivates a more controlled evaluation that directly measures cumulative visual drift alongside edit completion.

\begin{figure}[t]
  \centering
  \includegraphics[width=\linewidth]{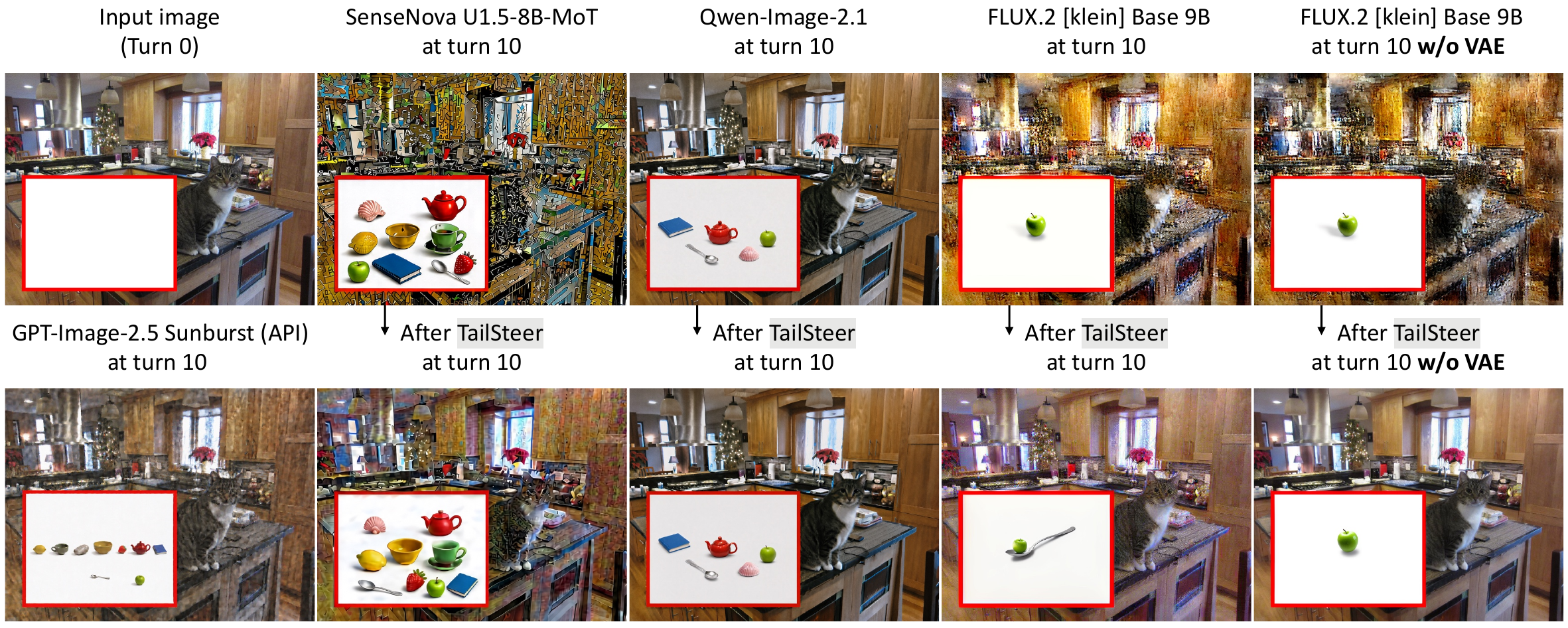}
  \vspace{-7mm}
    \caption{\textbf{Qualitative comparison after 10 editing turns.}
    TailSteer substantially reduces background drift while retaining the requested object edits across different backbones. For Klein, bypassing inter-turn VAE conversions further preserves the original background colors and details.}
    \vspace{-5mm}
  \label{fig_9_local_preserve}
\end{figure}

\paragraph{Multi-turn drift evaluation.}
To measure cumulative visual drift more directly, here we introduce \textbf{Add \& Preserve}, a controlled benchmark of 100 detail-rich natural images. For each image, we randomly place a white region enclosed by a red box and apply 10 successive instructions to add or remove objects within it (Fig.~\ref{fig_9_local_preserve}). Each edited output becomes the next input, while the surrounding content should remain unchanged. We track cumulative drift outside the box using PSNR, SSIM, and LPIPS against the original image, while Cumulative Object Accuracy (COA) evaluates whether the objects inside the box match the editing history. This design separately measures drift and edit completion, distinguishing genuine preservation gains from simply suppressing requested edits.

\begin{figure}[t]
  \centering
  \vspace{-14mm}
  \includegraphics[width=\linewidth]{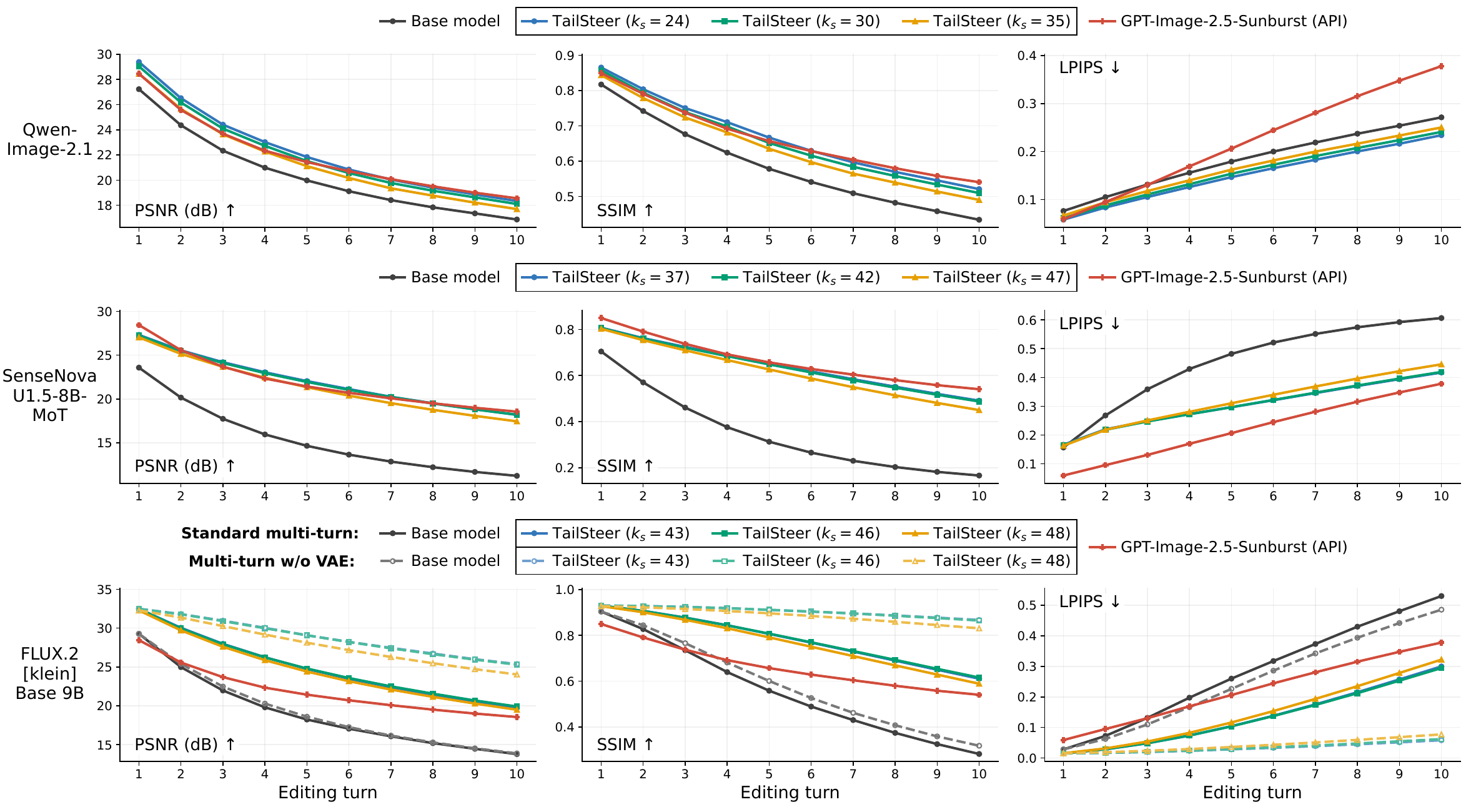}
  \vspace{-8mm}
    \caption{\textbf{Cumulative drift across turns.}
    Outside-box metrics on Add \& Preserve over 10 turns. TailSteer reduces drift across all backbones, approaching or surpassing GPT-Image-2.5-Sunburst in preservation. Dashed curves show further gains from bypassing inter-turn VAE for TailSteer-Klein.}
    \vspace{-3mm}
  \label{fig_8_preserve_plot}
\end{figure}

Despite training only on single-edit denoising rollouts, TailSteer substantially slows cumulative drift across turns (Fig.~\ref{fig_8_preserve_plot}). Notably, the best preservation occurs at earlier or intermediate steering onsets rather than the latest ones, highlighting the benefit of steering before the final few steps. Table~\ref{tab:multi-turn-preservation} confirms consistent gains across all backbones and steering onsets, with COA near base-model levels. TailSteer-Klein even surpasses GPT-Image-2.5 Sunburst on all three preservation metrics, while TailSteer-SenseNova reduces drift with nearly unchanged COA, demonstrating that TailSteer unlocks long-horizon robustness while largely retaining the base models' generative capabilities.

The turn-10 examples in Fig.~\ref{fig_9_local_preserve} further illustrate TailSteer's improved anti-drift capability, particularly for SenseNova and Klein. However, residual color shifts in VAE-based editors suggest another potential source of drift: repeated VAE reconstruction between turns. Inspired by prior work~\citep{joseph2024iterative,almog2025reedvae}, we evaluate Klein with latent feedback to bypass repeated VAE conversions between turns (note that all other experiments use standard image-feedback). In Fig.~\ref{fig_8_preserve_plot}, latent feedback alone offers limited improvement, but combining it with TailSteer achieves turn-10 PSNR comparable to standard feedback at turn 5, with less color distortion (Fig.~\ref{fig_9_local_preserve}), suggesting that latent reuse or improved VAE fidelity can further complement TailSteer's trajectory-tail steering.

\subsection{Ablation study}
\begin{figure}[t]
  \centering
  \vspace{-15mm}
  \includegraphics[width=\linewidth]{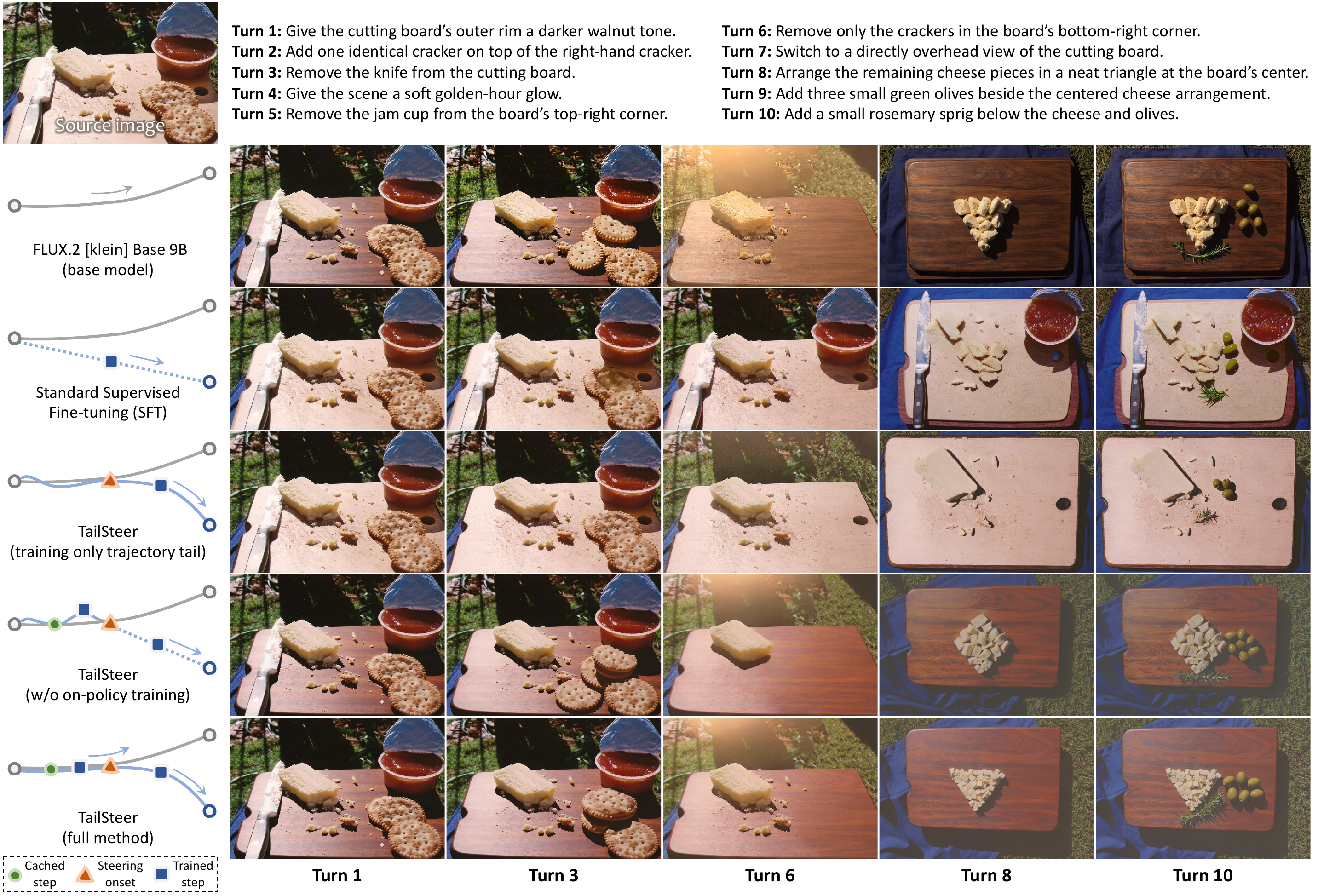}
  \vspace{-8mm}
    \caption{\textbf{Ablation of training strategies.}
    With the same training data, full TailSteer better balances preserving the base model's editing capabilities with enhancing multi-turn drift resistance.}
    \vspace{-2mm}
  \label{fig_10_ablative_study}
\end{figure}

\paragraph{Training strategy.}
We compare different training strategies using the same cached preservation data on FLUX.2 [klein] Base 9B (Table~\ref{tab:training-method-ablation}). Standard SFT reduces visual drift but weakens instruction following, as illustrated by its failure to remove objects in Fig.~\ref{fig_10_ablative_study} (turns~3,~6). Training only the trajectory tail achieves slightly higher PSNR but substantially degrades editing capability, while last-step-only training performs even worse under full-trajectory inference. Without on-policy training, preservation remains competitive, but spatial reasoning suffers: the model fails to arrange the cheese into the requested triangle at turn~8, consistent with its lower RISE score. In contrast, full TailSteer better preserves the cheese and cutting board at turn~6 while correctly executing the subsequent spatial rearrangement. Highlighting that preservation data alone is insufficient, retaining early dynamics and steering from on-policy states are key to retaining model's prior with drift resistance.

\paragraph{Data scale and trajectory types.}
We investigate how training data scale and trajectory type affect TailSteer, training each configuration for approximately one epoch (Table~\ref{tab:data-scale-ablation}). At smaller scales, local-edit trajectories provide substantially stronger preservation than identity-edit trajectories, but this gap narrows as the dataset grows. Larger datasets generally improve drift resistance, while mixing both trajectory types offers a better balance between preservation and editing capability. Our main setting combines local-edit and identity-edit training data at a $3{:}1$ ratio, highlighting the potential of scaling diverse self-generated preservation endpoints rollouts for robust multi-turn editing.

\begin{table}[!t]
  \centering
  \vspace{-3mm}
  \caption{\textbf{Training-strategy ablation.}
    All variants share same training data and full-steps sampling.}
  \label{tab:training-method-ablation}
  \begingroup
  \fontsize{6.5}{7.5}\selectfont
  \setlength{\tabcolsep}{0.8pt}
  \renewcommand{\arraystretch}{0.96}
  \setlength{\aboverulesep}{0.3ex}
  \setlength{\belowrulesep}{0.5ex}
  \definecolor{benchIncrease}{HTML}{147D73}
  \definecolor{benchDecrease}{HTML}{B65F2E}
  \colorlet{benchUp}{benchIncrease!75}
  \colorlet{benchDown}{benchDecrease!75}
  \newcommand{\benchdelta}[3][black!60]{\raisebox{-0.45ex}{{\fontsize{5}{6}\selectfont\textcolor{#1}{\makebox[0.65em][c]{$#2$}#3}}}}
  \newcommand{\benchscore}[2]{\mbox{#1\,#2}}
  \newcommand{\benchhead}[1]{{\fontsize{5.5}{6}\selectfont\mdseries #1}}
  \newcommand{\benchgroup}[1]{{\fontsize{5.5}{6}\selectfont\bfseries #1}}
  \begin{tabular*}{\linewidth}{@{\extracolsep{\fill}}l*{8}{c}@{}}
    \toprule
    Ablative training scheme & \multicolumn{4}{c}{\benchgroup{Single-turn image editing}} & \multicolumn{4}{c}{\benchgroup{Add \& Preserve}} \\
    \cmidrule(lr){2-5}\cmidrule(l){6-9}
    {\fontsize{6}{7}\selectfont (Use FLUX.2 [klein] Base 9B as backbone)} & \benchhead{GEdit-EN $\uparrow$} & \benchhead{GEdit-ZH $\uparrow$} & \benchhead{RISE Acc. (\%) $\uparrow$} & \benchhead{ImgEdit $\uparrow$}
    & \benchhead{COA (\%) $\uparrow$} & \benchhead{PSNR@10 $\uparrow$} & \benchhead{SSIM@10 $\uparrow$} & \benchhead{LPIPS@10 $\downarrow$} \\
    \midrule
    Training only last-step & \benchscore{2.24}{\benchdelta[benchDown]{\downarrow}{5.28}} & \benchscore{2.17}{\benchdelta[benchDown]{\downarrow}{5.22}} & \benchscore{\phantom{0}3.33}{\benchdelta[benchDown]{\downarrow}{10.28}} & \benchscore{2.15}{\benchdelta[benchDown]{\downarrow}{2.04}} & \benchscore{13.10}{\benchdelta[benchDown]{\downarrow}{18.73}} & \benchscore{14.93}{\benchdelta[benchDown]{\downarrow}{5.03}} & \benchscore{0.399}{\benchdelta[benchDown]{\downarrow}{0.222}} & \benchscore{0.594}{\benchdelta[benchDown]{\uparrow}{0.305}} \\
    Standard Supervised Fine-tuning (SFT) & \benchscore{5.24}{\benchdelta[benchDown]{\downarrow}{2.28}} & \benchscore{5.28}{\benchdelta[benchDown]{\downarrow}{2.11}} & \benchscore{\phantom{0}8.89}{\benchdelta[benchDown]{\downarrow}{\phantom{0}4.72}} & \benchscore{3.41}{\benchdelta[benchDown]{\downarrow}{0.77}} & \benchscore{31.07}{\benchdelta[benchDown]{\downarrow}{\phantom{0}0.76}} & \benchscore{19.77}{\benchdelta[benchDown]{\downarrow}{0.19}} & \benchscore{0.608}{\benchdelta[benchDown]{\downarrow}{0.013}} & \benchscore{0.287}{\benchdelta[benchUp]{\downarrow}{0.003}} \\
    Training only trajectory tail (from step-46) & \benchscore{6.09}{\benchdelta[benchDown]{\downarrow}{1.42}} & \benchscore{6.02}{\benchdelta[benchDown]{\downarrow}{1.37}} & \benchscore{10.28}{\benchdelta[benchDown]{\downarrow}{\phantom{0}3.33}} & \benchscore{3.92}{\benchdelta[benchDown]{\downarrow}{0.26}} & \benchscore{29.62}{\benchdelta[benchDown]{\downarrow}{\phantom{0}2.22}} & \benchscore{20.04}{\benchdelta[benchUp]{\uparrow}{0.08}} & \benchscore{0.606}{\benchdelta[benchDown]{\downarrow}{0.015}} & \benchscore{0.313}{\benchdelta[benchDown]{\uparrow}{0.023}} \\
    TailSteer w/o on-policy training & \benchscore{7.44}{\benchdelta[benchDown]{\downarrow}{0.08}} & \benchscore{7.40}{\benchdelta[benchUp]{\uparrow}{0.01}} & \benchscore{11.39}{\benchdelta[benchDown]{\downarrow}{\phantom{0}2.22}} & \benchscore{4.17}{\benchdelta[benchDown]{\downarrow}{0.01}} & \benchscore{31.94}{\benchdelta[benchUp]{\uparrow}{\phantom{0}0.11}} & \benchscore{19.81}{\benchdelta[benchDown]{\downarrow}{0.15}} & \benchscore{0.607}{\benchdelta[benchDown]{\downarrow}{0.014}} & \benchscore{0.305}{\benchdelta[benchDown]{\uparrow}{0.015}} \\
    \midrule
    TailSteer (full method, steering from step-46) & \benchscore{7.52}{\phantom{\benchdelta{\uparrow}{0.00}}} & \benchscore{7.39}{\phantom{\benchdelta{\uparrow}{0.00}}} & \benchscore{13.61}{\phantom{\benchdelta{\uparrow}{00.00}}} & \benchscore{4.19}{\phantom{\benchdelta{\uparrow}{0.00}}} & \benchscore{31.84}{\phantom{\benchdelta{\uparrow}{00.00}}} & \benchscore{19.96}{\phantom{\benchdelta{\uparrow}{0.00}}} & \benchscore{0.621}{\phantom{\benchdelta{\uparrow}{0.000}}} & \benchscore{0.290}{\phantom{\benchdelta{\uparrow}{0.000}}} \\
    \bottomrule
  \end{tabular*}
  \endgroup
\end{table}

\begin{table}[!t]
  \centering
  \vspace{-5mm}
  \caption{\textbf{Data scale and trajectory-type ablation.} 
  All variants are trained for nearly one epoch.}
  \label{tab:data-scale-ablation}
  \begingroup
  \fontsize{6.5}{7.5}\selectfont
  \setlength{\tabcolsep}{0.8pt}
  \renewcommand{\arraystretch}{0.96}
  \setlength{\aboverulesep}{0.3ex}
  \setlength{\belowrulesep}{0.5ex}
  \definecolor{benchIncrease}{HTML}{147D73}
  \definecolor{benchDecrease}{HTML}{B65F2E}
  \colorlet{benchUp}{benchIncrease!75}
  \colorlet{benchDown}{benchDecrease!75}
  \newcommand{\benchdelta}[3][black!60]{\raisebox{-0.45ex}{{\fontsize{5}{6}\selectfont\textcolor{#1}{\makebox[0.65em][c]{$#2$}#3}}}}
  \newcommand{\benchscore}[2]{\mbox{#1\,#2}}
  \newcommand{\benchhead}[1]{{\fontsize{5.5}{6}\selectfont\mdseries #1}}
  \newcommand{\benchgroup}[1]{{\fontsize{5.5}{6}\selectfont\bfseries #1}}
  \begin{tabular*}{\linewidth}{@{\extracolsep{\fill}}lc*{6}{c}@{}}
    \toprule
    Ablative training data scale & Cached
      & \multicolumn{2}{c}{\benchgroup{Single-turn image editing}} & \multicolumn{4}{c}{\benchgroup{Add \& Preserve}} \\
    \cmidrule(lr){3-4}\cmidrule(l){5-8}
    {\fontsize{6}{7}\selectfont (Use FLUX.2 [klein] Base 9B as backbone)} & Trajectories & \benchhead{GEdit-EN $\uparrow$} & \benchhead{GEdit-ZH $\uparrow$} & \benchhead{COA (\%) $\uparrow$}
      & \benchhead{PSNR@10 $\uparrow$} & \benchhead{SSIM@10 $\uparrow$} & \benchhead{LPIPS@10 $\downarrow$} \\
    \midrule
    Identity-edit data & \multirow{3}{*}{1,000} & \benchscore{7.05}{\benchdelta[benchDown]{\downarrow}{0.47}} & \benchscore{6.82}{\benchdelta[benchDown]{\downarrow}{0.56}} & \benchscore{30.27}{\benchdelta[benchDown]{\downarrow}{1.57}} & \benchscore{14.15}{\benchdelta[benchDown]{\downarrow}{5.81}} & \benchscore{0.482}{\benchdelta[benchDown]{\downarrow}{0.139}} & \benchscore{0.400}{\benchdelta[benchDown]{\uparrow}{0.110}} \\
    Local-edit data &  & \benchscore{7.40}{\benchdelta[benchDown]{\downarrow}{0.12}} & \benchscore{7.35}{\benchdelta[benchDown]{\downarrow}{0.04}} & \benchscore{30.31}{\benchdelta[benchDown]{\downarrow}{1.52}} & \benchscore{17.61}{\benchdelta[benchDown]{\downarrow}{2.35}} & \benchscore{0.557}{\benchdelta[benchDown]{\downarrow}{0.064}} & \benchscore{0.330}{\benchdelta[benchDown]{\uparrow}{0.041}} \\
    Identity-edit + Local-edit data &  & \benchscore{7.28}{\benchdelta[benchDown]{\downarrow}{0.24}} & \benchscore{7.26}{\benchdelta[benchDown]{\downarrow}{0.13}} & \benchscore{30.42}{\benchdelta[benchDown]{\downarrow}{1.41}} & \benchscore{16.88}{\benchdelta[benchDown]{\downarrow}{3.08}} & \benchscore{0.538}{\benchdelta[benchDown]{\downarrow}{0.083}} & \benchscore{0.349}{\benchdelta[benchDown]{\uparrow}{0.059}} \\
    \midrule
    Identity-edit data & \multirow{3}{*}{5,000} & \benchscore{6.92}{\benchdelta[benchDown]{\downarrow}{0.59}} & \benchscore{6.97}{\benchdelta[benchDown]{\downarrow}{0.42}} & \benchscore{31.23}{\benchdelta[benchDown]{\downarrow}{0.61}} & \benchscore{16.61}{\benchdelta[benchDown]{\downarrow}{3.35}} & \benchscore{0.540}{\benchdelta[benchDown]{\downarrow}{0.081}} & \benchscore{0.353}{\benchdelta[benchDown]{\uparrow}{0.063}} \\
    Local-edit data &  & \benchscore{7.26}{\benchdelta[benchDown]{\downarrow}{0.26}} & \benchscore{7.27}{\benchdelta[benchDown]{\downarrow}{0.12}} & \benchscore{31.11}{\benchdelta[benchDown]{\downarrow}{0.72}} & \benchscore{18.85}{\benchdelta[benchDown]{\downarrow}{1.11}} & \benchscore{0.614}{\benchdelta[benchDown]{\downarrow}{0.007}} & \benchscore{0.278}{\benchdelta[benchUp]{\downarrow}{0.012}} \\
    Identity-edit + Local-edit data &  & \benchscore{7.35}{\benchdelta[benchDown]{\downarrow}{0.17}} & \benchscore{7.24}{\benchdelta[benchDown]{\downarrow}{0.15}} & \benchscore{31.19}{\benchdelta[benchDown]{\downarrow}{0.64}} & \benchscore{19.01}{\benchdelta[benchDown]{\downarrow}{0.95}} & \benchscore{0.619}{\benchdelta[benchDown]{\downarrow}{0.003}} & \benchscore{0.270}{\benchdelta[benchUp]{\downarrow}{0.019}} \\
    \midrule
    Identity-edit data & \multirow{3}{*}{10,000} & \benchscore{7.05}{\benchdelta[benchDown]{\downarrow}{0.47}} & \benchscore{7.05}{\benchdelta[benchDown]{\downarrow}{0.33}} & \benchscore{30.63}{\benchdelta[benchDown]{\downarrow}{1.21}} & \benchscore{18.00}{\benchdelta[benchDown]{\downarrow}{1.96}} & \benchscore{0.559}{\benchdelta[benchDown]{\downarrow}{0.062}} & \benchscore{0.337}{\benchdelta[benchDown]{\uparrow}{0.047}} \\
    Local-edit data &  & \benchscore{7.32}{\benchdelta[benchDown]{\downarrow}{0.20}} & \benchscore{7.30}{\benchdelta[benchDown]{\downarrow}{0.08}} & \benchscore{30.98}{\benchdelta[benchDown]{\downarrow}{0.85}} & \benchscore{18.79}{\benchdelta[benchDown]{\downarrow}{1.17}} & \benchscore{0.630}{\benchdelta[benchUp]{\uparrow}{0.009}} & \benchscore{0.268}{\benchdelta[benchUp]{\downarrow}{0.022}} \\
    Identity-edit + Local-edit data &  & \benchscore{7.30}{\benchdelta[benchDown]{\downarrow}{0.22}} & \benchscore{7.17}{\benchdelta[benchDown]{\downarrow}{0.22}} & \benchscore{31.96}{\benchdelta[benchUp]{\uparrow}{0.12}} & \benchscore{18.94}{\benchdelta[benchDown]{\downarrow}{1.02}} & \benchscore{0.616}{\benchdelta[benchDown]{\downarrow}{0.005}} & \benchscore{0.280}{\benchdelta[benchUp]{\downarrow}{0.010}} \\
    \midrule
    Full data (main setting) & 38,540 & \benchscore{7.52}{\phantom{\benchdelta{\uparrow}{0.00}}} & \benchscore{7.39}{\phantom{\benchdelta{\uparrow}{0.00}}} & \benchscore{31.84}{\phantom{\benchdelta{\uparrow}{0.00}}} & \benchscore{19.96}{\phantom{\benchdelta{\uparrow}{0.00}}} & \benchscore{0.621}{\phantom{\benchdelta{\uparrow}{0.000}}} & \benchscore{0.290}{\phantom{\benchdelta{\uparrow}{0.000}}} \\
    \bottomrule
  \end{tabular*}
  \vspace{-5mm}
  \endgroup
\end{table}

\vspace{-1mm}
\section{Conclusion}
\label{sec:conclusion}
\vspace{-1mm}
We presented \textbf{TailSteer}, a final-stage training framework that unlocks the multi-turn editing potential of foundation models without relearning their existing capabilities. By concentrating preservation corrections near the denoising trajectory tail while retaining earlier dynamics, TailSteer substantially reduces cumulative drift across successive edits. Extensive experiments across FLUX, SenseNova, and Qwen demonstrate stronger long-horizon preservation while largely retaining the base models' generative capabilities. Our findings suggest that robust multi-turn editing need not be learned anew, but can be unlocked by refining how foundation models complete each edit.

\newpage
\bibliography{refs}

@misc{openai2025gptimage,
  author       = {{OpenAI}},
  title        = {Introducing Our Latest Image Generation Model in the {API}},
  year         = {2025},
  month        = apr,
  howpublished = {OpenAI product release},
  url          = {https://openai.com/index/image-generation-api/},
}

@misc{openai2025gptimage15,
  author       = {{OpenAI}},
  title        = {The New {ChatGPT Images} Is Here},
  year         = {2025},
  month        = dec,
  howpublished = {OpenAI product release},
  url          = {https://openai.com/index/new-chatgpt-images-is-here/},
}

@article{labs2025flux1kontextflowmatching,
  author       = {{Black Forest Labs} and Batifol, Stephen and Blattmann, Andreas and Boesel,
                  Frederic and Consul, Saksham and Diagne, Cyril and Dockhorn, Tim and English,
                  Jack and others},
  title        = {{FLUX.1 Kontext}: Flow Matching for In-Context Image Generation and Editing in
                  Latent Space},
  journal      = {arXiv preprint arXiv:2506.15742},
  year         = {2025},
}

@misc{flux22025,
  author       = {{Black Forest Labs}},
  title        = {{FLUX.2}: Frontier Visual Intelligence},
  year         = {2025},
  month        = nov,
  howpublished = {Technical blog post},
  url          = {https://bfl.ai/blog/flux-2},
}

@misc{google2025gemini25flashimage,
  author       = {{Google}},
  title        = {Image Editing in {Gemini} Just Got a Major Upgrade},
  year         = {2025},
  month        = aug,
  howpublished = {Google product announcement},
  url          = {https://blog.google/products-and-platforms/products/gemini/updated-image-editing-model/},
}

@misc{google2026nanobanana2,
  author       = {{Google}},
  title        = {{Nano Banana 2}: Combining Pro Capabilities with Lightning-Fast Speed},
  year         = {2026},
  month        = feb,
  howpublished = {Google product announcement},
  url          = {https://blog.google/innovation-and-ai/technology/ai/nano-banana-2/},
}

@misc{openai2026images25,
  author       = {{OpenAI}},
  title        = {Introducing {ChatGPT Images 2.5}},
  year         = {2026},
  month        = sep,
  howpublished = {OpenAI product release},
  url          = {https://openai.com/index/introducing-chatgpt-images-2-5/},
}

@misc{reve2026reve21,
  author       = {{Reve}},
  title        = {Launching {Reve 2.1}},
  year         = {2026},
  month        = jul,
  howpublished = {Reve product announcement},
  url          = {https://blog.reve.com/posts/launching-reve-2.1/},
}

@misc{meta2026museimage,
  author       = {{Meta}},
  title        = {Introducing {Muse Image} and {Muse Video}},
  year         = {2026},
  month        = jul,
  howpublished = {Meta AI research announcement},
  url          = {https://ai.meta.com/blog/introducing-muse-image-muse-video-msl/},
}

@inproceedings{couairon2023diffedit,
  author       = {Couairon, Guillaume and Verbeek, Jakob and Schwenk, Holger and Cord, Matthieu},
  title        = {{DiffEdit}: Diffusion-Based Semantic Image Editing with Mask Guidance},
  booktitle    = {International Conference on Learning Representations},
  year         = {2023},
}

@misc{dx8152flux2consistency,
  author       = {{dx8152}},
  title        = {{Flux2-Klein-9B-Consistency}},
  year         = {2026},
  howpublished = {Hugging Face model repository},
  note         = {Version 2, released April 17, 2026},
  url          = {https://huggingface.co/dx8152/Flux2-Klein-9B-Consistency},
}

@inproceedings{liao2026freqedit,
  author       = {Liao, Yucheng and Liang, Jiajun and Cui, Kaiqian and Zhao, Baoquan and Xie,
                  Haoran and Liu, Wei and Li, Qing and Mao, Xudong},
  title        = {{FreqEdit}: Preserving High-Frequency Features for Robust Multi-Turn Image
                  Editing},
  booktitle    = {Proceedings of the IEEE/CVF Conference on Computer Vision and Pattern
                  Recognition (CVPR)},
  pages        = {43525--43535},
  year         = {2026},
}

@article{wang2026ditdrift,
  author       = {Wang, Xiaoce and Zhou, Sifan and Wang, Kaifei and Xu, Leli and Qiu, Xuerui and
                  He, Tao and Li, Ming},
  title        = {Why Do {DiT} Editors Drift? Plug-and-Play Low Frequency Alignment in {VAE}
                  Latent Space},
  journal      = {arXiv preprint arXiv:2605.08250},
  year         = {2026},
}

@inproceedings{yao2025reconstruction,
  author       = {Yao, Jingfeng and Yang, Bin and Wang, Xinggang},
  title        = {Reconstruction vs. Generation: Taming Optimization Dilemma in Latent Diffusion
                  Models},
  booktitle    = {Proceedings of the IEEE/CVF Conference on Computer Vision and Pattern
                  Recognition (CVPR)},
  pages        = {15703--15712},
  year         = {2025},
}

@misc{dieleman2025latents,
  author       = {Dieleman, Sander},
  title        = {Generative Modelling in Latent Space},
  year         = {2025},
  month        = apr,
  howpublished = {Technical blog post},
  url          = {https://sander.ai/2025/04/15/latents.html},
}

@misc{flux2latent2025,
  author       = {{Black Forest Labs}},
  title        = {{FLUX.2}: Analyzing and Enhancing the Latent Space of {FLUX}---Representation
                  Comparison},
  year         = {2025},
  month        = nov,
  howpublished = {Technical report},
  url          = {https://bfl.ai/research/representation-comparison},
}

@article{diao2026sensenovau1,
  title={Sensenova-u1: Unifying multimodal understanding and generation with neo-unify architecture},
  author={Diao, Haiwen and Wu, Penghao and Deng, Hanming and Wang, Jiahao and Bai, Shihao and Wu, Silei and Fan, Weichen and Ye, Wenjie and Tong, Wenwen and Fan, Xiangyu and others},
  journal={arXiv preprint arXiv:2605.12500},
  year={2026}
}

@inproceedings{zhu2025kvedit,
  author       = {Zhu, Tianrui and Zhang, Shiyi and Shao, Jiawei and Tang, Yansong},
  title        = {{KV-Edit}: Training-Free Image Editing for Precise Background Preservation},
  booktitle    = {Proceedings of the IEEE/CVF International Conference on Computer Vision (ICCV)},
  pages        = {16607--16617},
  year         = {2025},
}

@inproceedings{kim2025layerwisememory,
  author       = {Kim, Daneul and Lee, Jaeah and Park, Jaesik},
  title        = {Improving Editability in Image Generation with Layer-wise Memory},
  booktitle    = {Proceedings of the IEEE/CVF Conference on Computer Vision and Pattern
                  Recognition (CVPR)},
  pages        = {7889--7898},
  year         = {2025},
}

@inproceedings{qian2024boosting,
  author       = {Qian, Yurui and Cai, Qi and Pan, Yingwei and Li, Yehao and Yao, Ting and Sun,
                  Qibin and Mei, Tao},
  title        = {Boosting Diffusion Models with Moving Average Sampling in Frequency Domain},
  booktitle    = {Proceedings of the IEEE/CVF Conference on Computer Vision and Pattern
                  Recognition (CVPR)},
  pages        = {8911--8920},
  year         = {2024},
}

@inproceedings{ning2023inputperturbation,
  author       = {Ning, Mang and Sangineto, Enver and Porrello, Angelo and Calderara, Simone and
                  Cucchiara, Rita},
  title        = {Input Perturbation Reduces Exposure Bias in Diffusion Models},
  booktitle    = {Proceedings of the 40th International Conference on Machine Learning},
  volume       = {202},
  pages        = {26245--26265},
  series       = {Proceedings of Machine Learning Research},
  publisher    = {PMLR},
  year         = {2023},
}

@inproceedings{ning2024elucidating,
  author       = {Ning, Mang and Li, Mingxiao and Su, Jianlin and Salah, Albert Ali and Onal
                  Ertugrul, Itir},
  title        = {Elucidating the Exposure Bias in Diffusion Models},
  booktitle    = {International Conference on Learning Representations},
  pages        = {15167--15189},
  year         = {2024},
}

@inproceedings{li2024timeshift,
  author       = {Li, Mingxiao and Qu, Tingyu and Yao, Ruicong and Sun, Wei and Moens,
                  Marie-Francine},
  title        = {Alleviating Exposure Bias in Diffusion Models through Sampling with Shifted Time
                  Steps},
  booktitle    = {International Conference on Learning Representations},
  pages        = {16816--16838},
  year         = {2024},
}

@article{diao2026sensenova,
  title={Sensenova-u1. 5: Towards native unified visual intelligence},
  author={Diao, Haiwen and Wang, Jiahao and Ding, Chenjing and Deng, Hanming and Chen, Jiangnan and Zhang, Ruixi and Wang, Ruohui and Tong, Wenwen and Fan, Xiangyu and Wang, Yubo and others},
  journal={arXiv preprint arXiv:2609.11929},
  year={2026}
}

@misc{qwen2026image21,
  author       = {{Qwen Team}},
  title        = {{Qwen-Image-2.1}},
  year         = {2026},
  howpublished = {Hugging Face model card},
  url          = {https://huggingface.co/Qwen/Qwen-Image-2.1},
}

@inproceedings{laban2026llms,
  author       = {Laban, Philippe and Hayashi, Hiroaki and Zhou, Yingbo and Neville, Jennifer},
  title        = {{LLMs} Get Lost In Multi-Turn Conversation},
  booktitle    = {International Conference on Learning Representations},
  pages        = {54738--54778},
  year         = {2026},
}

@inproceedings{joseph2024iterative,
  author       = {Joseph, K. J. and Udhayanan, Prateksha and Shukla, Tripti and Agarwal, Aishwarya
                  and Karanam, Srikrishna and Goswami, Koustava and Srinivasan, Balaji Vasan},
  title        = {Iterative Multi-Granular Image Editing Using Diffusion Models},
  booktitle    = {Proceedings of the IEEE/CVF Winter Conference on Applications of Computer Vision
                  (WACV)},
  pages        = {8107--8116},
  year         = {2024},
}

@article{almog2025reedvae,
  author       = {Almog, Gal and Shamir, Ariel and Fried, Ohad},
  title        = {{REED-VAE}: {RE-Encode} Decode Training for Iterative Image Editing with
                  Diffusion Models},
  journal      = {Computer Graphics Forum},
  volume       = {44},
  number       = {2},
  pages        = {e70020},
  year         = {2025},
  doi          = {10.1111/cgf.70020},
}

@inproceedings{li2026back,
  author       = {Li, Tianhong and He, Kaiming},
  title        = {Back to Basics: Let Denoising Generative Models Denoise},
  booktitle    = {Proceedings of the IEEE/CVF Conference on Computer Vision and Pattern
                  Recognition (CVPR)},
  pages        = {36115--36125},
  year         = {2026},
}

@inproceedings{yu2026pixeldit2,
  author       = {Yongsheng Yu and Wei Xiong and Yichen Sheng and Shiqiu Liu and Jiebo Luo},
  title        = {{PixelDiT2}: Representation-Grounded Pixel Diffusion Transformers},
  booktitle    = {Conference on Neural Information Processing Systems (NeurIPS)},
  year         = {2026},
}

@inproceedings{avrahami2022blended,
  author       = {Avrahami, Omri and Lischinski, Dani and Fried, Ohad},
  title        = {Blended Diffusion for Text-Driven Editing of Natural Images},
  booktitle    = {Proceedings of the IEEE/CVF Conference on Computer Vision and Pattern
                  Recognition (CVPR)},
  pages        = {18208--18218},
  year         = {2022},
}

@inproceedings{yin2025consistedit,
  author       = {Yin, Zixin and Chen, Ling-Hao and Ni, Lionel and Dai, Xili},
  title        = {{ConsistEdit}: Highly Consistent and Precise Training-free Visual Editing},
  booktitle    = {Proceedings of the SIGGRAPH Asia 2025 Conference Papers},
  pages        = {1--11},
  publisher    = {Association for Computing Machinery},
  year         = {2025},
  doi          = {10.1145/3757377.3763909},
}

@inproceedings{zhou2025multiturn,
  author       = {Zhou, Zijun and Deng, Yingying and He, Xiangyu and Dong, Weiming and Tang, Fan},
  title        = {Multi-turn Consistent Image Editing},
  booktitle    = {Proceedings of the IEEE/CVF International Conference on Computer Vision (ICCV)},
  pages        = {15792--15801},
  year         = {2025},
}

@inproceedings{jiang2026flowdc,
  author       = {Jiang, Yilei and Wang, Zhen and Wang, Yanghao and Yu, Jun and Zhuang, Yueting
                  and Xiao, Jun and Chen, Long},
  title        = {{FlowDC}: Flow-Based Decoupling-Decay for Complex Image Editing},
  booktitle    = {Proceedings of the IEEE/CVF Conference on Computer Vision and Pattern
                  Recognition (CVPR)},
  pages        = {25757--25766},
  year         = {2026},
}

@inproceedings{kim2025reflex,
  author       = {Kim, Jimyeong and Park, Jungwon and Song, Yeji and Kwak, Nojun and Rhee,
                  Wonjong},
  title        = {{ReFlex}: Text-Guided Editing of Real Images in Rectified Flow via Mid-Step
                  Feature Extraction and Attention Adaptation},
  booktitle    = {Proceedings of the IEEE/CVF International Conference on Computer Vision (ICCV)},
  pages        = {15939--15948},
  year         = {2025},
}

@inproceedings{yu2025contextmemory,
  author       = {Yu, Jiwen and Bai, Jianhong and Qin, Yiran and Liu, Quande and Wang, Xintao and
                  Wan, Pengfei and Zhang, Di and Liu, Xihui},
  title        = {Context as Memory: Scene-Consistent Interactive Long Video Generation with
                  Memory Retrieval},
  booktitle    = {Proceedings of the SIGGRAPH Asia 2025 Conference Papers},
  pages        = {1--11},
  publisher    = {Association for Computing Machinery},
  year         = {2025},
  doi          = {10.1145/3757377.3763833},
}

@inproceedings{dalal2025oneminute,
  author       = {Dalal, Karan and Koceja, Daniel and Xu, Jiarui and Zhao, Yue and Han, Shihao and
                  Cheung, Ka Chun and Kautz, Jan and Choi, Yejin and Sun, Yu and Wang, Xiaolong},
  title        = {One-Minute Video Generation with Test-Time Training},
  booktitle    = {Proceedings of the IEEE/CVF Conference on Computer Vision and Pattern
                  Recognition (CVPR)},
  pages        = {17702--17711},
  year         = {2025},
}

@inproceedings{qu2026vincie,
  author       = {Qu, Leigang and Cheng, Feng and Yang, Ziyan and Zhao, Qi and Lin, Shanchuan and
                  Shi, Yichun and Li, Yicong and Wang, Wenjie and Chua, Tat-Seng and Jiang, Lu},
  title        = {{VINCIE}: Unlocking In-context Image Editing from Video},
  booktitle    = {International Conference on Learning Representations},
  pages        = {114873--114918},
  year         = {2026},
}

@article{huang2026mteditflow,
  author       = {Huang, Jiahui and Zhang, Yasi and Chen, Tianyu and Wang, Shu and Xie, Jianwen
                  and Leong, Oscar and Zhou, Mingyuan and Wang, Nanzhu and Wu, Ying Nian},
  title        = {{MT-EditFlow}: Reinforcement Learning for Multi-Turn Image Editing with Flow
                  Matching},
  journal      = {arXiv preprint arXiv:2606.01985},
  year         = {2026},
}

@article{kuznetsova2020openimages,
  author       = {Kuznetsova, Alina and Rom, Hassan and Alldrin, Neil and Uijlings, Jasper and
                  Krasin, Ivan and Pont-Tuset, Jordi and Kamali, Shahab and Popov, Stefan and others},
  title        = {The {Open Images} Dataset {V4}: Unified Image Classification, Object Detection,
                  and Visual Relationship Detection at Scale},
  journal      = {International Journal of Computer Vision},
  volume       = {128},
  number       = {7},
  pages        = {1956--1981},
  year         = {2020},
  doi          = {10.1007/s11263-020-01316-z},
}

@article{liu2025step1xedit,
  author       = {Liu, Shiyu and Han, Yucheng and Xing, Peng and Yin, Fukun and Wang, Rui and
                  Cheng, Wei and Liao, Jiaqi and Wang, Yingming and Fu, Honghao and Han, Chunrui
                  and Li, Guopeng and Peng, Yuang and Sun, Quan and Wu, Jingwei and Cai, Yan and others},
  title        = {{Step1X-Edit}: A Practical Framework for General Image Editing},
  journal      = {arXiv preprint arXiv:2504.17761},
  year         = {2025},
}

@inproceedings{zhao2025envisioning,
  author       = {Zhao, Xiangyu and Zhang, Peiyuan and Tang, Kexian and Zhu, Xiaorong and Li, Hao
                  and Chai, Wenhao and Zhang, Zicheng and Xia, Renqiu and Zhai, Guangtao and others},
  title        = {Envisioning Beyond the Pixels: Benchmarking Reasoning-Informed Visual Editing},
  booktitle    = {Advances in Neural Information Processing Systems},
  volume       = {38},
  publisher    = {Curran Associates, Inc.},
  year         = {2025},
  doi          = {10.52202/085713-5791},
}

@inproceedings{ye2025imgedit,
  author       = {Ye, Yang and He, Xianyi and Li, Zongjian and Lin, Bin and Yuan, Shenghai and
                  Yan, Zhiyuan and Hou, Bohan and Yuan, Li},
  title        = {{ImgEdit}: A Unified Image Editing Dataset and Benchmark},
  booktitle    = {Advances in Neural Information Processing Systems},
  volume       = {38},
  publisher    = {Curran Associates, Inc.},
  year         = {2025},
  doi          = {10.52202/085713-4387},
}

@misc{flux2klein2026,
  author       = {{Black Forest Labs}},
  title        = {{FLUX.2} [klein] {Base 9B}},
  year         = {2026},
  howpublished = {Hugging Face model card},
  url          = {https://huggingface.co/black-forest-labs/FLUX.2-klein-base-9B},
}

@inproceedings{li2026qwenimagebench,
  title={Qwen-image-bench: From generation to creation in text-to-image evaluation},
  author={Li, Niantong and Hu, Guangzheng and Qiao, Weixu and Ba, Ying and Hong, Qichen and Shen, Shijun and Wang, Jinlin and Zhou, Fan and Kang, Jianye and Shang, Xin and others},
  booktitle={Proceedings of the 2026 Conference on Empirical Methods in Natural Language Processing},
  pages={22760--22787},
  year={2026}
}

@article{li2026bizgeneval,
  author       = {Li, Yan and Zeng, Zezi and Zhou, Ziwei and Gao, Xin and Tian, Muzhao and Yang,
                  Yifan and Cheng, Mingxi and Dai, Qi and Yang, Yuqing and Qiu, Lili and Wang,
                  Zhendong and Yang, Zhengyuan and Yang, Xue and Wang, Lijuan and Li, Ji and Luo,
                  Chong},
  title        = {{BizGenEval}: A Systematic Benchmark for Commercial Visual Content Generation},
  journal      = {arXiv preprint arXiv:2603.25732},
  year         = {2026},
}

@inproceedings{chen2026edival,
  author       = {Chen, Tianyu and Zhang, Yasi and Zhang, Zhi and Yu, Peiyu and Wang, Shu and
                  Wang, Zhendong and Lin, Kevin and Wang, Xiaofei and Yang, Zhengyuan and Li,
                  Linjie and others},
  title        = {{EdiVal-Agent}: An Object-Centric Framework for Automated, Fine-Grained
                  Evaluation of Multi-Turn Editing},
  booktitle    = {International Conference on Learning Representations},
  year         = {2026},
}

@article{aubin2026monet,
  title={MONET: A Massive, Open, Non-redundant and Enriched Text-to-image dataset},
  author={Aubin, Benjamin and Quintana, Gonzalo I{\~n}aki and Tasar, Onur and Sreetharan, Sanjeev and Czerwinska, Urszula and Henry, Damien and Chadebec, Cl{\'e}ment},
  journal={arXiv preprint arXiv:2605.21272},
  year={2026}
}

@article{otsu1979threshold,
  author  = {Otsu, Nobuyuki},
  title   = {A Threshold Selection Method from Gray-Level Histograms},
  journal = {IEEE Transactions on Systems, Man, and Cybernetics},
  volume  = {9},
  number  = {1},
  pages   = {62--66},
  year    = {1979},
  doi     = {10.1109/TSMC.1979.4310076},
}
\bibliographystyle{iclr2027_conference}

\clearpage
\appendix
\graphicspath{{appendix/figures/}}
\setcounter{figure}{0}
\setcounter{table}{0}
\setcounter{equation}{0}
\renewcommand{\thefigure}{S\arabic{figure}}
\renewcommand{\thetable}{S\arabic{table}}
\renewcommand{\theequation}{S\arabic{equation}}
\renewcommand{\theHfigure}{app.\arabic{figure}}
\renewcommand{\theHtable}{app.\arabic{table}}
\renewcommand{\theHequation}{app.\arabic{equation}}
\crefname{figure}{Fig.}{Figs.}
\Crefname{figure}{Figure}{Figures}
\crefname{table}{Tab.}{Tabs.}
\Crefname{table}{Table}{Tables}
\crefname{equation}{Eq.}{Eqs.}
\crefname{section}{Sec.}{Secs.}
\crefname{subsection}{Sec.}{Secs.}
\crefname{appendix}{Sec.}{Secs.}
\hypersetup{hidelinks}

\pdfbookmark[0]{Appendix}{tailsteer-appendix}
\section*{Appendix}
\subsection*{Overview}
\begingroup
\fontsize{10.5}{13.3}\selectfont
\setlength{\parindent}{0pt}
\setlength{\parskip}{0.53\baselineskip}
\newcommand{\overviewsection}[1]{%
  \hyperref[#1]{\textbf{\ref*{#1}.\enspace\nameref*{#1}}\,\textcolor{blue!55!black}{\(\scriptstyle\nearrow\)}}%
  \enspace\ignorespaces}
\newcommand{\overviewnumber}[1]{\textcolor{blue!55!black}{\labelcref{#1}}}
\newcommand{\overviewref}[1]{\hyperref[#1]{\textcolor{blue!55!black}{\cref*{#1}}}}

% \overviewsection{app:detailed-related-work}
% Small background changes can accumulate when each edit starts from the previous output.
% We review how image representations, inference-time guidance, visual memory, and multi-turn
% supervision address this drift. TailSteer targets the final denoising steps of a single edit.

\overviewsection{app:dataset-construction}
We describe how to construct TailSteer's training data, from selecting source
images (\overviewref{fig:supp-source-categories}) and generating editing instructions
(\overviewref{fig:supp-annotation-prompt}) to applying automated locality and identity
checks and building preservation masks (\overviewref{fig:supp-screening-flow}).
Rejected examples illustrate the purpose of these checks
(\overviewref{fig:supp-screening-examples}).

\overviewsection{app:training-details}
For readers looking to reproduce TailSteer, we provide the training configurations
for each base model (\overviewref{tab:supp-training-configs}) and the full training
algorithm (\hyperref[alg:supp-tailsteer]{Alg.~\ref*{alg:supp-tailsteer}}).
We explain how on-policy rollouts, step sampling, and velocity blending work
together to preserve earlier dynamics while steering the tail.

\overviewsection{app:training-analysis}
For readers adapting TailSteer to their own base models, we share practical
insights on which parameters to update (\overviewref{fig:supp-scope-updates}),
how to choose a learning rate (\overviewref{fig:supp-qwen-rate}), and how to
configure step sampling (\overviewref{fig:supp-klein-step-sampling}) and velocity
blending (\overviewref{fig:supp-odds-tail-path}).
We also discuss numerical precision and explore which parts of the base models
change during training and how much they change (\overviewref{tab:supp-parameter-changes}).

\overviewsection{app:evaluation-protocols}
We detail inference settings and scoring protocols and explain why editing
capability and drift resistance should be evaluated separately.
Examples from MSE-Bench (\overviewref{fig:supp-mse-history}) show how earlier
drift may make later edits easier or failures harder to detect.
We introduce the design and metrics of Add \& Preserve, which measures
cumulative drift and object retention separately, with the evaluation prompt
in~\overviewref{fig:supp-coa-prompt}.

\overviewsection{app:quantitative-results}
For a closer look at the main paper's quantitative results, we provide detailed
scores for two text-to-image and three single-turn editing benchmarks
(\overviewref{tab:supp-qib-breakdown}, \overviewnumber{tab:supp-bizgen-breakdown},
\overviewnumber{tab:supp-gedit-breakdown} and~\overviewnumber{tab:supp-rise-breakdown}).

\overviewsection{app:qualitative-results}
To see how TailSteer retains the base models' T2I generative priors, readers can
compare the matched text-to-image samples
(Figs.~\overviewnumber{fig:supp-t2i-klein}, \overviewnumber{fig:supp-t2i-sensenova}, and~\overviewnumber{fig:supp-t2i-qwen}).
Our single-turn comparisons show how TailSteer preserves challenging non-local
editing capabilities while improving pixel-level fidelity in local edits,
with RGB error maps and masked PSNR
(Figs.~\overviewnumber{fig:supp-gedit-klein}, \overviewnumber{fig:supp-gedit-sensenova}, and~\overviewnumber{fig:supp-gedit-qwen}).
For single-input multi-turn editing, we organize the Klein comparisons into
three types: local edits
(Figs.~\overviewnumber{fig:supp-single-input-l1}, \overviewnumber{fig:supp-single-input-l2}, and~\overviewnumber{fig:supp-single-input-l3}),
identity-preserving edits
(Figs.~\overviewnumber{fig:supp-single-input-i2} and~\overviewnumber{fig:supp-single-input-i3}),
and general edits
(Figs.~\overviewnumber{fig:supp-single-input-m1}, \overviewnumber{fig:supp-single-input-m2}, and~\overviewnumber{fig:supp-single-input-m3}).
We show how TailSteer reduces drift over ten turns with SenseNova and
twenty turns with Qwen
(Figs.~\overviewnumber{fig:supp-single-input-sensenova} and~\overviewnumber{fig:supp-single-input-qwen}).
For readers interested in editing with visual references, our
\hyperref[app:qualitative-multi-input]{multi-input experiments (Sec.~\ref*{app:qualitative-multi-input})}
cover multi-object compositing and skeleton-guided editing
(Figs.~\overviewnumber{fig:supp-multi-input-klein}, \overviewnumber{fig:supp-multi-input-sensenova}, and~\overviewnumber{fig:supp-multi-input-qwen}).
Finally, seven \hyperref[app:simulated-creative-workflows]{simulated creative workflows (Sec.~\ref*{app:simulated-creative-workflows})}
show how TailSteer supports iterative creation with recurring people and products
over ten or twenty turns
(Figs.~\overviewnumber{fig:supp-creative-workflows-klein}, \overviewnumber{fig:supp-creative-workflows-sensenova}, and~\overviewnumber{fig:supp-creative-workflows-qwen}).

\overviewsection{app:limitations}
We discuss why TailSteer's multi-turn capabilities still depend on the base model
(\overviewref{fig:supp-identity-limitations}) and outline two directions toward
truly persistent editing: improving representation fidelity between turns and
learning stable tail corrections from the model's own multi-turn trajectories.
\par\endgroup

\clearpage
\section{Training Dataset Construction}
\label{app:dataset-construction}

\subsection{Source Images and Instruction Generation}
\label{app:data-sources}
\begin{figure}[!t]
  \centering
  \includegraphics[width=\linewidth]{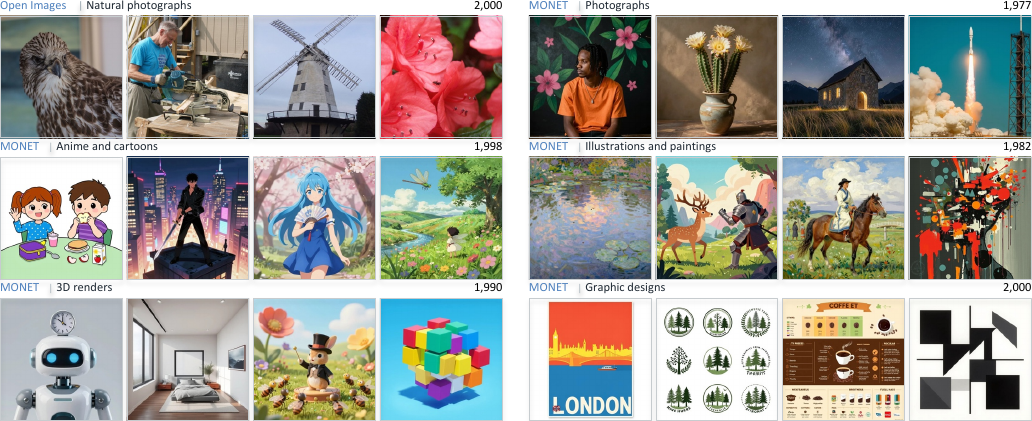}
  \caption{\textbf{Source images used for instruction generation.} Counts give the number of source images in each group before filtering.}
  \label{fig:supp-source-categories}
\end{figure}

We begin with 2,000 natural photographs from Open Images~\citep{kuznetsova2020openimages} and 9,947 images from MONET~\citep{aubin2026monet}. MONET subset spans five visual categories: photographs, anime and cartoons, illustrations and paintings, 3D renders, and graphic designs.~\cref{fig:supp-source-categories} shows four examples from each category after center cropping.

\begin{figure}[htbp]
  \centering
  \fbox{\begin{minipage}{\dimexpr\linewidth-2\fboxsep-2\fboxrule\relax}
    \small\raggedright
    \textbf{Visual review}\\
    First identify a clearly referable main subject of sufficient visible size in THIS crop. Then assess that subject's suitability for local editing BEFORE inventing any instructions.\par
    \smallskip\textbf{Instruction generation}\\
    Produce exactly three independent English instructions, each applied to this SAME original image, NOT a sequence. \emph{[\ldots]} Each instruction expresses one main local change, clearly names its target, and may preserve everything else.\par
    \smallskip\textbf{Response format}\\
    Return ONLY a JSON object with accepted (boolean), rejection\_category (null if accepted), and edits (array).
  \end{minipage}}
  \caption{\textbf{Excerpts from the annotation prompt.} Bold headings separate the prompt components for readability.}
  \label{fig:supp-annotation-prompt}
\end{figure}

\begin{figure}[t]
  \centering
  \includegraphics[width=\textwidth]{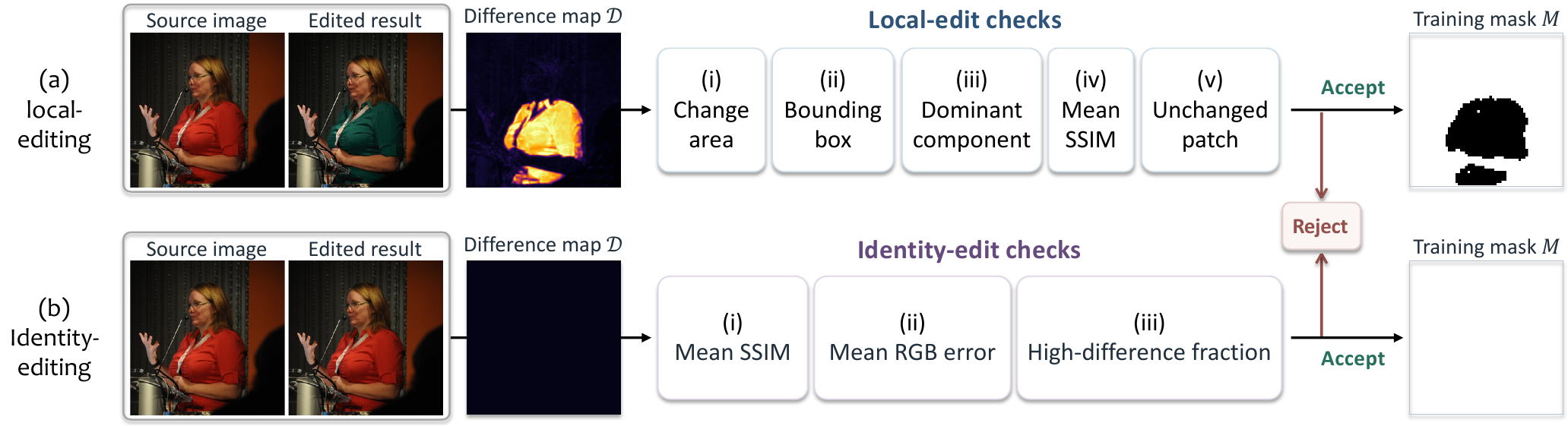}
    \caption{\textbf{Automated results screening and mask construction.} Local and identity edits undergo separate checks, detailed in~\cref{app:data-checks}. Accepted results produce training masks, where white regions are preserved from the source and black regions retain the edited result.}
  \label{fig:supp-screening-flow}
\end{figure}

To generate editing instructions, we use \texttt{gpt-6-sol} via its API to annotate each $1024\times1024$ center crop. Given the image and an annotation prompt (\cref{fig:supp-annotation-prompt}), the model first checks whether the crop contains a clearly identifiable subject suitable for local editing. If accepted, it generates three independent English instructions, each requesting a single change to the original image. We additionally include an identity instruction, \emph{Keep the image exactly unchanged}, to provide examples of full-image preservation. Applying all four instructions independently to each source image with each frozen backbone yields 35,841 local-edit requests and 11,947 identity requests per base model.

\subsection{Automated Identity and Locality Checks}
\label{app:data-checks}

Not every editing results from a frozen base model is suitable for preservation training, and manually screening the outputs of each base model would be costly at the scale of our experiments. We therefore adopt conservative automated checks with stringent thresholds to approximate careful human screening, prioritizing reliable samples over higher acceptance rates. Although manual review might further improve data quality, this automated approach has already yielded strong results in our all experiments.

To screen each rollout, we compare its decoded output with the source crop (\cref{fig:supp-screening-flow}). Both images are resized to $512\times512$, with RGB values normalized to $[0,1]$. At each pixel, we compute the mean absolute RGB difference $E$ after Gaussian blurring and the channel-averaged local SSIM $S$. We then combine color error and structural dissimilarity into a smoothed difference map
\begin{equation}
  D = G_{1.5} * \bigl(0.55\,E + 0.45\,\operatorname{clip}(1-S,0,1)\bigr),
  \label{eq:supp-image-difference}
\end{equation}
where $G_{1.5}$ is a Gaussian kernel ($\sigma=1.5$ pixels).

\paragraph{Local edits.}
For local edits, we seek a coherent change confined to a limited region, with sufficient unchanged content for preservation training. We first detect changed regions by thresholding $D$ using Otsu's method~\citep{otsu1979threshold}, closing small gaps, and removing small components. We then apply five checks: (i) \emph{changed area}, requiring the detected components to cover 0.2--20\% of the image, (ii) \emph{bounding box}, limited to 35\% of the image, (iii) \emph{dominant component}, accounting for at least 65\% of the detected area, and (iv) \emph{mean SSIM}, which must be at least 0.72. Finally, (v) \emph{unchanged patch} requires at least one fully preserved $256\times256$-pixel square aligned with the backbone's spatial grid, after expanding the detected region by a margin equivalent to 48 pixels in the original crop.

\paragraph{Identity edits.}
For identity edits, the entire image should remain unchanged. We therefore apply three checks: (i) \emph{mean SSIM} of at least 0.82, (ii) \emph{mean RGB error} of at most 0.065, and (iii) \emph{high-difference fraction}, allowing at most 10\% of pixels with $D\geq0.22$. The first two measure overall similarity by averaging $S$ and $E$, while the third limits the extent of pronounced changes.

\subsection{Preservation Mask Construction}
\label{app:mask-construction}

For accepted local edits, we reuse the detected change components to construct the preservation mask. We expand these components by only five pixels in the original $1024\times1024$ crop, rather than the wider 48-pixel margin used for screening. This narrower margin allows preservation supervision closer to the edit boundary. On the base model's spatial grid ($64\times64$ for FLUX.2 Klein), a cell receives $M=1$ only if every pixel it covers lies outside the expanded region, and $M=0$ otherwise.

The training target takes the source state wherever $M=1$ and the frozen rollout endpoint wherever $M=0$. As illustrated in~\cref{fig:supp-screening-flow}, the black cells retain the edited shirt from the rollout, while the surrounding white cells use the source states. For accepted identity edits, we set $M=1$ everywhere, yielding the all-white mask.

\begin{figure}[t]
  \centering
  \vspace{-6mm}
  \includegraphics[width=\textwidth]{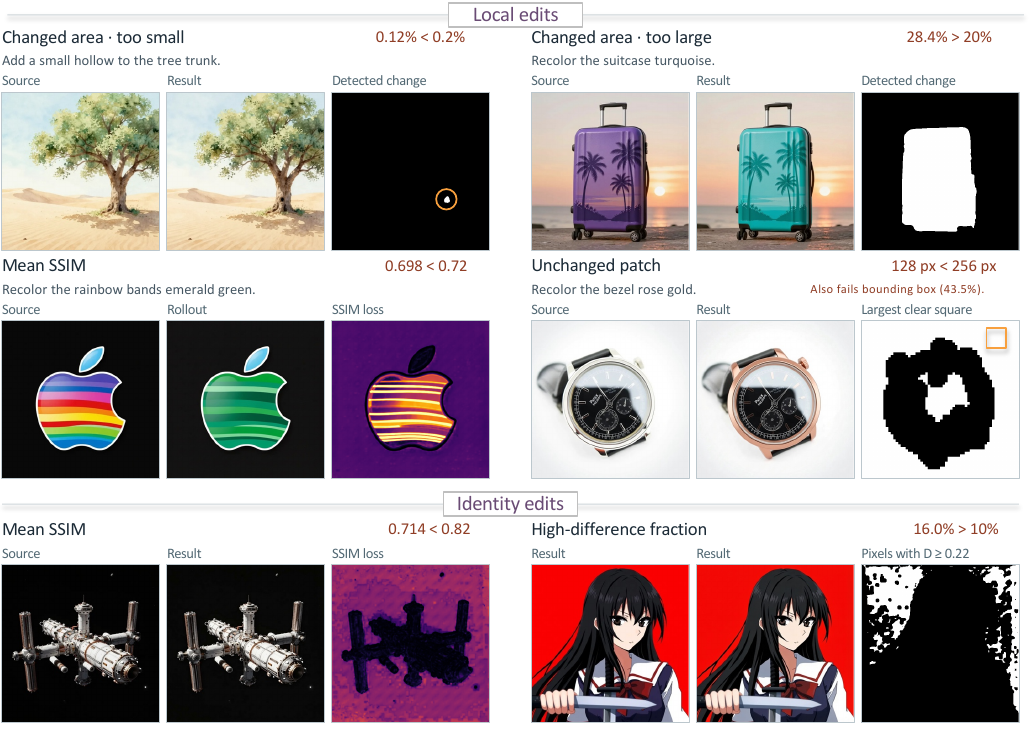}
  \vspace{-6mm}
    \caption{\textbf{Examples of rejected rollouts from FLUX.2 Klein.} Local edits illustrate failures across different locality checks, while identity edits highlight structural and spatial inconsistencies. Each example shows the diagnostic visualization with the failed criterion.}
    \vspace{-3mm}
    \label{fig:supp-screening-examples}
\end{figure}

\subsection{Filtering Statistics and Dataset Composition}
\label{app:data-statistics}

We apply the same 47,788 editing requests to each base model and screen their results using the checks described in~\cref{app:data-checks}. The number of retained trajectories are shown in~\cref{tab:supp-data-composition}. To illustrate how the checks filter out unsuitable rollouts,~\cref{fig:supp-screening-examples} presents rejection examples from FLUX.2 Klein. The calendar example shows why bounding-box coverage matters: although only 14.2\% of pixels are changed, their bounding box spans 68.1\% of the image. Conversely, the sneaker edit remains spatially confined, but its largest connected component accounts for only 34.5\% of the detected area. These cases illustrate how different checks capture distinct forms of unsuitable changes.

\begin{table}[t]
  \centering
  \caption{\textbf{Retained trajectories by different base models.} Counts are followed by acceptance rates. Each backbone starts with 35,841 local and 11,947 identity requests.}
  \label{tab:supp-data-composition}
  \small
  \setlength{\tabcolsep}{2pt}
  \begin{tabular*}{\linewidth}{@{\extracolsep{\fill}}lrrr@{}}
    \toprule
    Base model & Local edits & Identity edits & Total retained \\
    \midrule
    FLUX.1 Kontext & 17,936 (50.0\%) &  9,698 (81.2\%) & 27,634 (57.8\%) \\
    FLUX.2 Klein   & 28,444 (79.4\%) & 11,916 (99.7\%) & 40,360 (84.5\%) \\
    SenseNova-U1.5 & 22,393 (62.5\%) & 10,033 (84.0\%) & 32,426 (67.9\%) \\
    Qwen-Image-2.1 & 24,276 (67.7\%) &  9,942 (83.2\%) & 34,218 (71.6\%) \\
    \bottomrule
  \end{tabular*}
  \vspace{-6mm}
\end{table}

Identity edits show a different rejection pattern. Of the 31 rejected FLUX.2 Klein identity rollouts, 29 fall below the SSIM threshold and 12 exceed the high-difference fraction limit, with 10 failing both. Notably, all 11,947 identity rollouts satisfy the mean RGB error threshold, highlighting the importance of structural and spatial checks beyond average color differences.

After identity and locality checks, we split the retained trajectories by source image, reserving approximately 5\% of sources for validation and keeping all edits associated with the same image in the same split.

\section{Training Details and Algorithm}
\label{app:training-details}

\subsection{Base Models and Training Configurations}
\label{app:training-configurations}
\label{app:base-model-configurations}

For each base model, we initialize a trainable refined model and a frozen base model from the same checkpoint, as listed in~\cref{tab:supp-training-configs}. For FLUX.1 Kontext [dev]~\citep{labs2025flux1kontextflowmatching}, FLUX.2 [klein] Base 9B~\citep{flux2klein2026}, and Qwen-Image-2.1~\citep{qwen2026image21}, we update the denoising transformer while keeping the VAEs and conditioning encoders frozen. SenseNova-U1.5-8B-MoT~\citep{diao2026sensenova} uses a mixture-of-transformers (MoT) architecture, for which we also allow gradients to flow through the conditioning path.

\begin{table}[!htbp]
\centering
\vspace{-6mm}
\caption{\textbf{Training configurations for the four base models.} Each steering onset corresponds to a separate training run. Bold onsets identify the checkpoints used for the qualitative comparisons in the main paper.}
\label{tab:supp-training-configs}
\small
\setlength{\tabcolsep}{2.5pt}
\renewcommand{\arraystretch}{1.07}
\resizebox{\linewidth}{!}{%
  \begin{tabular}{@{}lcccc@{}}
\toprule
\textbf{Setting} & \textbf{FLUX.1 Kontext [dev]} & \textbf{FLUX.2 [klein] Base 9B} & \textbf{SenseNova-U1.5-8B-MoT} & \textbf{Qwen-Image-2.1} \\
\midrule
\multicolumn{5}{@{}l}{\textit{Settings by base model}} \\
Trainable modules & Full denoiser & Full denoiser & Full MoT & Full denoiser \\
Trainable parameters & 11.90B & 9.08B & 17.53B & 7.12B \\
Denoising state & Unpacked VAE latent & Unpacked VAE latent & Packed RGB patches & Normalized VAE latent \\
Guidance scale & Embedded, 2.5 & Text CFG, 4 & Text 4, image 1 & True CFG, 1 \\
Timestep shift & Resolution-dependent & Resolution-dependent & Fixed, 3 & Resolution-dependent \\
Denoising steps per edit & 30 & 50 & 50 & 40 \\
Forward indices & $0$--$29$ & $0$--$49$ & $1$--$50$ & $0$--$39$ \\
Steering onset $k_s$ (step) & $22,\,\mathbf{25},\,27$ & $43,\,\mathbf{46},\,48$ & $37,\,\mathbf{42},\,47$ & $24,\,\mathbf{30},\,35$ \\
Peak learning rate & $10^{-5}$ & $10^{-5}$ & $10^{-5}$ & $10^{-6}$ \\
Final-step loss weight & 1 & 1 & 1 & 0.1 \\
\midrule
\multicolumn{5}{@{}l}{\textit{Shared settings}} \\
Training examples & \multicolumn{4}{l}{Local and identity edits on $1024\times1024$ center crops (retained counts in~\cref{tab:supp-data-composition})} \\
Data split and seed & \multicolumn{4}{l}{Approximately 95\% training and 5\% validation, grouped by source image. Training and split seed 42.} \\
Optimizer & \multicolumn{4}{l}{AdamW, $\beta_1=0.9$, $\beta_2=0.999$, $\epsilon=10^{-8}$, weight decay $0.01$, gradient norm clipped to $1.0$} \\
Training schedule & \multicolumn{4}{l}{2,000 optimizer updates. Linear warmup for 100 updates, cosine decay on a 10,000-update horizon.} \\
Training loss & \multicolumn{4}{l}{Mean squared velocity error over the full state. Unit weight before the final forward pass.} \\
Effective batch size & \multicolumn{4}{l}{16 trajectories per update ($1$ per GPU $\times$ $8$ GPUs $\times$ $2$ gradient accumulation steps)} \\
Hardware and sharding & \multicolumn{4}{l}{8 NVIDIA H100 GPUs per run, fully sharded data parallelism (FSDP2)} \\
Precision & \multicolumn{4}{l}{BF16 trainable weights and forward passes. FP32 velocity targets, loss, and gradient reduction.} \\
Step sampling & \multicolumn{4}{l}{One forward step per edit, sampled over native time intervals with $\rho=0.8$ and $\beta=3$ (\cref{eq:supp-training-sampling})} \\
Preservation endpoint & \multicolumn{4}{l}{Source state in preserved regions, frozen rollout endpoint in edited regions (\cref{app:mask-construction})} \\
Velocity target & \multicolumn{4}{l}{Frozen base velocity through $k_s$, then blend toward the preservation velocity with odds ratio $r=1$ (\cref{eq:supp-training-blend})} \\
\bottomrule
\end{tabular}%
  }
\vspace{-6mm}
\end{table}

We retain each base model's native guidance configuration during training. Kontext uses a guidance embedding, while Klein employs CFG with an empty negative prompt. SenseNova applies separate text and image guidance scales without guidance normalization, whereas Qwen uses only the conditional prediction at a guidance scale of 1.

\subsection{On-Policy Training}
\label{app:training-algorithm}
\label{app:trajectory-caching}

Each training example consists of a source image, an editing instruction, and a preservation-aware endpoint $x_1^{\mathrm{tgt}}$ constructed as described in~\cref{app:mask-construction}. While the target endpoint remains fixed for each example, the intermediate states encountered by the refined model change as training progresses. To account for this evolving state distribution, we let the refined model perform several denoising steps before supervising its prediction at a sampled step. These self-rollouts take place entirely within a single editing turn, without requiring rollouts across successive edits.

\newcounter{suppalgorithm}
\renewcommand{\thesuppalgorithm}{S\arabic{suppalgorithm}}
\begin{figure}[htbp]
  \centering
  \begin{minipage}{\linewidth}
    \refstepcounter{suppalgorithm}\label{alg:supp-tailsteer}
    \small
    \hrule
    \smallskip
    \textbf{Algorithm \thesuppalgorithm. TailSteer training with on-policy rollouts}\par
    \smallskip
    \hrule
    \smallskip
    \begin{tabular}{@{}r@{\hspace{.55em}}p{\dimexpr\linewidth-1.8em\relax}@{}}
    1 & Sample a training example with editing condition $c$, cached base-model states, and target endpoint $x_1^{\mathrm{tgt}}$. Sample a denoising step $k$ following~\cref{app:training-step-sampling}. \\
    2 & For $k\leq k_s$, select the nearest preceding cached state (or the initial noise state for the first step). For $k>k_s$, select $x_{k_s}^{\mathrm{base}}$. \\
    3 & Roll out the refined model from the selected state to $x_k$ without tracking gradients. \\
    4 & At $x_k$, evaluate the frozen base model's velocity $v_k^{\mathrm{base}}=v^{\mathrm{base}}(x_k,k,c)$ and the preservation velocity $v_k^{\mathrm{pres}}=(x_1^{\mathrm{tgt}}-x_k)/(1-k)$. \\
    5 & Set $v_k^*=v_k^{\mathrm{base}}$ for $k\leq k_s$; otherwise, blend $v_k^{\mathrm{base}}$ and $v_k^{\mathrm{pres}}$ using $\lambda(k)$ as defined in~\cref{app:training-velocity-blending}. \\
    6 & Backpropagate $\mathrm{MSE}(v_\theta(x_k,k,c),\mathrm{sg}[v_k^*])$ with the loss weight specified in~\cref{tab:supp-training-configs}. \\
    7 & Accumulate gradients over two microbatches, clip gradients, and perform one AdamW update. \\
    \end{tabular}
    \smallskip
    \hrule
  \end{minipage}
\end{figure}

Algorithm~\ref{alg:supp-tailsteer} summarizes the training procedure using the denoising coordinate $k$ defined in the main paper, which equals one minus the scheduled noise level and increases from 0 (noise) to 1 (output). In the prior region ($k\leq k_s$), each self-rollout starts from the nearest strictly preceding cached base-model state, at most three sampling steps earlier. In the transition region ($k>k_s$), all self-rollouts instead start from the fixed steering-onset state $x_{k_s}^{\mathrm{base}}$, allowing the refined model to determine the subsequent trajectory without intermediate restarts from the base model. At the resulting state $x_k$, we recompute the frozen base model's velocity and the preservation velocity to construct the supervision target. Gradients flow only through the refined model's prediction at the sampled step, while the preceding self-rollout remains gradient-free.

\begin{figure}[t]
  \centering
  \vspace{-6mm}
  \includegraphics[width=0.85\linewidth]{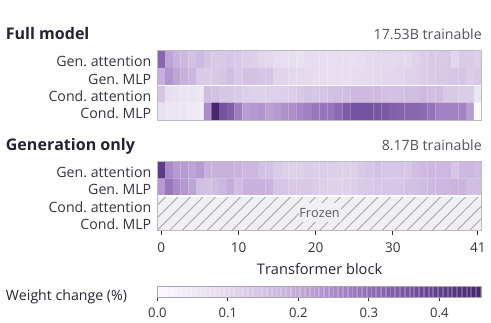}
    \caption{\textbf{Layerwise parameter changes under different training scopes.} Full-model training yields larger changes in conditioning MLPs but smaller changes in generation MLPs, suggesting the importance of adapting the conditioning pathway. Each cell shows relative weight change in a SenseNova block. Both runs share the same initialization, with hatching indicating frozen modules.}
  \label{fig:supp-scope-updates}
\end{figure}

\subsection{Training-Step Sampling Strategy}
\label{app:training-step-sampling}

During training, we supervise one denoising step per sampled edit. Steps before the steering onset are trained to retain the base model's earlier dynamics, while later steps learn preservation corrections. To emphasize these corrections without neglecting earlier dynamics, we sample steps throughout the trajectory with a bias toward its tail. Specifically, we use a mixture of a uniform distribution and a distribution concentrated near the clean endpoint:
\begin{equation}
  p(k)=\rho+(1-\rho)\,\beta k^{\beta-1}, \qquad 0\leq k\leq 1.
  \label{eq:supp-training-sampling}
\end{equation}
Here $\rho$ controls the fraction of uniform sampling, while $\beta>1$ determines how strongly the remaining probability mass favors later steps. We set $\rho=0.8$ and $\beta=3$. Since each base model follows a nonuniform denoising schedule, we convert this continuous density into discrete sampling probabilities by integrating over the interval between each forward pass and the next, with the final interval ending at $k=1$. We further examine the effects of different sampling strategies in~\cref{app:on-policy-sampling}.

\subsection{Velocity Blending}
\label{app:training-velocity-blending}

While the sampling strategy determines which steps receive supervision, velocity blending controls the balance between preserving the base model's editing dynamics and correcting drift at each sampled step. Up to the steering onset $k_s$, we set $\lambda(k)=1$ to match the frozen base model's velocity. Beyond $k_s$, we progressively shift supervision toward the preservation velocity, which directs the current state toward the target endpoint $x_1^{\mathrm{tgt}}$. Let $\tau$ denote the final denoising step before reaching the clean endpoint. We define the base velocity weight in the tail as
\begin{equation}
  \lambda(k)=\frac{r(\tau-k)}{r(\tau-k)+(k-k_s)},
  \qquad k_s<k\leq\tau.
  \label{eq:supp-training-blend}
\end{equation}
where $r>0$ controls the shape of the transition. With our default $r=1$, $\lambda(k)$ decreases linearly from 1 at $k_s$ to 0 at $\tau$. A larger $r$ retains more of the base model's editing velocity, while a smaller $r$ introduces stronger preservation supervision earlier in the tail. Regardless of $r$, the final step uses only the preservation velocity $v_\tau^{\mathrm{pres}}$, evaluated at the refined model's own on-policy state. For example, with the FLUX.2 [klein] Base 9B steering onset at step 46, steps 47--49 receive preservation supervision, with step 49 relying on it exclusively. We analyze the effects of steering onset and blending schedules in~\cref{app:velocity-blending}.

\section{Practical Training Insights and Analysis}
\label{app:training-analysis}

The preceding section describes TailSteer's training procedure and the configurations used in our experiments. Here, we examine the training choices behind these settings and draw practical insights for adapting TailSteer to other base models. We assess both single-turn editing capability and cumulative drift resistance when comparing training strategies. Unless otherwise noted, all comparisons use checkpoints after 2,000 updates.

\subsection{Trainable Parameter Selection}
\label{app:trainable-parameters}

We first investigate whether updating the generation branch alone is sufficient, or whether the conditioning pathway should also adapt. In SenseNova-U1.5-8B-MoT, the vision--language branch encodes the source image and editing instruction into conditioning tokens, which the generation branch attends to when predicting velocities for noisy image patches. At $k_s=42$, we compare generation-only training, including its input and output modules (8.17B parameters), with full-model training (17.53B parameters). The latter also allows preservation supervision to update the source representations used by the denoiser. Both runs share the same initialization, data, and training schedule.

\begin{table}[t]
  \centering
  \vspace{-10mm}
  \caption{\textbf{Effect of trainable parameter selection in SenseNova.} RISE Bench accuracy and cumulative object accuracy (COA) are percentages. COA is averaged over all ten turns. PSNR (dB), SSIM, and LPIPS are measured outside the edit box at turn 10.}
  \label{tab:supp-sensenova-scope}
  \small
  \setlength{\tabcolsep}{1.5pt}
  \begin{tabular*}{\linewidth}{@{\extracolsep{\fill}}lrrrrr@{}}
    \toprule
    & \multicolumn{1}{c}{RISE} & \multicolumn{4}{c}{Add \& Preserve} \\
    \cmidrule(lr){2-2}\cmidrule(l){3-6}
    Trainable scope & Acc.$\uparrow$ & COA$\uparrow$ & PSNR$\uparrow$ & SSIM$\uparrow$ & LPIPS$\downarrow$ \\
    \midrule
    Generation only & 26.94 & 92.63 & \textbf{18.34} & 0.470 & 0.455 \\
    Full model & \textbf{29.44} & \textbf{94.78} & 18.22 & \textbf{0.489} & \textbf{0.417} \\
    \bottomrule
  \end{tabular*}
\end{table}

As shown in \cref{tab:supp-sensenova-scope}, full-model training improves RISEBench accuracy by 2.50 percentage points, while GEdit-Bench scores in both languages and the ImgEdit-Bench Basic score each change by less than 0.05 points. It also improves COA and exterior SSIM and LPIPS, with only a slight decrease in PSNR. These results suggest that jointly updating the conditioning and generation branches achieves a better balance between editing capability and preservation than updating generation branch alone.

The layerwise analysis in \cref{fig:supp-scope-updates} suggests that effective preservation refinement benefits from adapting how source-image information interacts with noisy image features, rather than updating the generation branch alone. Freezing the conditioning pathway may limit this adaptation, leaving residual preservation errors that accumulate across editing turns. For models with separate conditioning and generation branches, we therefore recommend exploring joint training, although the contribution of individual modules remains unclear.

\begin{figure}[t]
  \centering
  \vspace{-2mm}
  \includegraphics[width=\textwidth]{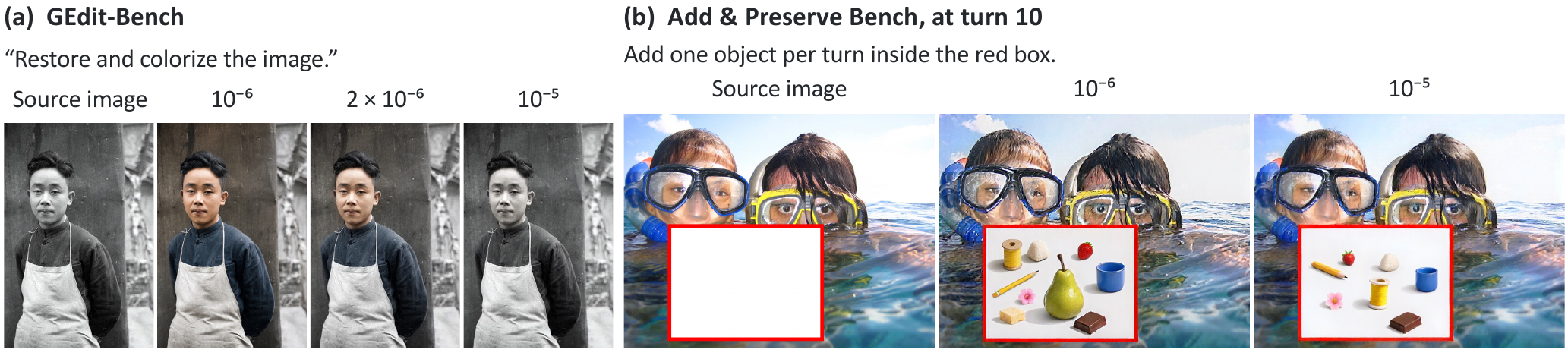}
  \vspace{-6mm}
  \caption{\textbf{Excessive learning rates can suppress editing in Qwen-Image-2.1.} (a) Under fixed-step sampling, lower-rate checkpoints successfully colorize the portrait, while $10^{-5}$ leaves it nearly unchanged. (b) With time-mixture sampling (\cref{app:on-policy-sampling}), the $10^{-5}$ checkpoint loses previously added objects, including the pear, by turn 10. All uses identical sources, instructions, and inference seeds.}
  \vspace{-3mm}
  \label{fig:supp-qwen-rate}
\end{figure}

\subsection{Learning Rate Selection}
\label{app:learning-rate-analysis}

Although on-policy training and cached rollouts help preserve the base model's early denoising dynamics, overly strong updates in the trajectory tail may still suppress edits established earlier in sampling. The learning rate must therefore be chosen carefully to improve preservation without disrupting the base model's learned editing priors.

As shown in \cref{tab:supp-update-strength}, higher learning rates improve preservation but affect editing capability differently across base models. Klein tolerates stronger updates, whereas Qwen-Image-2.1 is more sensitive. At learning rate $10^{-5}$, Qwen leaves the grayscale portrait nearly unchanged and loses previously added objects despite better exterior preservation (\cref{fig:supp-qwen-rate}), a trend that persists under alternative step sampling (\cref{app:on-policy-sampling}). This sensitivity may partly stem from Qwen's timestep-independent image-conditioning features, which remain fixed throughout denoising rather than adapting to each step. Such conditioning may make aggressive tail updates more likely to suppress intended edits.

We therefore recommend starting at a peak learning rate of $10^{-6}$ and increasing it gradually while monitoring both editing capability and preservation. Updates that are too small may leave drift insufficiently corrected, while overly large updates risk suppressing the base model's editing prior. The goal is to find a suitable correction strength that improves long-term preservation without compromising the model's original capabilities. Accordingly, we select $10^{-5}$ for Klein and $10^{-6}$ for Qwen (\cref{tab:supp-training-configs}).

\begin{table}[!t]
  \centering
  \vspace{-10mm}
  \caption{\textbf{Effect of peak learning rate after 2,000 updates.} Higher rates can improve preservation at the cost of editing capability. Add \& Preserve metrics are evaluated at turn 10. Bold indicates selected learning rates and best scores.}
  \label{tab:supp-update-strength}
  \small
  \setlength{\tabcolsep}{1.0pt}
  \begin{tabular*}{\linewidth}{@{\extracolsep{\fill}}llcccccc@{}}
    \toprule
    & & \multicolumn{2}{c}{Single-turn} & \multicolumn{4}{c}{Add \& Preserve} \\
    \cmidrule(lr){3-4}\cmidrule(l){5-8}
    Model & Peak LR & GEdit$\uparrow$ & RISE$\uparrow$ & COA$\uparrow$ & PSNR$\uparrow$ & SSIM$\uparrow$ & LPIPS$\downarrow$ \\
    \midrule
    Klein & $10^{-6}$ & \textbf{7.47} & 11.67 & 11.0 & 16.33 & 0.506 & 0.356 \\
          & {\boldmath$10^{-5}$} & 7.36 & \textbf{13.33} & \textbf{11.4} & \textbf{19.50} & \textbf{0.610} & \textbf{0.293} \\
    \midrule
    Qwen  & {\boldmath$10^{-6}$} & \textbf{8.00} & \textbf{19.72} & 61.6 & 18.31 & 0.522 & 0.236 \\
          & $2\!\times\!10^{-6}$ & 7.77 & 15.56 & \textbf{63.8} & 19.03 & 0.542 & 0.270 \\
          & $10^{-5}$ & 6.62 & 12.22 & 37.9 & \textbf{21.32} & \textbf{0.679} & \textbf{0.219} \\
    \bottomrule
  \end{tabular*}
\end{table}

\begin{figure}[t]
  \centering
   \vspace{-3mm}
  \includegraphics[width=0.8\linewidth]{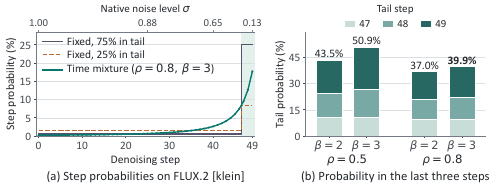}
   \vspace{-3mm}
     \caption{\textbf{Balancing early- and tail-step supervision.} (a) Sampling probabilities on FLUX.2 [klein] Base 9B, with the shaded region marking the three steps after $k_s=46$. The 25\% fixed split is shown for comparison. (b) Tail probabilities under different time-mixture parameters. Our setting ($\rho=0.8$, $\beta=3$) allocates 39.9\% of samples to the tail while retaining broader early-step coverage.}
     \vspace{-3mm}
  \label{fig:supp-klein-step-sampling}
\end{figure}

\subsection{Training-Step Sampling Strategies}
\label{app:on-policy-sampling}

After choosing how strongly to update the model, we examine how to distribute supervision across denoising steps. As described in \cref{app:training-step-sampling}, TailSteer learns preservation corrections near the trajectory tail while matching the base model's earlier dynamics to retain its editing priors. Allocating more samples to the tail may strengthen preservation, but could also leave earlier steps insufficiently supervised. To study this trade-off, we compare our time mixture with a deliberately tail-heavy fixed split that assigns 75\% of samples to the tail. For Klein, this concentrates most training updates on just three of its 50 denoising steps (\cref{fig:supp-klein-step-sampling}).

The results in \cref{tab:supp-supervision-distribution} show that stronger tail emphasis does not consistently improve performance. The time mixture improves cumulative object retention across all three base models while largely maintaining single-turn editing capability, although other metrics show mixed results. One possible explanation is that excessive tail sampling leaves too little supervision to keep the refined model's earlier dynamics aligned with those of the frozen base model, potentially weakening its learned editing priors. These observations suggest that effective training requires sufficient supervision for both early-dynamics matching and tail refinement. In practice, we recommend adjusting this balance according to each base model's native noise schedule.

\begin{table}[!t]
  \centering
  \vspace{-10mm}
  \caption{\textbf{Effect of training-step sampling after 2,000 updates.} The time mixture redistributes supervision toward earlier steps, with model-dependent effects on editing and preservation. Tail denotes the percentage of samples drawn after the steering onset.}
  \label{tab:supp-supervision-distribution}
    \footnotesize
    \setlength{\tabcolsep}{1.5pt}
    \resizebox{\linewidth}{!}{%
    \begin{tabular}{@{}lllcccccccccc@{}}
      \toprule
      & & & & \multicolumn{2}{c}{Text-to-image} & \multicolumn{3}{c}{Single-turn editing} & \multicolumn{2}{c}{Multi-turn editing} & \multicolumn{2}{c}{Add \& Preserve} \\
      \cmidrule(lr){5-6}\cmidrule(lr){7-9}\cmidrule(lr){10-11}\cmidrule(l){12-13}
      Model & Peak LR & Sampler & Tail (\%) & Qwen-IB$\uparrow$ & BizGen$\uparrow$ & GEdit$\uparrow$ & RISE$\uparrow$ & ImgEdit$\uparrow$ & SR@5$\uparrow$ & CC@3$\uparrow$ & COA$\uparrow$ & LPIPS$\downarrow$ \\
      \midrule
      Klein & $10^{-5}$ & Fixed split & 75.0 & \textbf{49.72} & 9.34 & 7.27 & 11.67 & 4.17 & \textbf{35} & \textbf{87.65} & 30.77 & \textbf{0.285} \\
       &  & Time mixture & 39.9 & 49.42 & \textbf{10.97} & \textbf{7.52} & \textbf{13.61} & \textbf{4.19} & 33 & 87.57 & \textbf{31.84} & 0.290 \\
      \midrule
      SenseNova & $10^{-5}$ & Fixed split & 75.0 & 51.69 & \textbf{47.98} & \textbf{8.28} & 28.89 & 4.41 & 47 & 76.81 & 94.05 & \textbf{0.414} \\
       &  & Time mixture & 43.9 & \textbf{52.24} & 45.73 & 8.27 & \textbf{29.44} & 4.41 & \textbf{55} & \textbf{77.03} & \textbf{94.78} & 0.417 \\
      \midrule
      Qwen & $10^{-6}$ & Fixed split & 75.0 & \textbf{53.25} & 56.85 & 8.00 & \textbf{19.72} & 4.42 & 40 & 87.35 & 65.85 & \textbf{0.236} \\
       &  & Time mixture & 42.3 & 53.06 & \textbf{57.38} & \textbf{8.04} & 18.06 & 4.42 & \textbf{44} & \textbf{89.70} & \textbf{66.94} & 0.239 \\
      \bottomrule
    \end{tabular}%
    }
\end{table}

\begin{table}[!t]
  \centering
  \vspace{-5mm}
  \caption{\textbf{Velocity-blending trade-offs on FLUX.2 [klein] Base 9B.} All runs use 2,000 updates, varying only $r$. $\mathrm{COA}_{1:10}$ averages over all turns and sequences, while $\mathrm{COA}_{10}$ measures final-turn retention of all ten requested objects.}
  \label{tab:supp-odds-sweep}
  \begingroup
    \footnotesize
    \setlength{\tabcolsep}{2.0pt}
    \resizebox{\linewidth}{!}{%
    \begin{tabular}{@{}lccccccccccc@{}}
      \toprule
      & \multicolumn{2}{c}{Pres.\ weight (\%)} & \multicolumn{4}{c}{Single-turn editing} & \multicolumn{5}{c}{Add \& Preserve} \\
      \cmidrule(lr){2-3}\cmidrule(lr){4-7}\cmidrule(l){8-12}
      $r$ & Step 47 & Step 48 & GEdit-EN$\uparrow$ & GEdit-ZH$\uparrow$ & RISE$\uparrow$ & ImgEdit$\uparrow$ & COA$_{1:10}\uparrow$ & COA$_{10}\uparrow$ & PSNR$\uparrow$ & SSIM$\uparrow$ & LPIPS$\downarrow$ \\
      \midrule
      0.01 & 97.4 & 99.3 & 7.48 & 7.40 & 10.28 & 4.19 & 31.30 & \textbf{12.1} & \textbf{19.98} & \textbf{0.623} & \textbf{0.284} \\
      0.1 & 78.8 & 93.7 & \textbf{7.53} & \textbf{7.49} & \textbf{13.89} & 4.23 & 32.01 & 11.7 & 19.90 & 0.615 & 0.297 \\
      \textbf{1} (default) & 27.1 & 59.7 & 7.52 & 7.39 & 13.61 & 4.19 & 31.84 & 12.0 & 19.96 & 0.621 & 0.290 \\
      10 & 3.6 & 12.9 & 7.51 & 7.41 & 13.61 & \textbf{4.24} & 31.96 & 11.4 & 19.73 & 0.608 & 0.301 \\
      100 & 0.4 & 1.5 & 7.32 & 7.37 & 12.22 & 4.18 & \textbf{32.38} & 11.7 & 19.77 & 0.607 & 0.304 \\
      \bottomrule
    \end{tabular}%
    }
    \par
  \endgroup
  \medskip
  \includegraphics[width=\textwidth]{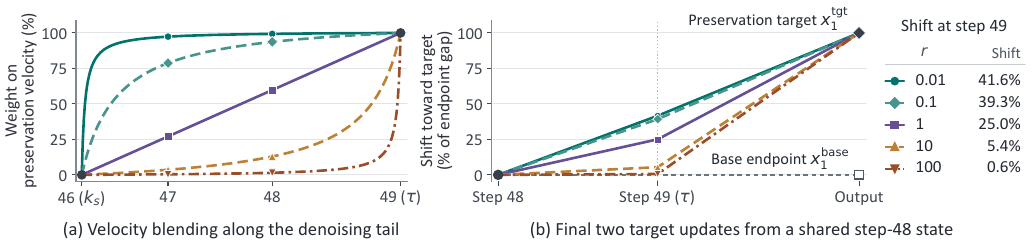}
  \captionof{figure}{\textbf{Controlling the timing of tail-preservation corrections.} (a) Smaller $r$ increases the preservation-velocity weight earlier in the tail, while larger $r$ delays the correction. (b) From the same step-48 state, different blending schedules produce different shifts toward preservation at step 49, despite sharing the same final target. These paths illustrate supervision targets, not actual refined-model trajectories.}
  \vspace{-3mm}
  \label{fig:supp-odds-tail-path}
\end{table}

\subsection{Velocity Blending}
\label{app:velocity-blending}

While step sampling determines where supervision is applied, velocity blending controls how the model transitions from its original editing dynamics toward preservation. As defined in \cref{eq:supp-training-blend}, the odds ratio $r$ adjusts the balance between the frozen base model's velocity and the preservation velocity after the steering onset. Smaller $r$ introduces stronger preservation corrections earlier in the tail, whereas larger $r$ keeps supervision closer to the base model's editing dynamics for longer, with both ultimately converging to full preservation supervision at the final step.

\Cref{fig:supp-odds-tail-path} illustrates this effect. Starting from the same step-48 state, smaller $r$ shifts the target update further toward preservation before the final step. Although all blending schedules ultimately point to the same preservation endpoint, they guide the model through different tail trajectories. The refined model learns to approximate these targets on its own states, so the timing of correction can affect how well it preserves both image details and the intended edits.

The results in \cref{tab:supp-odds-sweep} show that correction timing affects preservation and editing capability in different ways. Very early corrections ($r=0.01$) achieve the strongest exterior preservation but weaken instruction following, suggesting that excessive preservation pressure may interfere with intended edits. Moderate values improve single-turn performance while maintaining strong preservation. Delaying corrections further offers no consistent benefit: although $r=100$ achieves the highest average COA across turns, its final-turn retention and single-turn performance are weaker. We therefore adopt $r=1$ as a balanced linear default and recommend adjusting the correction timing according to the model's preservation and editing performance.

\begin{table}[t]
  \centering
    \caption{\textbf{Parameter changes across three base models after TailSteer training.} Count and Share report the number and percentage of BF16 weights changed after 2,000 updates, while $\Delta=100\lVert W-W_0\rVert_F/\lVert W_0\rVert_F$ measures relative change within each parameter group. Darker cells indicate larger changes on a shared scale. Dashes denote absent modules.}
  \label{tab:supp-parameter-changes}
  \small
  \setlength{\tabcolsep}{2pt}
  \renewcommand{\arraystretch}{1.07}
  \definecolor{weightteal}{HTML}{236774}
  \newcommand{\weightdelta}[2]{{\setlength{\fboxsep}{0pt}%
    \colorbox{weightteal!#1}{\makebox[3.2em][r]{\strut #2\hspace{0.25em}}}}}
  \newcommand{\nogroup}{\multicolumn{3}{c}{\textcolor{black!55}{---}}}
  \resizebox{\linewidth}{!}{%
  \begin{tabular}{@{}lrrcrrcrrc@{}}
    \toprule
    & \multicolumn{3}{c}{\textbf{FLUX.2 [klein] Base 9B}}
    & \multicolumn{3}{c}{\textbf{Qwen-Image-2.1}}
    & \multicolumn{3}{c}{\textbf{SenseNova-U1.5-8B-MoT}} \\
    & \multicolumn{3}{c}{9.08B denoiser weights, $k_s=46$}
    & \multicolumn{3}{c}{7.12B denoiser weights, $k_s=30$}
    & \multicolumn{3}{c}{17.53B MoT weights, $k_s=42$} \\
    \cmidrule(lr){2-4}\cmidrule(lr){5-7}\cmidrule(l){8-10}
    Parameter group & Count & Share (\%) & $\Delta$ (\%)
                    & Count & Share (\%) & $\Delta$ (\%)
                    & Count & Share (\%) & $\Delta$ (\%) \\
    \midrule
    Timestep embedding & 4.02M & 22.55 & \weightdelta{68}{2.386} & 2.23M & 12.52 & \weightdelta{62}{1.982} & 2.05M & 11.51 & \weightdelta{41}{0.889} \\
    Block modulation & 57.5M & 22.83 & \weightdelta{38}{0.758} & 2.95M & 4.39 & \weightdelta{25}{0.333} & \nogroup \\
    Noise-scale embedding & \nogroup & \nogroup & 1.93M & 10.84 & \weightdelta{43}{0.974} \\
    Image input & 97.4k & 18.58 & \weightdelta{19}{0.184} & 4.12k & 1.57 & \weightdelta{7}{0.027} & 2.71M & 7.71 & \weightdelta{22}{0.248} \\
    Text input & 5.73M & 11.38 & \weightdelta{31}{0.510} & 302.1k & 0.90 & \weightdelta{9}{0.038} & 5.26M & 0.84 & \weightdelta{12}{0.077} \\
    Denoiser output & 25.7M & 75.48 & \weightdelta{49}{1.252} & 1.23M & 7.21 & \weightdelta{24}{0.301} & 1.20M & 12.12 & \weightdelta{21}{0.238} \\
    Language output head & \nogroup & \nogroup & 0 & 0 & \weightdelta{0}{0} \\
    \addlinespace[2.5pt]
    Denoiser attention & 49.9M & 9.30 & \weightdelta{24}{0.295} & 23.0M & 1.07 & \weightdelta{12}{0.075} & 60.4M & 3.43 & \weightdelta{15}{0.118} \\
    Conditioning attention & 46.6M & 8.67 & \weightdelta{21}{0.225} & \nogroup & 108.0M & 6.13 & \weightdelta{16}{0.130} \\
    Denoiser MLP & 116.4M & 9.63 & \weightdelta{26}{0.344} & 48.6M & 1.01 & \weightdelta{12}{0.076} & 206.8M & 3.26 & \weightdelta{15}{0.118} \\
    Conditioning MLP & 115.4M & 9.55 & \weightdelta{22}{0.243} & \nogroup & 471.3M & 7.43 & \weightdelta{19}{0.179} \\
    Fused attention and MLP & 529.6M & 10.12 & \weightdelta{34}{0.595} & \nogroup & \nogroup \\
    Transformer norms & 0 & 0 & \weightdelta{0}{0} & 1 & 0.01 & \weightdelta{1}{0.0002} & 1.85k & 0.26 & \weightdelta{3}{0.003} \\
    \midrule
    \textbf{All trainable weights} & 950.9M & 10.47 & \weightdelta{31}{0.498} & 78.3M & 1.10 & \weightdelta{12}{0.078} & 859.6M & 4.90 & \weightdelta{16}{0.134} \\
    Net update norm $\lVert W-W_0\rVert_F$ & \multicolumn{3}{c}{10.42} & \multicolumn{3}{c}{2.90} & \multicolumn{3}{c}{11.05} \\
    \bottomrule
  \end{tabular}%
  }
\end{table}

\subsection{Numerical Precision and Training Stability}
\label{app:numerical-precision}
\label{app:optimization-precision}

We recommend using BF16 trainable weights for TailSteer, which proved sufficient and more reliable in our experiments. In a FLUX.2 [klein] Base 9B pilot with BF16 forward passes, FP32 master weights produced overly smooth outputs with reduced fine detail, suggesting greater sensitivity to parameter updates. One possible explanation is that FP32 retains small updates that BF16 would otherwise round away, potentially disrupting the base model's learned editing priors. While FP32 may work with careful tuning, we recommend BF16 as a more stable starting point for preservation refinement.

\subsection{How TailSteer Changes the Base Model}
\label{app:parameter-analysis}
\label{app:parameter-functional-changes}

Having examined the training choices, we next investigate how much TailSteer actually changes the base models. As shown in \cref{tab:supp-parameter-changes}, only 1.10--10.47\% of BF16 weights differ from initialization across Klein, Qwen, and SenseNova, with overall relative changes below 0.5\%. These modest parameter changes accompany substantial improvements in long-term preservation, while single-turn editing capability and cumulative object retention remain largely intact.

More importantly, the changes are not evenly distributed. Time-related embeddings and modulation modules show some of the largest relative changes, while most attention and MLP weights remain close to their initialization. In SenseNova, conditioning MLPs also exhibit a higher proportion of changed weights than generation MLPs, consistent with the benefits of joint training observed in \cref{app:trainable-parameters}. This suggests that adapting how the model responds to denoising time and source-image conditioning may be particularly important for preservation refinement.

Together, these observations suggest that TailSteer improves preservation through modest, nonuniform parameter updates, particularly in time-related and conditioning modules. These changes may refine how the model uses denoising time and source-image information to predict velocities, enabling more targeted corrections near the trajectory tail while largely retaining its learned editing priors. By reducing residual preservation errors within each edit, such refinements may prevent errors from accumulating across turns, helping explain the substantial gains in long-term preservation despite limited changes to the base model.

\section{Evaluation Protocols}
\label{app:evaluation-protocols}

\subsection{Inference and Baseline Implementations}
\label{app:inference-settings}
\label{app:inference-baselines}

For a controlled comparison, we evaluate each base model and its TailSteer-refined counterpart using identical prompts, source images, and generation seeds. TailSteer uses the refined model's weights at every denoising step, with step counts and guidance scales following~\cref{tab:supp-training-configs}. We use seed 42 for text-to-image generation and a base seed of 0 for the five public editing benchmarks, while Add \& Preserve specifies a seed for each image and turn.

For comparison methods, we retain their respective inference settings. FreqEdit~\citep{liao2026freqedit} follows its native noise schedule, which is also used for the FLUX.1 Kontext [dev] base-model control on Add \& Preserve. VAE-LFA~\citep{wang2026ditdrift} applies momentum correction starting from the second turn, while Consistency LoRA V2~\citep{dx8152flux2consistency} runs on FLUX.2 [klein] 9B with guidance 1. For qualitative comparisons, MTC~\citep{zhou2025multiturn} follows its inversion and editing procedure, and VINCIE~\citep{qu2026vincie} conditions on the image and instruction history. We evaluate GPT-Image-2.5 Sunburst~\citep{openai2026images25} through its API calls.

\subsection{Text-to-Image and Single-Turn Evaluation}
\label{app:single-turn-protocols}\label{app:single-turn-evaluation}

\paragraph{Text-to-image generation.}
For Qwen-Image-Bench~\citep{li2026qwenimagebench}, we generate one $1024\times1024$ image for each of 1,000 English and 1,000 Chinese prompts. We evaluate the outputs locally using Q-Judger and the benchmark's checklists. The scoring script first averages applicable criteria within each dimension, then aggregates the scores across dimensions and prompts separately for each language. BizGenEval~\citep{li2026bizgeneval} contains 400 prompts spanning five commercial design domains. We follow the requested aspect ratios on a 32-pixel grid with a 1,024-pixel long edge, defaulting to a square canvas when no ratio is specified. We use Gemini-3-Flash-Preview to answer the benchmark's checklist questions covering attributes, layout, text, and knowledge. For each image, the Error Score is $\max(0,1-0.1e)$, where $e$ is the number of incorrect answers. We report the mean score across images as a percentage.

\paragraph{Single-turn editing.}
For GEdit-Bench~\citep{liu2025step1xedit}, we edit the same 606 source cases once in English and once in Chinese. GPT-4.1 evaluates semantic consistency and perceptual quality, and the official Overall score is computed by taking their geometric mean per image, then averaging the 11 edit-type means within each language. RISEBench~\citep{zhao2025envisioning} uses GPT-4.1 to assess reasoning and task-specific visual criteria. Accuracy is measured as the fraction of all 360 edits receiving full marks on every required criterion. For the Basic split of ImgEdit-Bench~\citep{ye2025imgedit}, GPT-4o evaluates 737 edits using the official task-specific rubrics. Our parser requires a numeric score for every requested field, after which we average scores within each of the nine edit types and weight the types equally.

\begin{figure}[t]
  \centering
  \vspace{-10mm}
  \includegraphics[width=\textwidth]{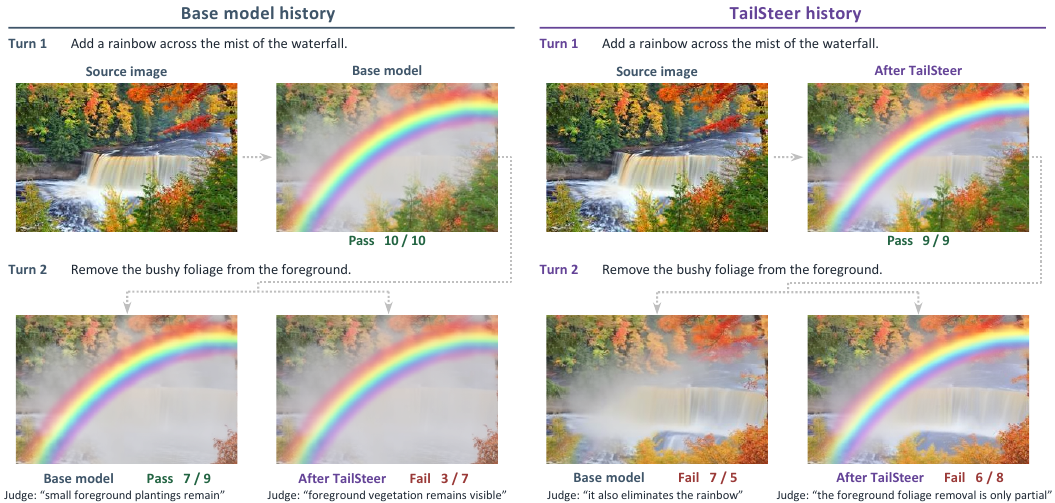}
    \caption{\textbf{Higher MSE-Bench scores do not necessarily indicate a better multi-turn model.} In sequence 6, the base model smooths away foliage when adding a rainbow, whereas TailSteer preserves more detail. At turn 2, both models edit each first-turn output with the same instruction and seed. On the smoothed base-model history (left), the judge accepts the base model's removal (7/9) despite residual foliage, but rejects TailSteer's nearly identical output (3/7), where the remaining plants are clearer. On TailSteer's history (right), the base model also removes the rainbow (7/5), while TailSteer retains it but leaves some foliage (6/8). Earlier drift may thus simplify later edits or conceal failures from the evaluator. Scores indicate prompt following/consistency, both of which must exceed 6 to pass.}
    \vspace{-3mm}
  \label{fig:supp-mse-history}
\end{figure}

\subsection{General Multi-Turn Evaluation}
\label{app:multi-turn-protocols}\label{app:general-multi-turn-evaluation}

Unlike single-turn evaluation, multi-turn benchmarks assess models as they repeatedly edit their own outputs. Each sequence begins with the benchmark source image, and each method uses its preceding output as input to the next turn. Multi-input qualitative examples additionally provide a pose or object reference at the specified turn. In the FLUX.2 latent-feedback ablation, we pass the final latent state directly to the next turn and decode each result for evaluation. All other comparisons use image feedback.

MSE-Bench~\citep{qu2026vincie} consists of 100 five-turn sequences. At each turn, GPT-5-nano evaluates the latest edit using the full prompt and image history, scoring prompt following and consistency with the input image from 0 to 10. Both scores must exceed 6 to pass. Upon failure, the evaluator retains the current scores but zeros all subsequent turns. Accordingly, SR@5 measures the fraction of sequences passing all five turns, while PF@5 and Cons.@5 average fifth-turn scores, including those zeroed after earlier failures.

EdiVal-Bench~\citep{chen2026edival} consists of 572 three-turn sequences. Its multi-pass evaluator uses Grounding DINO and Qwen2-VL-7B-Instruct to assess instruction following. IF@3 measures the fraction of sequences passing all three turns. For CC@3, we separately average third-turn DINOv3 similarity for unchanged objects and the background over their respective eligible cases, then weight the two means equally. Unlike instruction-following success, CC@3 includes all eligible cases regardless of whether the requested edits succeed.

\paragraph{Separating editing capability from drift resistance.}
Reliable multi-turn editing requires both strong single-turn editing capability and resistance to cumulative visual drift. While the preceding single-turn benchmarks assess the former, sequence-success metrics such as MSE-Bench's SR@5 and EdiVal-Bench's IF@3 combine both abilities, making it difficult to evaluate drift resistance independently. More importantly, detail lost in earlier edits may make subsequent instructions easier to execute, even when preservation is worse. For instance, FLUX.2 [klein] Base 9B completes all five turns on 40 of 100 MSE-Bench sequences, compared with 33 for TailSteer ($k_s=46$), despite TailSteer's stronger preservation performance. This seemingly counterintuitive result becomes clearer when we look beyond the scores to the editing histories behind them.

\Cref{fig:supp-mse-history} illustrates how earlier preservation can affect both subsequent editing and its evaluation. At turn 1, both models add a rainbow, but the base model smooths away surrounding foliage that TailSteer better preserves. When the next instruction asks to remove this foliage while keeping the rainbow, we test both models on each first-turn output using the same instruction and seed. On the base model's smoothed history (left), the two outputs are nearly identical (RGB SSIM of 0.979), yet the judge accepts the base model's removal (7/9) and rejects TailSteer's (3/7), where residual foliage is more discernible. On TailSteer's more detailed history (right), the base model removes the rainbow along with the foliage (7/5), while TailSteer retains the rainbow but leaves some plants (6/8). Thus, an earlier loss of detail may make subsequent edits easier or their failures less apparent, without indicating stronger preservation.

This example reveals a counterintuitive problem: greater drift in an earlier edit may make subsequent instructions easier to execute, or their failures harder for the evaluator to detect. Conversely, better preservation does not guarantee successful editing, which remains limited by the model's original capabilities. MSE-Bench further assigns zero to subsequent scores after an earlier failure, regardless of actual preservation quality. Thus, sequence success alone is not a reliable measure of drift resistance. \textbf{Since the preceding single-turn benchmarks already assess editing capability, we evaluate cumulative preservation separately, while checking whether requested edits remain intact.}

\subsection{Add \& Preserve Benchmark Construction}
\label{app:add-preserve-construction}

To evaluate cumulative drift independently of instruction success, we introduce Add \& Preserve, which confines successive object additions to a fixed region while using the surrounding scene as a preservation reference. The benchmark consists of 100 detail-rich natural images from Open Images~\cite{kuznetsova2020openimages} and MONET~\cite{aubin2026monet}, each paired with ten successive object additions. For each image, we place a rectangle spanning half its width and height, manually checking its position to ensure sufficient detail remains outside. We fill the rectangle with white and draw a 12-pixel red border inside its perimeter. The resulting image serves as both the first-turn input and the fixed reference for later evaluation.

For each image, we sample ten distinct objects without replacement using a fixed dataset seed of 42. At each turn, the instruction asks the model to add one small object in the remaining white space while preserving earlier additions, the border, and the surrounding scene. For example, a sequence may begin with a strawberry followed by a pink eraser, both of which must remain visible at turn two. Each output then becomes the input to the next turn, allowing drift to accumulate while keeping the editing region and preservation reference fixed.

\begin{figure}[b]
  \centering
  \fbox{\begin{minipage}{\dimexpr\linewidth-2\fboxsep-2\fboxrule\relax}
    \small\raggedright
    \textbf{System prompt}\par
    This is a crop of the allowed editing area. For each requested object, check whether a recognizable instance with the specified color/material is visibly present and its body is fully inside the crop. Ignore size, shadows and background quality. Answer yes, no, or unsure, with brief visual evidence. Return only JSON:
    {\ttfamily\char123{}\char34{}items\char34{}:[\char123{}\char34{}id\char34{}:\char34{}o01\char34{},\allowbreak\char34{}verdict\char34{}:\char34{}yes\char124{}no\char124{}unsure\char34{},\allowbreak
    \char34{}evidence\char34{}:\char34{}brief evidence\char34{}\char125{}]\char125{}.}
    Include every requested ID once.\par
    \smallskip\textbf{User message (example, turn 1)}\par
    Requested objects:\par
    {\ttfamily[\char123{}\char34{}id\char34{}:\char34{}o01\char34{}, \char34{}object\char34{}:\char34{}a strawberry\char34{}\char125{}]}\par
  \end{minipage}}
  \caption{\textbf{Object-presence evaluation prompt for Add \& Preserve.} The judge receives the current-turn crop and the cumulative list of requested objects, illustrated here for turn one.}
  \label{fig:supp-coa-prompt}
\end{figure}

\subsection{Add \& Preserve Evaluation Metrics}
\label{app:add-preserve-metrics}

\paragraph{Preservation outside the box.}
To measure cumulative non-target drift, we compare each output with the initial marked image using a fixed binary mask covering the exterior pixels, excluding the border. This provides a consistent reference across turns, regardless of whether the requested additions succeed. Since viewpoint changes or zooming would disrupt pixel correspondence, this evaluation is restricted to the localized additions in Add \& Preserve.

We compute PSNR from mean squared RGB error over the masked pixels, with intensities in $[0,1]$. For SSIM and LPIPS, we first replace the box interior and border in each output with the reference pixels, preventing requested edits from affecting neighborhoods across the boundary. We then average each spatial score map over the exterior pixels, using an $11\times11$ Gaussian window ($\sigma=1.5$) for SSIM and AlexNet v0.1 for LPIPS. We report the mean across images at each turn, with turn-10 results in the main table.

\paragraph{Object retention inside the box.}
To complement exterior preservation, Cumulative Object Accuracy (COA) measures how many objects requested up to each turn remain visible. We crop the editing region just inside its red border and provide the judge with this crop and the cumulative object list (\cref{fig:supp-coa-prompt}). At turn $r$, COA is the fraction of the $r$ requested objects receiving a \emph{yes} verdict, with \emph{no} and \emph{unsure} receiving zero credit. For example, if only the eraser remains at turn two, the score is $1/2$. We first average the ten turn scores within each complete sequence, then across sequences, giving each turn equal weight despite the increasing number of requested objects. Final score is reported as a percentage.

\section{Additional Quantitative Results}
\label{app:quantitative-results}

\begin{table}[t]
  \centering
    \caption{\textbf{Qwen-Image-Bench results by language and criterion.} Each language includes 1,000 prompts. Criterion scores average over applicable prompts; Overall averages applicable criteria within each prompt, then weights all prompts equally.}
  \label{tab:supp-qib-breakdown}
  \footnotesize
  \setlength{\tabcolsep}{1.2pt}
  \renewcommand{\arraystretch}{0.91}
  \resizebox{\linewidth}{!}{%
  \begin{tabular}{@{}l*{12}{r}@{}}
    \toprule
    & \multicolumn{4}{c}{\textbf{FLUX.2 [klein] Base 9B}} & \multicolumn{4}{c}{\textbf{SenseNova-U1.5-8B-MoT}} & \multicolumn{4}{c}{\textbf{Qwen-Image-2.1}} \\
    \cmidrule(lr){2-5}\cmidrule(lr){6-9}\cmidrule(lr){10-13}
    Criterion & Base & $k_s{=}43$ & $k_s{=}46$ & $k_s{=}48$ & Base & $k_s{=}37$ & $k_s{=}42$ & $k_s{=}47$ & Base & $k_s{=}24$ & $k_s{=}30$ & $k_s{=}35$ \\
    \midrule
    \multicolumn{13}{@{}l}{\emph{English prompts}} \\
    Quality & 49.51 & 49.94 & 50.06 & 49.93 & 51.31 & 50.54 & 50.13 & 50.50 & 54.41 & 54.11 & 54.30 & 54.09 \\
    Aesthetics & 52.26 & 52.58 & 52.35 & 53.12 & 55.50 & 54.45 & 54.14 & 54.49 & 54.20 & 55.01 & 54.17 & 54.57 \\
    Alignment & 50.58 & 50.28 & 50.66 & 51.18 & 54.91 & 53.85 & 53.94 & 53.83 & 53.65 & 52.94 & 52.84 & 52.88 \\
    Real-world fidelity & 44.68 & 44.96 & 43.20 & 44.19 & 48.86 & 48.39 & 48.17 & 48.42 & 51.11 & 51.95 & 51.03 & 51.58 \\
    Creative generation & 44.91 & 46.35 & 45.75 & 45.11 & 54.26 & 53.27 & 52.63 & 53.68 & 52.20 & 52.07 & 51.42 & 52.66 \\
    \cmidrule[0.25pt]{1-13}
    \textbf{Overall} & \textbf{49.27} & \textbf{49.68} & \textbf{49.42} & \textbf{49.74} & \textbf{53.43} & \textbf{52.45} & \textbf{52.24} & \textbf{52.55} & \textbf{53.39} & \textbf{53.49} & \textbf{53.06} & \textbf{53.32} \\
    \addlinespace[2pt]
    \multicolumn{13}{@{}l}{\emph{Chinese prompts}} \\
    Quality & 49.75 & 49.69 & 49.91 & 49.85 & 51.68 & 51.45 & 50.95 & 51.24 & 54.03 & 53.75 & 54.12 & 53.65 \\
    Aesthetics & 52.08 & 52.00 & 52.12 & 52.06 & 54.58 & 53.47 & 54.23 & 54.55 & 54.25 & 53.35 & 54.00 & 53.91 \\
    Alignment & 50.45 & 50.83 & 50.92 & 51.04 & 54.89 & 54.55 & 54.19 & 54.77 & 53.54 & 53.52 & 53.61 & 53.12 \\
    Real-world fidelity & 43.22 & 44.09 & 43.49 & 44.06 & 48.96 & 48.77 & 47.86 & 48.23 & 49.94 & 49.22 & 49.75 & 49.37 \\
    Creative generation & 44.88 & 46.19 & 44.91 & 45.03 & 53.93 & 53.89 & 53.48 & 54.07 & 53.06 & 52.57 & 52.30 & 53.47 \\
    \cmidrule[0.25pt]{1-13}
    \textbf{Overall} & \textbf{49.16} & \textbf{49.48} & \textbf{49.37} & \textbf{49.46} & \textbf{53.25} & \textbf{52.78} & \textbf{52.58} & \textbf{53.02} & \textbf{53.34} & \textbf{52.86} & \textbf{53.20} & \textbf{53.06} \\
    \bottomrule
  \end{tabular}%
  }
\end{table}

\begin{table}[t]
  \centering
    \caption{\textbf{BizGenEval results by design domain and prompt dimension.} The same 400 prompts are grouped by domain (80 each) and dimension (100 each). Entries report mean image-level Error Scores (\%, higher is better).}
  \label{tab:supp-bizgen-breakdown}
  \footnotesize
  \setlength{\tabcolsep}{1.2pt}
  \renewcommand{\arraystretch}{0.91}
  \begin{tabular*}{\textwidth}{@{\extracolsep{\fill}}l*{12}{r}@{}}
    \toprule
    & \multicolumn{4}{c}{\textbf{FLUX.2 [klein] Base 9B}} & \multicolumn{4}{c}{\textbf{SenseNova-U1.5-8B-MoT}} & \multicolumn{4}{c}{\textbf{Qwen-Image-2.1}} \\
    \cmidrule(lr){2-5}\cmidrule(lr){6-9}\cmidrule(lr){10-13}
    Subset & Base & $k_s{=}43$ & $k_s{=}46$ & $k_s{=}48$ & Base & $k_s{=}37$ & $k_s{=}42$ & $k_s{=}47$ & Base & $k_s{=}24$ & $k_s{=}30$ & $k_s{=}35$ \\
    \midrule
    \multicolumn{13}{@{}l}{\emph{Design domain}} \\
    Charts & 5.38 & 3.25 & 4.25 & 3.00 & 39.75 & 34.12 & 35.25 & 39.62 & 36.25 & 34.12 & 33.88 & 35.75 \\
    Posters & 11.75 & 12.25 & 15.62 & 13.50 & 47.62 & 47.50 & 49.50 & 49.00 & 61.88 & 60.00 & 58.75 & 59.12 \\
    Scientific figures & 1.00 & 1.62 & 2.37 & 1.62 & 47.50 & 49.50 & 49.12 & 50.00 & 59.00 & 61.75 & 62.12 & 60.38 \\
    Slides & 15.75 & 11.62 & 15.38 & 14.88 & 44.88 & 38.50 & 43.88 & 43.50 & 62.38 & 61.38 & 63.75 & 61.50 \\
    Webpages & 15.50 & 12.75 & 17.25 & 13.87 & 49.50 & 49.12 & 50.88 & 48.75 & 65.62 & 64.00 & 68.38 & 66.62 \\
    \addlinespace[2pt]
    \multicolumn{13}{@{}l}{\emph{Prompt dimension}} \\
    Attribute & 7.50 & 4.60 & 7.50 & 6.50 & 66.40 & 61.00 & 64.90 & 67.40 & 67.30 & 68.60 & 69.50 & 67.80 \\
    Layout & 30.00 & 25.90 & 34.50 & 28.90 & 76.00 & 75.00 & 79.00 & 78.10 & 77.40 & 76.70 & 74.90 & 77.50 \\
    Text & 1.70 & 2.00 & 1.90 & 2.00 & 35.40 & 35.10 & 34.50 & 35.30 & 77.60 & 77.90 & 80.00 & 78.60 \\
    Knowledge & 0.30 & 0.70 & 0.00 & 0.10 & 5.60 & 3.90 & 4.50 & 3.90 & 5.80 & 1.80 & 5.10 & 2.80 \\
    \cmidrule[0.25pt]{1-13}
    \textbf{Overall} & \textbf{9.87} & \textbf{8.30} & \textbf{10.97} & \textbf{9.37} & \textbf{45.85} & \textbf{43.75} & \textbf{45.73} & \textbf{46.17} & \textbf{57.03} & \textbf{56.25} & \textbf{57.38} & \textbf{56.67} \\
    \bottomrule
  \end{tabular*}
\end{table}

The following tables compare each base model with TailSteer at three steering onsets $k_s$, using the protocols in \cref{app:single-turn-protocols}. The single-turn editing tables also include FreqEdit and ConsLoRA for FLUX.2 [klein] Base 9B.

The text-to-image breakdowns in \cref{tab:supp-qib-breakdown,tab:supp-bizgen-breakdown} show that overall performance remains broadly close to the base models, with gains and losses across individual criteria and design tasks. The editing results in \cref{tab:supp-gedit-breakdown,tab:supp-rise-breakdown,tab:supp-imgedit-breakdown} reveal similar task-specific variation. These breakdowns clarify where capabilities remain stable and where aggregate scores can obscure task-specific changes.

\begin{table}[t]
  \centering
  \vspace{-10mm}
    \caption{\textbf{GEdit-Bench results by edit type and language.} Each language includes 606 edits, scored by the geometric mean of semantic consistency and perceptual quality (0--10). Overall weights the 11 edit types equally.}
  \label{tab:supp-gedit-breakdown}
  \footnotesize
  \setlength{\tabcolsep}{1.2pt}
  \renewcommand{\arraystretch}{0.91}
  \resizebox{\linewidth}{!}{%
  \begin{tabular}{@{}l*{14}{r}@{}}
    \toprule
    & \multicolumn{6}{c}{\textbf{FLUX.2 [klein] Base 9B}} & \multicolumn{4}{c}{\textbf{SenseNova-U1.5-8B-MoT}} & \multicolumn{4}{c}{\textbf{Qwen-Image-2.1}} \\
    \cmidrule(lr){2-7}\cmidrule(lr){8-11}\cmidrule(lr){12-15}
    Edit type & Base & FreqEdit & ConsLoRA & $k_s{=}43$ & $k_s{=}46$ & $k_s{=}48$ & Base & $k_s{=}37$ & $k_s{=}42$ & $k_s{=}47$ & Base & $k_s{=}24$ & $k_s{=}30$ & $k_s{=}35$ \\
    \midrule
    \multicolumn{15}{@{}l}{\emph{English prompts}} \\
    Subject addition & 8.31 & 8.16 & 8.17 & 8.35 & 8.33 & 8.32 & 8.61 & 8.86 & 8.71 & 8.84 & 8.95 & 8.83 & 8.94 & 8.89 \\
    Subject removal & 7.48 & 6.98 & 6.25 & 7.50 & 7.58 & 7.34 & 8.51 & 8.51 & 8.60 & 8.47 & 7.03 & 6.82 & 6.87 & 6.84 \\
    Subject replacement & 7.47 & 7.34 & 6.29 & 7.81 & 7.79 & 7.81 & 8.44 & 8.31 & 8.42 & 8.42 & 8.48 & 8.47 & 8.53 & 8.52 \\
    Background change & 7.91 & 7.72 & 6.03 & 7.63 & 7.50 & 7.41 & 8.47 & 8.22 & 8.31 & 8.25 & 8.39 & 8.26 & 8.36 & 8.20 \\
    Color change & 8.51 & 8.15 & 8.46 & 8.06 & 8.19 & 8.26 & 8.10 & 8.02 & 7.77 & 7.99 & 8.05 & 8.04 & 8.05 & 8.20 \\
    Material change & 7.33 & 7.09 & 6.22 & 6.93 & 7.21 & 7.23 & 7.94 & 7.83 & 7.78 & 7.78 & 7.82 & 7.66 & 7.56 & 7.64 \\
    Motion change & 8.09 & 7.56 & 6.15 & 7.92 & 7.95 & 7.77 & 8.68 & 8.67 & 8.50 & 8.47 & 8.37 & 8.36 & 8.23 & 8.27 \\
    Human editing & 5.57 & 5.79 & 5.48 & 5.64 & 5.82 & 5.81 & 8.19 & 8.08 & 8.17 & 8.15 & 7.18 & 6.62 & 7.33 & 7.04 \\
    Style change & 6.95 & 7.04 & 5.38 & 7.00 & 7.18 & 6.96 & 7.57 & 7.56 & 7.51 & 7.63 & 7.32 & 7.08 & 7.10 & 7.21 \\
    Text change & 7.03 & 5.88 & 5.60 & 7.04 & 7.40 & 7.26 & 8.91 & 8.64 & 8.86 & 8.90 & 8.95 & 8.91 & 8.88 & 8.80 \\
    Tone transfer & 7.53 & 7.45 & 8.34 & 7.80 & 7.74 & 7.32 & 8.27 & 8.56 & 8.39 & 8.39 & 8.63 & 8.49 & 8.62 & 8.57 \\
    \cmidrule[0.25pt]{1-15}
    \textbf{Overall} & \textbf{7.47} & \textbf{7.20} & \textbf{6.58} & \textbf{7.43} & \textbf{7.52} & \textbf{7.41} & \textbf{8.33} & \textbf{8.30} & \textbf{8.27} & \textbf{8.30} & \textbf{8.11} & \textbf{7.96} & \textbf{8.04} & \textbf{8.02} \\
    \addlinespace[2pt]
    \multicolumn{15}{@{}l}{\emph{Chinese prompts}} \\
    Subject addition & 8.36 & 8.38 & 7.84 & 8.30 & 8.19 & 8.25 & 9.04 & 8.97 & 9.00 & 8.92 & 8.62 & 8.72 & 8.76 & 8.64 \\
    Subject removal & 7.17 & 6.22 & 5.30 & 6.74 & 7.43 & 7.48 & 8.48 & 8.53 & 8.47 & 8.52 & 5.44 & 5.48 & 5.38 & 5.38 \\
    Subject replacement & 7.92 & 7.52 & 5.91 & 8.01 & 7.97 & 8.04 & 8.20 & 8.27 & 8.27 & 8.10 & 7.58 & 7.69 & 7.71 & 7.80 \\
    Background change & 7.60 & 7.33 & 5.16 & 7.54 & 7.52 & 7.31 & 8.32 & 7.99 & 8.13 & 8.07 & 7.61 & 7.53 & 7.67 & 7.62 \\
    Color change & 8.07 & 7.86 & 8.65 & 7.47 & 7.42 & 7.56 & 8.03 & 8.12 & 8.15 & 7.92 & 8.12 & 8.00 & 7.98 & 8.01 \\
    Material change & 7.25 & 7.37 & 6.68 & 7.06 & 7.23 & 7.00 & 7.75 & 7.61 & 8.04 & 7.65 & 7.94 & 8.14 & 7.96 & 7.80 \\
    Motion change & 7.49 & 7.68 & 6.05 & 7.73 & 7.76 & 8.18 & 8.49 & 8.61 & 8.37 & 8.48 & 8.51 & 8.25 & 8.37 & 8.36 \\
    Human editing & 6.14 & 5.88 & 5.59 & 5.96 & 5.97 & 6.15 & 8.27 & 8.23 & 8.29 & 8.24 & 7.23 & 6.66 & 6.59 & 6.88 \\
    Style change & 6.81 & 7.21 & 5.32 & 7.12 & 7.12 & 7.01 & 7.58 & 7.55 & 7.41 & 7.66 & 7.33 & 7.27 & 7.17 & 7.10 \\
    Text change & 7.18 & 5.75 & 5.72 & 7.24 & 6.96 & 7.04 & 8.74 & 8.73 & 8.79 & 8.83 & 8.60 & 8.73 & 8.62 & 8.84 \\
    Tone transfer & 7.67 & 7.72 & 8.54 & 7.71 & 7.68 & 7.78 & 8.41 & 8.46 & 8.46 & 8.37 & 8.86 & 8.55 & 8.67 & 8.82 \\
    \cmidrule[0.25pt]{1-15}
    \textbf{Overall} & \textbf{7.42} & \textbf{7.18} & \textbf{6.43} & \textbf{7.35} & \textbf{7.39} & \textbf{7.44} & \textbf{8.30} & \textbf{8.28} & \textbf{8.31} & \textbf{8.25} & \textbf{7.80} & \textbf{7.73} & \textbf{7.72} & \textbf{7.75} \\
    \bottomrule
  \end{tabular}%
  }
\end{table}

\begin{table}[t]
  \centering
    \caption{\textbf{RISEBench results by reasoning type.} Accuracy (\%) requires full marks on all applicable criteria, with sample counts in parentheses. Overall pools all 360 edits; the final three rows report mean rubric scores (1--5).}
  \label{tab:supp-rise-breakdown}
  \footnotesize
  \setlength{\tabcolsep}{1.2pt}
  \renewcommand{\arraystretch}{0.91}
  \resizebox{\linewidth}{!}{%
  \begin{tabular}{@{}l*{14}{r}@{}}
    \toprule
    & \multicolumn{6}{c}{\textbf{FLUX.2 [klein] Base 9B}} & \multicolumn{4}{c}{\textbf{SenseNova-U1.5-8B-MoT}} & \multicolumn{4}{c}{\textbf{Qwen-Image-2.1}} \\
    \cmidrule(lr){2-7}\cmidrule(lr){8-11}\cmidrule(lr){12-15}
    Measure & Base & FreqEdit & ConsLoRA & $k_s{=}43$ & $k_s{=}46$ & $k_s{=}48$ & Base & $k_s{=}37$ & $k_s{=}42$ & $k_s{=}47$ & Base & $k_s{=}24$ & $k_s{=}30$ & $k_s{=}35$ \\
    \midrule
    \multicolumn{15}{@{}l}{\emph{Accuracy (\%)}} \\
    Temporal (85) & 8.24 & 5.88 & 7.06 & 7.06 & 8.24 & 8.24 & 24.71 & 29.41 & 22.35 & 24.71 & 14.12 & 11.76 & 10.59 & 10.59 \\
    Causal (90) & 14.44 & 10.00 & 8.89 & 14.44 & 17.78 & 13.33 & 36.67 & 41.11 & 37.78 & 40.00 & 27.78 & 26.67 & 23.33 & 23.33 \\
    Spatial (100) & 28.00 & 23.00 & 16.00 & 27.00 & 23.00 & 21.00 & 46.00 & 43.00 & 43.00 & 44.00 & 30.00 & 27.00 & 29.00 & 31.00 \\
    Logical (85) & 3.53 & 3.53 & 4.71 & 2.35 & 3.53 & 1.18 & 15.29 & 10.59 & 11.76 & 15.29 & 8.24 & 5.88 & 7.06 & 7.06 \\
    \cmidrule[0.25pt]{1-15}
    \textbf{Overall} & \textbf{14.17} & \textbf{11.11} & \textbf{9.44} & \textbf{13.33} & \textbf{13.61} & \textbf{11.39} & \textbf{31.39} & \textbf{31.67} & \textbf{29.44} & \textbf{31.67} & \textbf{20.56} & \textbf{18.33} & \textbf{18.06} & \textbf{18.61} \\
    \addlinespace[2pt]
    \multicolumn{15}{@{}l}{\emph{Rubric scores (1--5)}} \\
    Instruction reasoning & 2.58 & 2.44 & 2.20 & 2.55 & 2.59 & 2.51 & 3.39 & 3.39 & 3.33 & 3.34 & 2.67 & 2.62 & 2.59 & 2.60 \\
    Appearance consistency & 3.68 & 4.04 & 4.28 & 3.65 & 3.78 & 3.76 & 4.54 & 4.53 & 4.56 & 4.57 & 4.45 & 4.45 & 4.42 & 4.46 \\
    Visual plausibility & 4.71 & 4.63 & 4.78 & 4.67 & 4.69 & 4.67 & 4.56 & 4.51 & 4.54 & 4.56 & 4.73 & 4.71 & 4.73 & 4.72 \\
    \bottomrule
  \end{tabular}%
  }
\end{table}

\begin{table}[!t]
  \centering
  \caption{\textbf{ImgEdit-Bench Basic results by edit type.} Grouped into nine types, each reported as a mean score on a 1--5 scale.}
  \label{tab:supp-imgedit-breakdown}
  \footnotesize
  \setlength{\tabcolsep}{1.2pt}
  \renewcommand{\arraystretch}{0.91}
  \resizebox{\linewidth}{!}{%
  \begin{tabular}{@{}l*{14}{r}@{}}
    \toprule
    & \multicolumn{6}{c}{\textbf{FLUX.2 [klein] Base 9B}} & \multicolumn{4}{c}{\textbf{SenseNova-U1.5-8B-MoT}} & \multicolumn{4}{c}{\textbf{Qwen-Image-2.1}} \\
    \cmidrule(lr){2-7}\cmidrule(lr){8-11}\cmidrule(lr){12-15}
    Edit type & Base & FreqEdit & ConsLoRA & $k_s{=}43$ & $k_s{=}46$ & $k_s{=}48$ & Base & $k_s{=}37$ & $k_s{=}42$ & $k_s{=}47$ & Base & $k_s{=}24$ & $k_s{=}30$ & $k_s{=}35$ \\
    \midrule
    Add & 4.53 & 4.47 & 4.57 & 4.56 & 4.54 & 4.53 & 4.49 & 4.52 & 4.55 & 4.54 & 4.76 & 4.74 & 4.67 & 4.67 \\
    Remove & 4.12 & 3.76 & 3.08 & 4.07 & 3.98 & 4.34 & 4.39 & 4.10 & 4.35 & 4.13 & 4.55 & 4.52 & 4.57 & 4.59 \\
    Replace & 4.51 & 4.44 & 3.76 & 4.55 & 4.54 & 4.56 & 4.84 & 4.72 & 4.73 & 4.69 & 4.67 & 4.70 & 4.71 & 4.68 \\
    Extract & 2.54 & 2.34 & 1.93 & 2.29 & 2.49 & 2.42 & 4.11 & 4.18 & 4.08 & 4.06 & 3.61 & 3.53 & 3.48 & 3.53 \\
    Adjust & 4.38 & 4.18 & 4.64 & 4.18 & 4.32 & 4.46 & 4.53 & 4.59 & 4.54 & 4.60 & 4.59 & 4.68 & 4.65 & 4.65 \\
    Background & 4.37 & 4.34 & 4.28 & 4.32 & 4.30 & 4.31 & 4.50 & 4.32 & 4.36 & 4.36 & 4.56 & 4.41 & 4.49 & 4.46 \\
    Style & 4.75 & 4.83 & 3.97 & 4.82 & 4.77 & 4.80 & 4.77 & 4.86 & 4.82 & 4.86 & 4.81 & 4.72 & 4.79 & 4.82 \\
    Compose & 3.80 & 3.67 & 3.77 & 3.85 & 3.84 & 3.82 & 3.77 & 3.74 & 3.65 & 3.74 & 3.87 & 3.61 & 4.10 & 3.87 \\
    Action & 4.82 & 4.75 & 4.73 & 4.91 & 4.89 & 4.89 & 4.49 & 4.44 & 4.60 & 4.50 & 4.21 & 4.32 & 4.33 & 4.24 \\
    \cmidrule[0.25pt]{1-15}
    \textbf{Overall} & \textbf{4.20} & \textbf{4.09} & \textbf{3.86} & \textbf{4.17} & \textbf{4.19} & \textbf{4.24} & \textbf{4.43} & \textbf{4.39} & \textbf{4.41} & \textbf{4.39} & \textbf{4.40} & \textbf{4.36} & \textbf{4.42} & \textbf{4.39} \\
    \bottomrule
  \end{tabular}%
  }
\end{table}

\clearpage
\section{Additional Qualitative Results}
\label{app:qualitative-results}

\subsection{Text-to-Image Generation}
\label{app:qualitative-t2i}\label{app:text-to-image-generation}

\begin{figure}[p]
  \centering
    \makebox[\linewidth][c]{%
      \includegraphics[
        width=1.25\linewidth,
        keepaspectratio
      ]{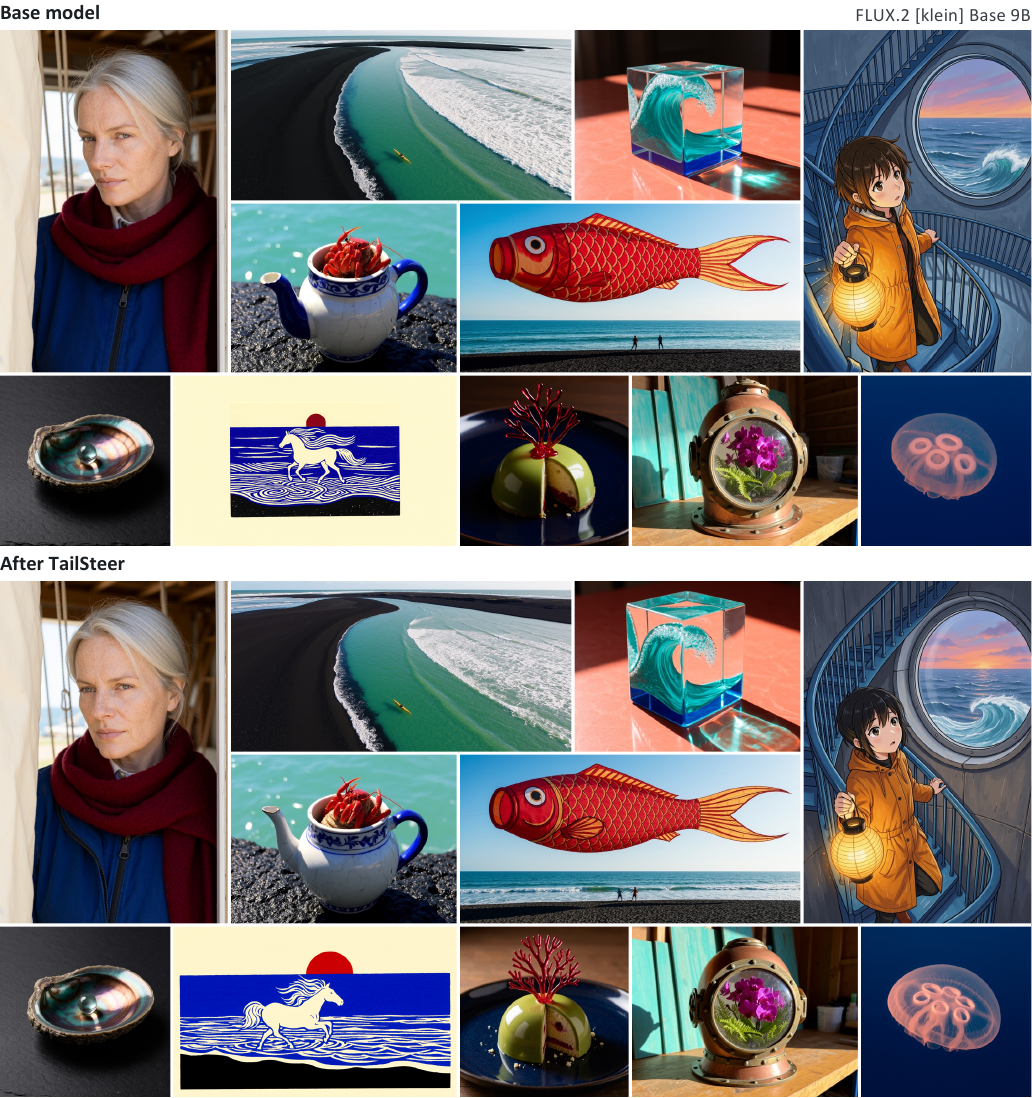}%
    }\par
    \caption{\textbf{Preserving text-to-image capabilities after TailSteer.} FLUX.2 [klein] Base 9B (top) and TailSteer ($k_s=46$, bottom) generate images using identical prompts, seeds, and sampling settings. Across diverse photographic and illustrative styles, TailSteer largely retains the base model's compositions and visual characteristics, with some results showing finer details, such as reflections in the glass cube, the teapot, and the intricate horse print and dessert garnish.}
  \label{fig:supp-FLUX2-klein-T2I}\label{fig:supp-t2i-klein}
\end{figure}

\begin{figure}[p]
  \centering
      \makebox[\linewidth][c]{%
      \includegraphics[
        width=1.25\linewidth,
        keepaspectratio
      ]{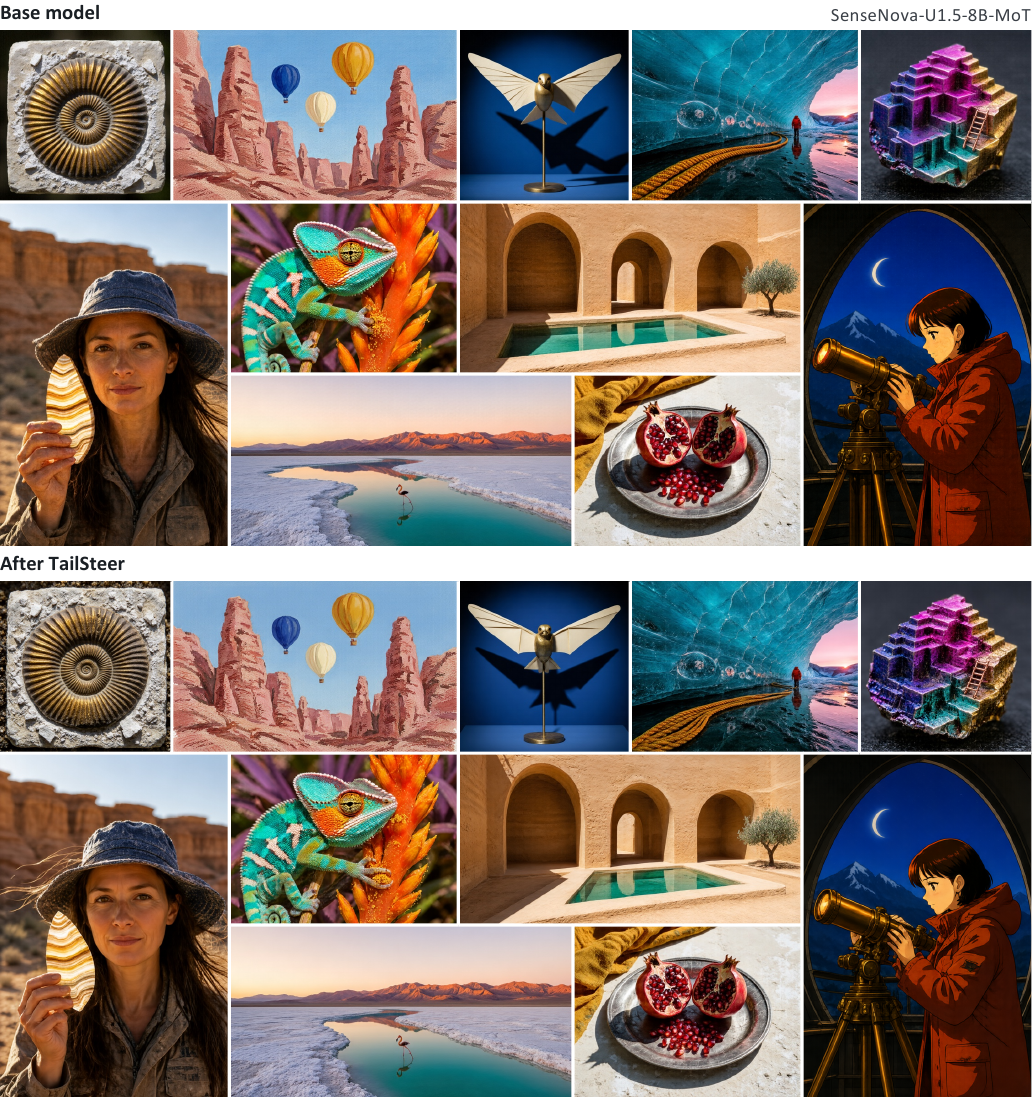}%
    }\par
    \caption{\textbf{Preserving text-to-image capabilities after TailSteer.} SenseNova-U1.5-8B-MoT (top) and TailSteer ($k_s=42$, bottom) generate images with identical prompts, seeds, and sampling settings. TailSteer closely retains the base model's compositions and visual styles, while often rendering finer details, including the winged sculpture and its shadow, ice-wall textures, crystalline structures, facial features, earthen architecture, and the pomegranate plate and tabletop.}
  \label{fig:supp-SenseNova-U1-5-T2I}\label{fig:supp-t2i-sensenova}
\end{figure}

\begin{figure}[p]
  \centering
    \makebox[\linewidth][c]{%
      \includegraphics[
        width=1.25\linewidth,
        keepaspectratio
      ]{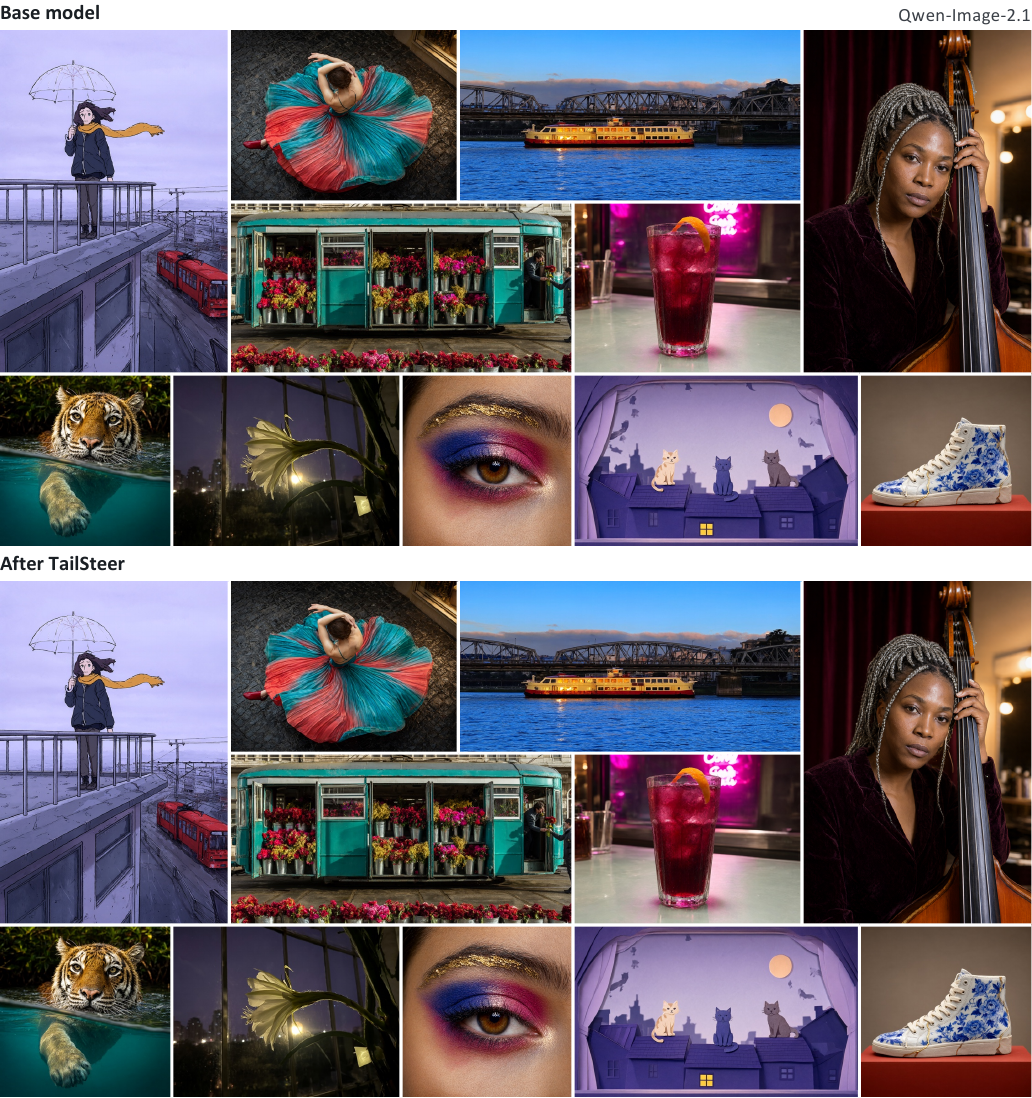}%
    }\par
    \caption{\textbf{Preserving text-to-image capabilities after TailSteer.} Qwen-Image-2.1 (top) and TailSteer ($k_s=30$, bottom) generate images with identical prompts, seeds, and sampling settings. TailSteer closely preserves the base model's compositions and visual styles, with some results showing richer fine-grained details, particularly in the flowing dress, water ripples, floral textures, and facial features.}
  \label{fig:supp-Qwen-Image-2-1-T2I-T2I}\label{fig:supp-t2i-qwen}
\end{figure}

To examine whether TailSteer affects the base models' generative capabilities, we compare their outputs before and after training using eleven English prompts per model, with identical seeds, sampling schedules, and image dimensions. As shown in \cref{fig:supp-t2i-klein,fig:supp-t2i-sensenova,fig:supp-t2i-qwen}, the refined models closely retain the original compositions, colors, and styles, occasionally producing even finer local details. Together with the quantitative results in \cref{tab:supp-qib-breakdown,tab:supp-bizgen-breakdown}, these examples show that TailSteer's preservation gains come with little disruption to the base models' learned generative priors.

\subsection{Single-Turn Image Editing}
\label{app:qualitative-single-turn}\label{app:single-turn-image-editing}

\begin{figure}[p]
  \centering
  \vspace{-10mm}
    \makebox[\linewidth][c]{%
      \includegraphics[
        width=1.1\linewidth,
        keepaspectratio
      ]{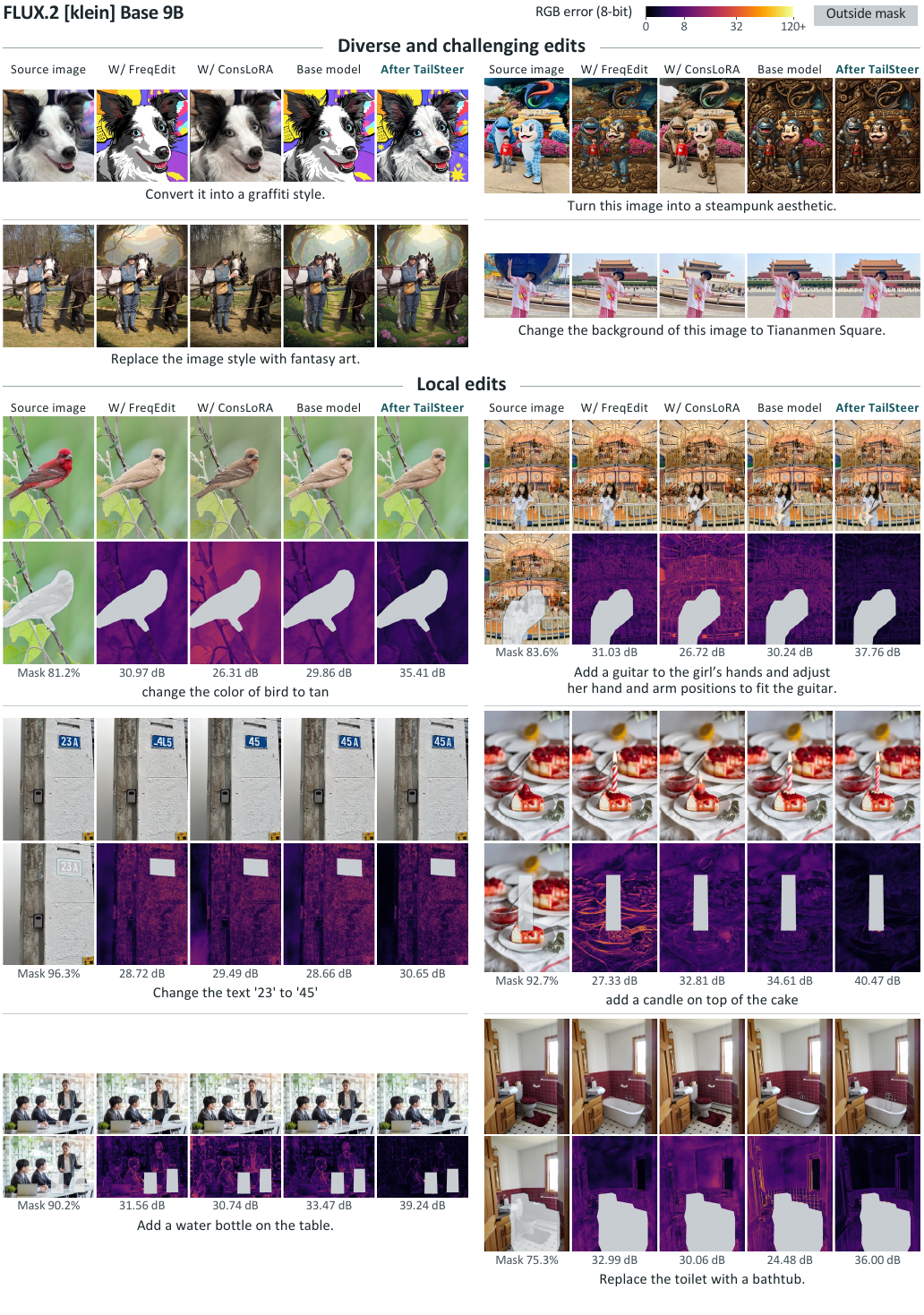}%
    }\par
    \vspace{-6mm}
    \caption{\textbf{Preserving single-turn editing capabilities with improved fidelity after TailSteer.} On GEdit-Bench (English), TailSteer ($k_s=46$) closely matches FLUX.2 [klein] Base 9B on challenging style transformations and semantic edits (top), sometimes rendering richer details. In contrast, FreqEdit and ConsLoRA often produce incomplete or incorrect edits, compromising the base model's learned capabilities. For local edits (bottom), RGB error maps reveal substantially fewer unintended changes in non-target regions. TailSteer achieves the highest preservation PSNR across all six examples, exceeding 40\,dB while faithfully executing the requested edits.}
  \label{fig:supp-gedit-klein}
\end{figure}

\begin{figure}[p]
  \centering
    \makebox[\linewidth][c]{%
      \includegraphics[
        width=1.25\linewidth,
        keepaspectratio
      ]{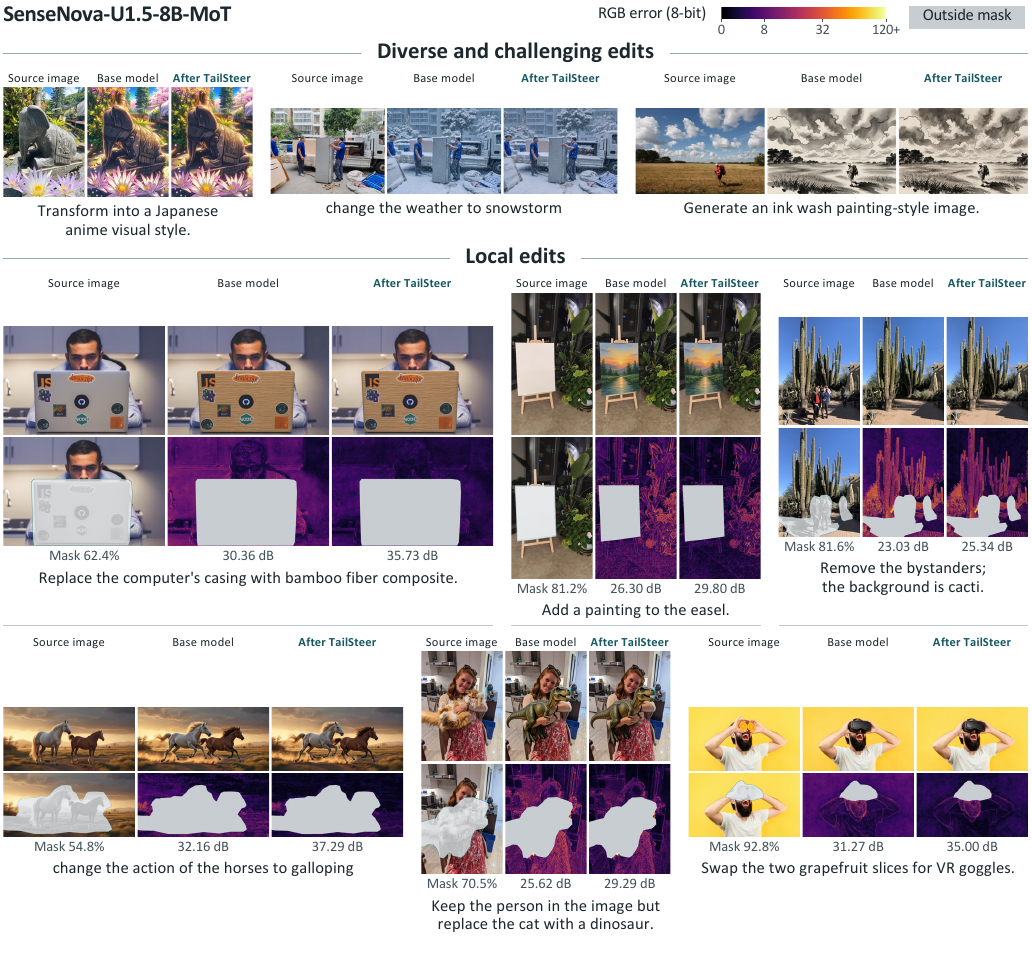}%
    }\par
    \caption{\textbf{Preserving single-turn editing capabilities with improved fidelity after TailSteer.} On GEdit-Bench (English), TailSteer ($k_s=42$) closely retains SenseNova-U1.5-8B-MoT's editing behavior across challenging style and weather transformations (top), maintaining its learned editing priors and visual quality. For local edits (bottom), TailSteer faithfully executes the requested changes while substantially reducing RGB errors in non-target regions. All six examples achieve higher preservation PSNR than the base model, reaching 37\,dB, with finer details retained in subjects and surrounding scenes.}
  \label{fig:supp-gedit-sensenova}
\end{figure}

\begin{figure}[p]
  \centering
      \makebox[\linewidth][c]{%
      \includegraphics[
        width=1.25\linewidth,
        keepaspectratio
      ]{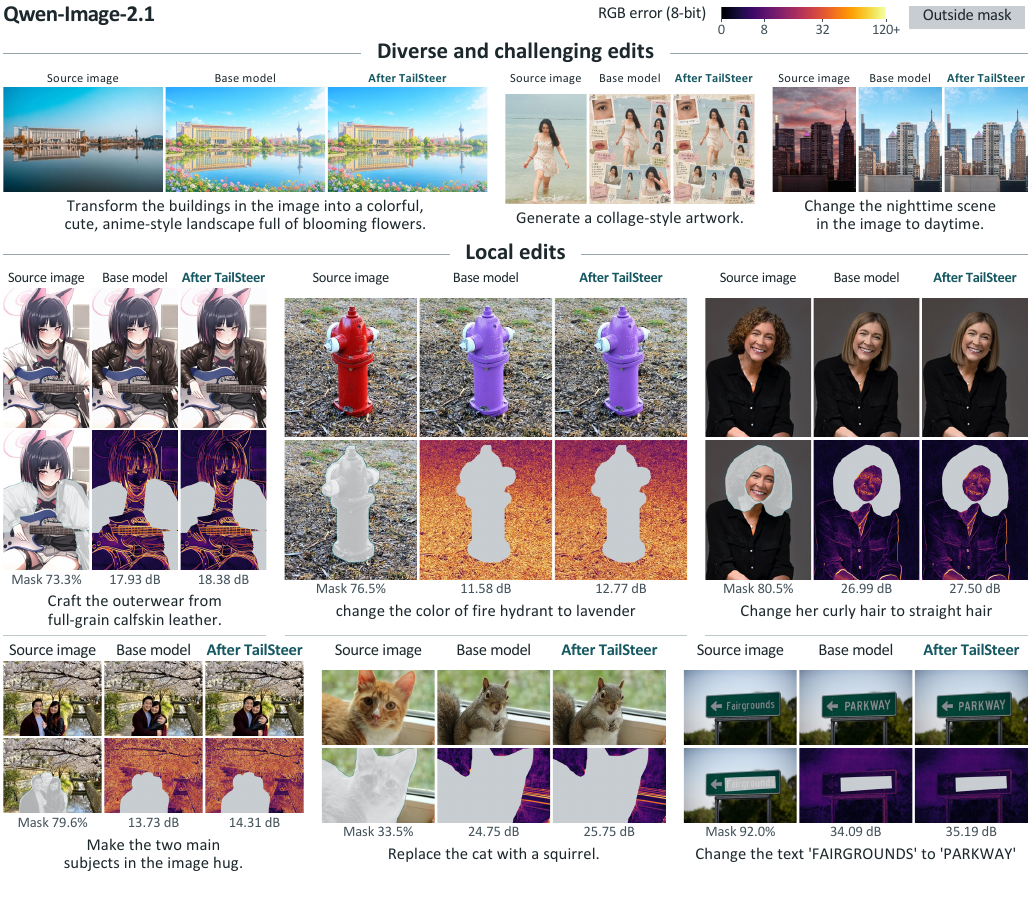}%
    }\par
    \caption{\textbf{Preserving single-turn editing capabilities with improved fidelity after TailSteer.} On GEdit-Bench (English), TailSteer ($k_s=30$) closely retains Qwen-Image-2.1's editing behavior across challenging style transformations, collage generation, and lighting changes (top). For local edits (bottom), TailSteer faithfully executes the requested changes while reducing RGB errors in non-target regions, better preserving facial details, background textures, and scene structures. All six examples achieve higher PSNR in the preservation region than the base model.}
  \label{fig:supp-gedit-qwen}
\end{figure}

We further examine whether TailSteer retains single-turn editing capabilities using matched GEdit-Bench~\citep{liu2025step1xedit} inputs, instructions, and seeds (\cref{fig:supp-gedit-klein,fig:supp-gedit-sensenova,fig:supp-gedit-qwen}). Across challenging non-local edits, TailSteer closely matches the base models' editing behavior, sometimes rendering finer details, whereas FreqEdit and ConsLoRA struggle with several edits on Klein. For local edits, RGB error maps reveal substantially fewer unintended changes, with higher preservation PSNR in every example shown. Together with \cref{tab:supp-gedit-breakdown}, these results show that TailSteer retains learned editing priors while improving pixel-level fidelity, laying the foundation for drift-resistant multi-turn editing.

\subsection{Single-Input Multi-Turn Editing}
\label{app:qualitative-single-input}\label{app:single-input-multi-turn-editing}
\begin{figure}[p]
  \centering
  \vspace{-10mm}
    \makebox[\linewidth][c]{%
      \includegraphics[
        width=1.25\linewidth,
        keepaspectratio
      ]{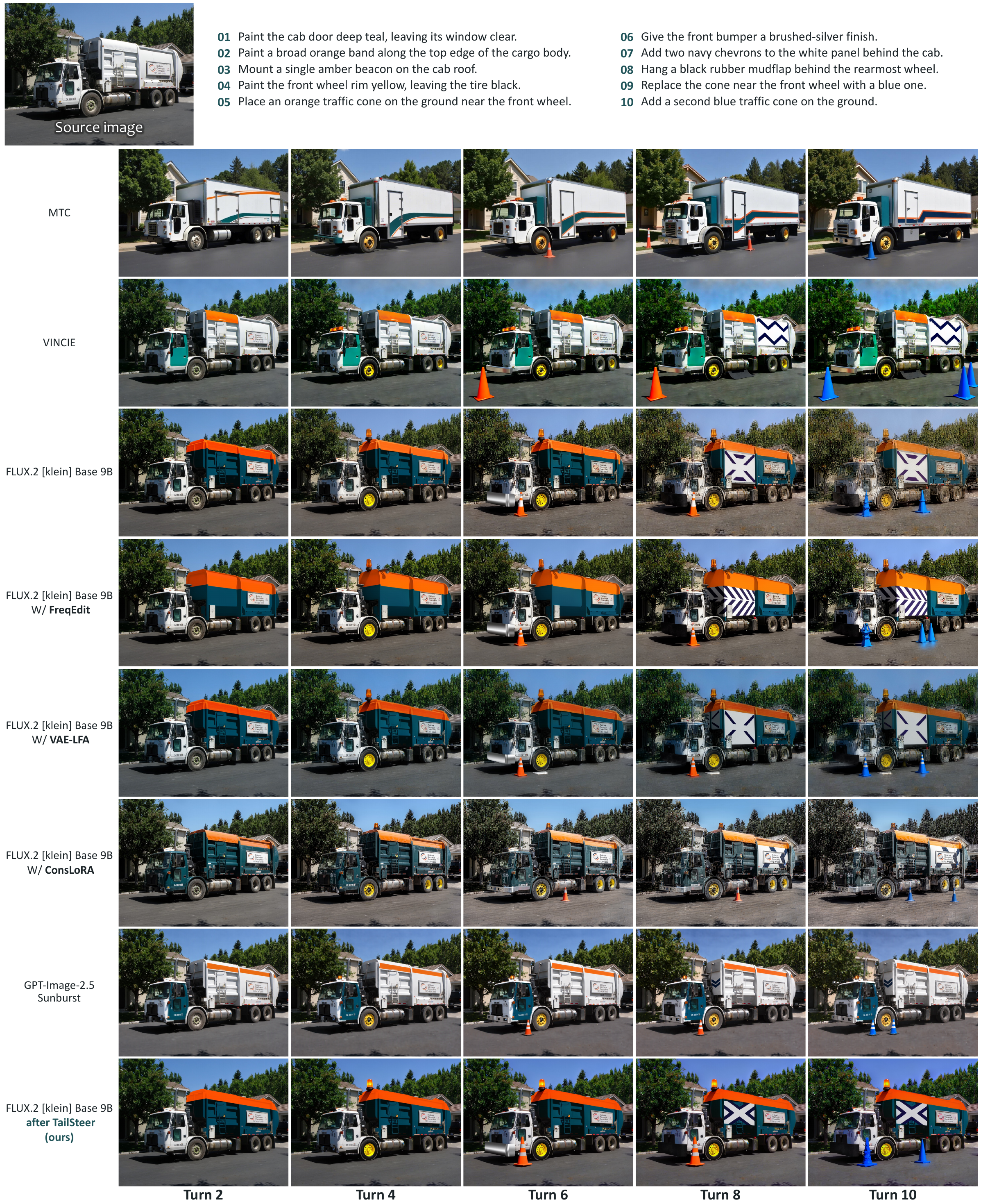}%
    }\par

    \caption{\textbf{Unlocking drift-resistant multi-turn editing after TailSteer: Local edits.} TailSteer faithfully executes all ten editing instructions while retaining the base model's fine-grained editing capabilities. During the initial recoloring, FreqEdit smooths away details on the truck body, whereas TailSteer closely matches the detailed result of the base model. Even after ten successive edits, TailSteer preserves the truck cab, distant houses, sky, trees, and pavement with remarkable fidelity, demonstrating substantially reduced cumulative drift without compromising editing quality.}

  \label{fig:supp-single-input-l1}
\end{figure}

\begin{figure}[p]
  \centering
    \vspace{-10mm}
    \makebox[\linewidth][c]{%
      \includegraphics[
        width=1.25\linewidth,
        keepaspectratio
      ]{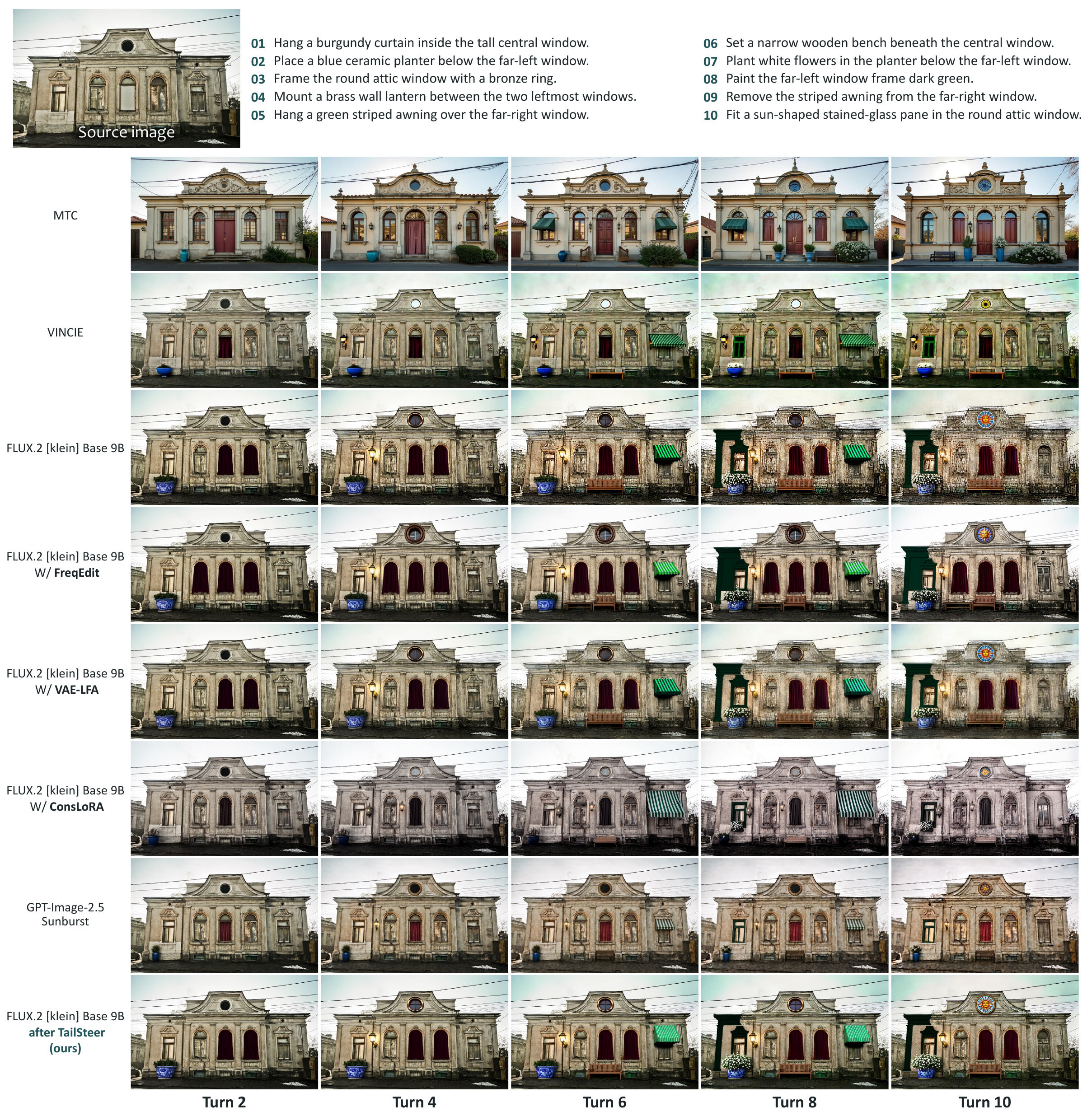}%
    }\par
    \caption{\textbf{Unlocking drift-resistant multi-turn editing after TailSteer: Local edits.} Across ten successive edits, TailSteer retains the base model's editing capabilities while preserving the original sky tones, weathered stone textures, and fine architectural details. Compared with the baselines, it exhibits substantially less color drift and texture degradation, keeping previously added elements, including the lantern, flower-filled blue planter, and wooden bench, clearly defined. The final two edits successfully remove the striped awning and introduce the sun-shaped stained-glass pane without compromising the overall fidelity of the scene.}

  \label{fig:supp-single-input-l2}
\end{figure}

\begin{figure}[p]
  \centering
      \vspace{-10mm}
    \makebox[\linewidth][c]{%
      \includegraphics[
        width=1.25\linewidth,
        keepaspectratio
      ]{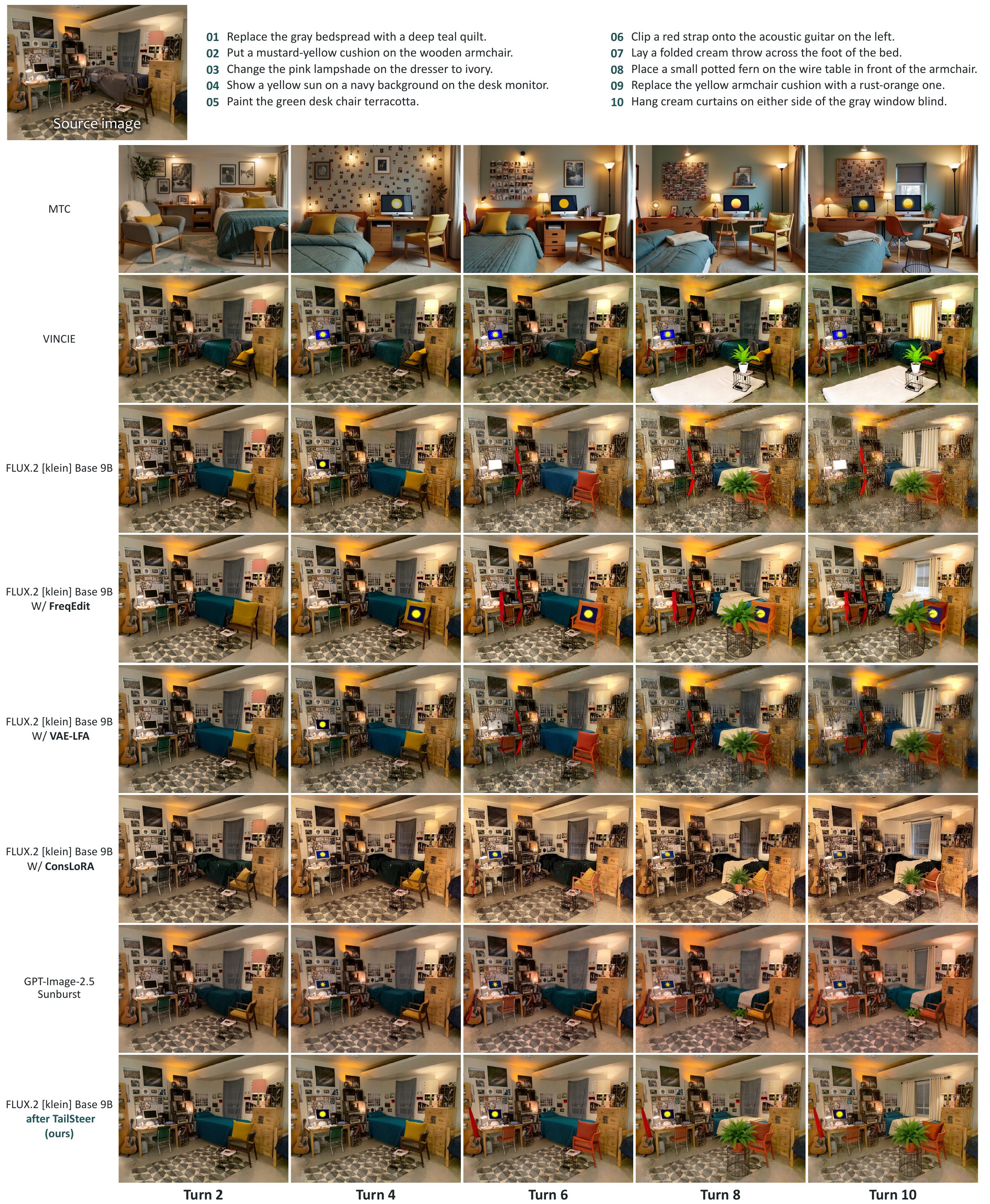}%
    }\par

    \caption{\textbf{Unlocking drift-resistant multi-turn editing after TailSteer: Local edits.} TailSteer follows the successive editing instructions while retaining the room's original color palette, ambient lighting, intricate rug patterns, wall photographs, and small furnishings. In contrast, several baselines introduce severe texture degradation, color shifts, or misplaced content, such as FreqEdit transferring the sun motif from the monitor to the armchair. Even at turn~10, TailSteer naturally integrates the newly added objects and decorative elements into the original scene, maintaining both editing quality and fine-grained visual fidelity.}

  \label{fig:supp-single-input-l3}
\end{figure}

\begin{figure}[p]
  \centering
      \vspace{-10mm}
    \makebox[\linewidth][c]{%
      \includegraphics[
        width=1.25\linewidth,
        keepaspectratio
      ]{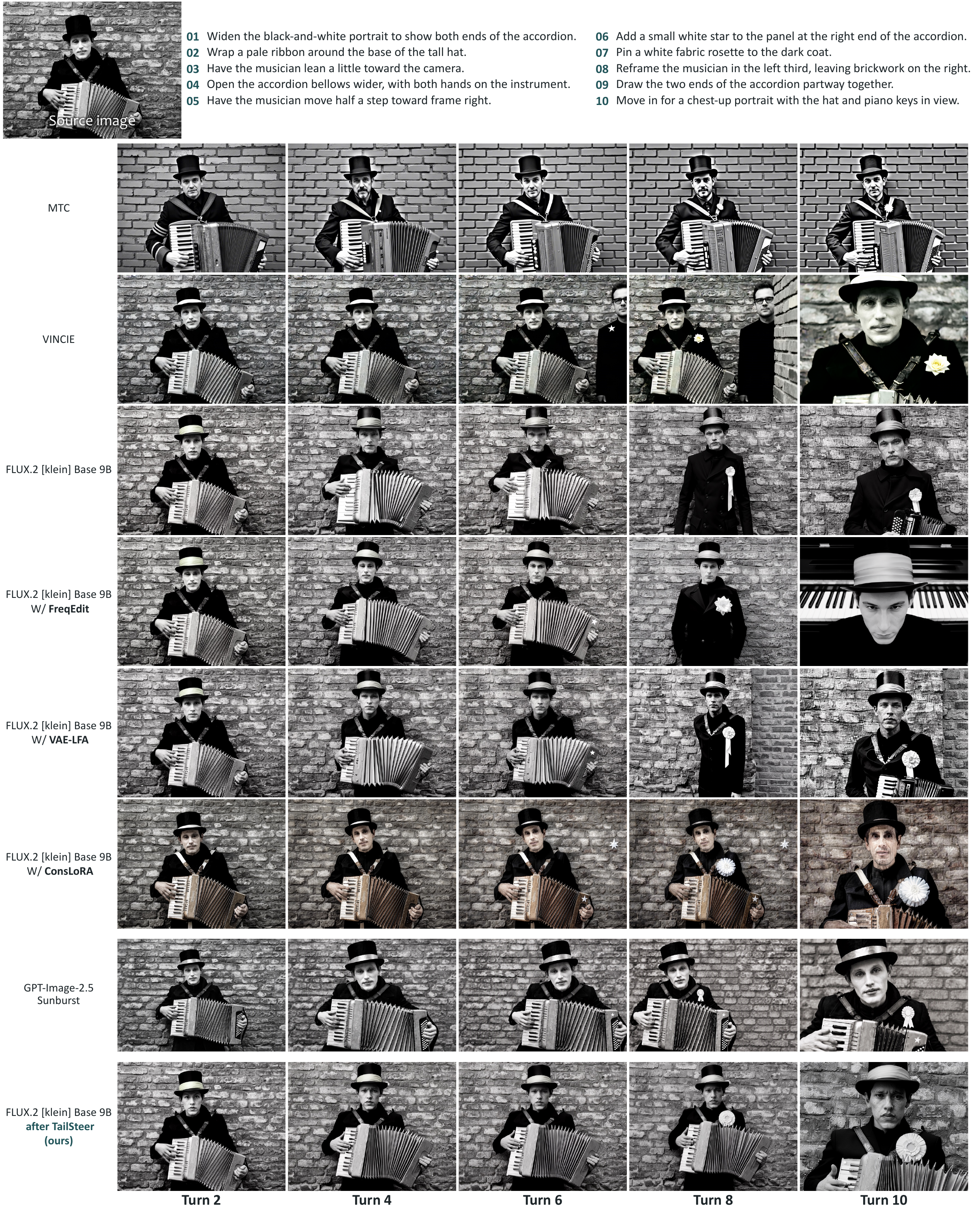}%
    }\par
    \vspace{-2mm}
    \caption{\textbf{Unlocking drift-resistant multi-turn editing after TailSteer: Identity-preserving edits.} Across ten turns combining local edits, non-rigid pose changes, and camera reframing, TailSteer preserves the musician's facial features, the accordion, and the fine textures of the brick wall. In contrast, several baselines introduce unintended poses or lose the instrument by turn~8. Such accumulated drift can also disrupt subsequent edits, as illustrated by FreqEdit's erroneous final close-up with a piano keyboard in the background. TailSteer successfully performs the requested zoom-in at turn~10, retaining clear facial features and instrument details without compromising fidelity.}

  \label{fig:supp-single-input-i2}
\end{figure}

\begin{figure}[p]
  \centering
      \vspace{-12mm}
    \makebox[\linewidth][c]{%
      \includegraphics[
        width=\linewidth,
        keepaspectratio
      ]{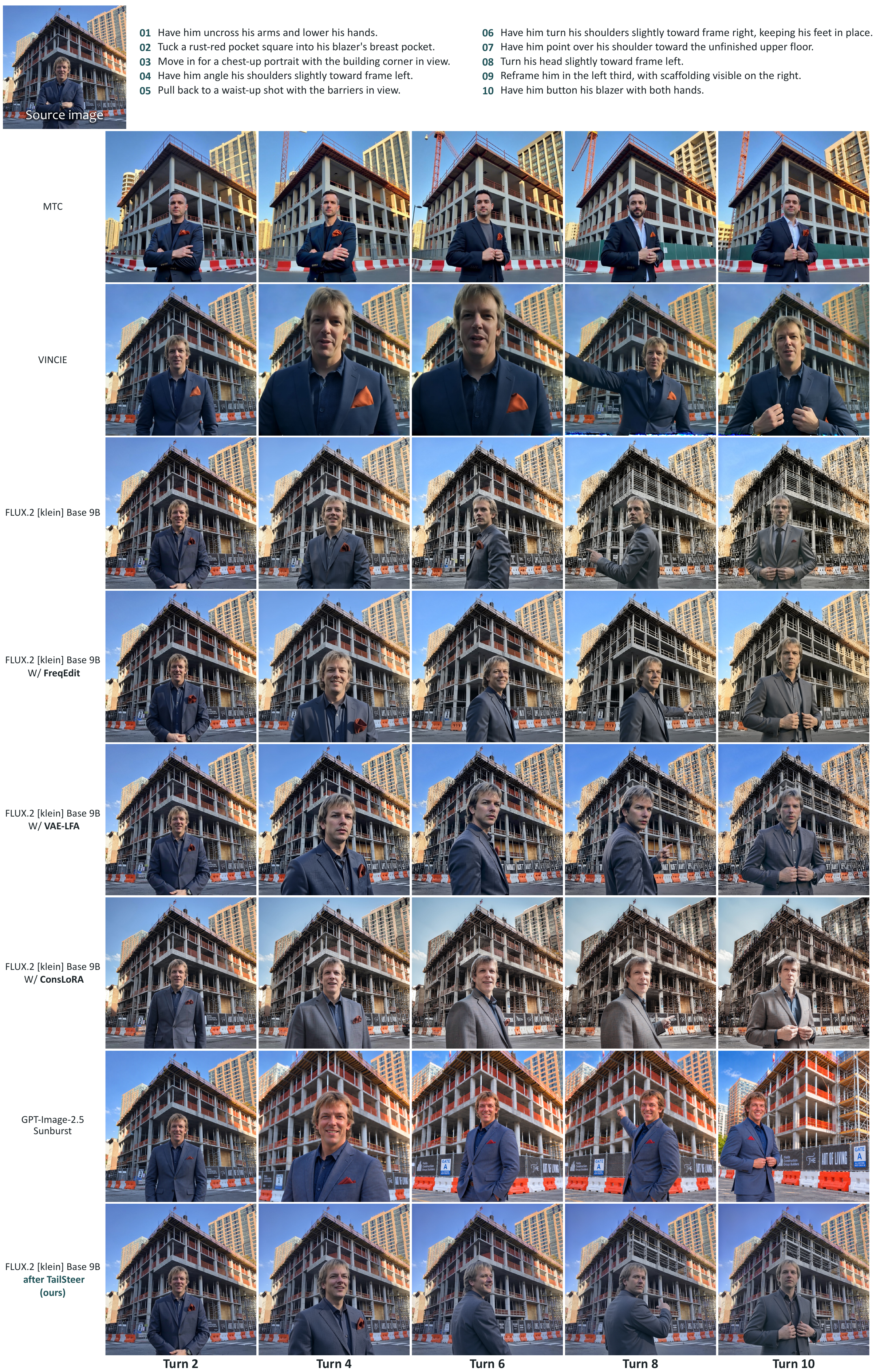}%
    }\par
    \vspace{-2mm}
    \caption{\textbf{Unlocking drift-resistant multi-turn editing after TailSteer: Identity-preserving edits.} Across ten turns combining local edits, non-rigid pose changes, and camera reframing, TailSteer preserves the man's facial features, natural skin tones, and fine architectural details. In contrast, several baselines introduce unintended poses, reddish skin tones, painterly artifacts, and progressive degradation of the building and scaffolding. TailSteer maintains the original photographic appearance while faithfully following the requested edits, from pointing toward the unfinished upper floor to buttoning the blazer at turn~10, demonstrating strong identity and scene preservation without compromising editing capabilities.}

  \label{fig:supp-single-input-i3}
\end{figure}

\begin{figure}[p]
  \centering
    \vspace{-10mm}
    \makebox[\linewidth][c]{%
      \includegraphics[
        width=1.25\linewidth,
        keepaspectratio
      ]{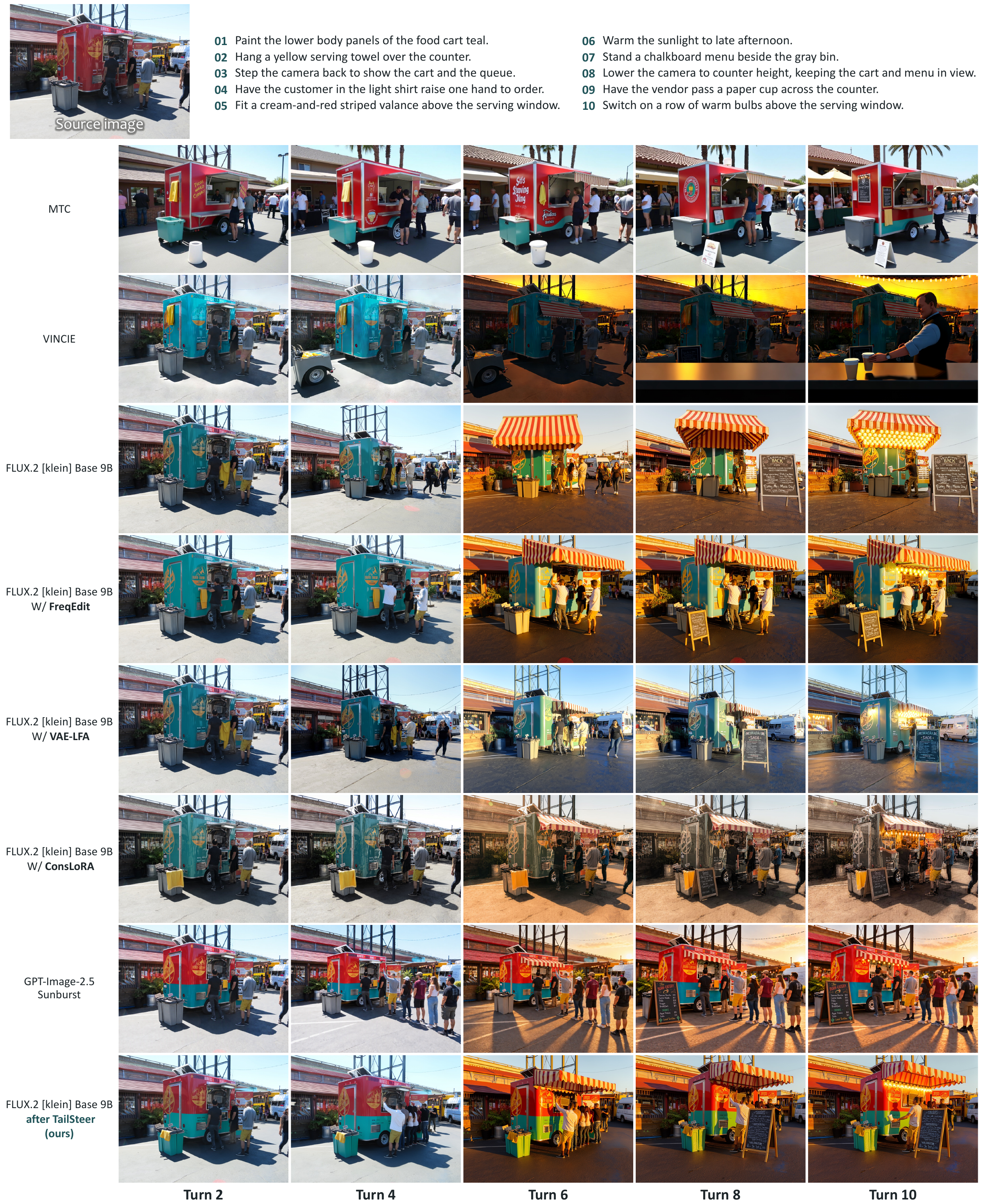}%
    }\par
    \vspace{-2mm}
    \caption{\textbf{Unlocking drift-resistant multi-turn editing after TailSteer: General edits.} Across ten challenging edits involving local modifications, lighting changes, human actions, and camera movements, TailSteer retains the base model's editing capabilities while preserving the food cart's structure, decorative details, and surrounding scene. As drift accumulates, several baselines exhibit progressive texture degradation and increasingly implausible edits, such as the obstructing foreground counter introduced by VINCIE. In contrast, TailSteer naturally integrates the striped awning, chalkboard, and warm lighting into the scene, maintaining fine-grained details and visual coherence through turn~10 despite substantial changes in appearance and viewpoint.}

  \label{fig:supp-single-input-m1}
\end{figure}

\begin{figure}[p]
  \centering
    \vspace{-10mm}
    \makebox[\linewidth][c]{%
      \includegraphics[
        width=1.25\linewidth,
        keepaspectratio
      ]{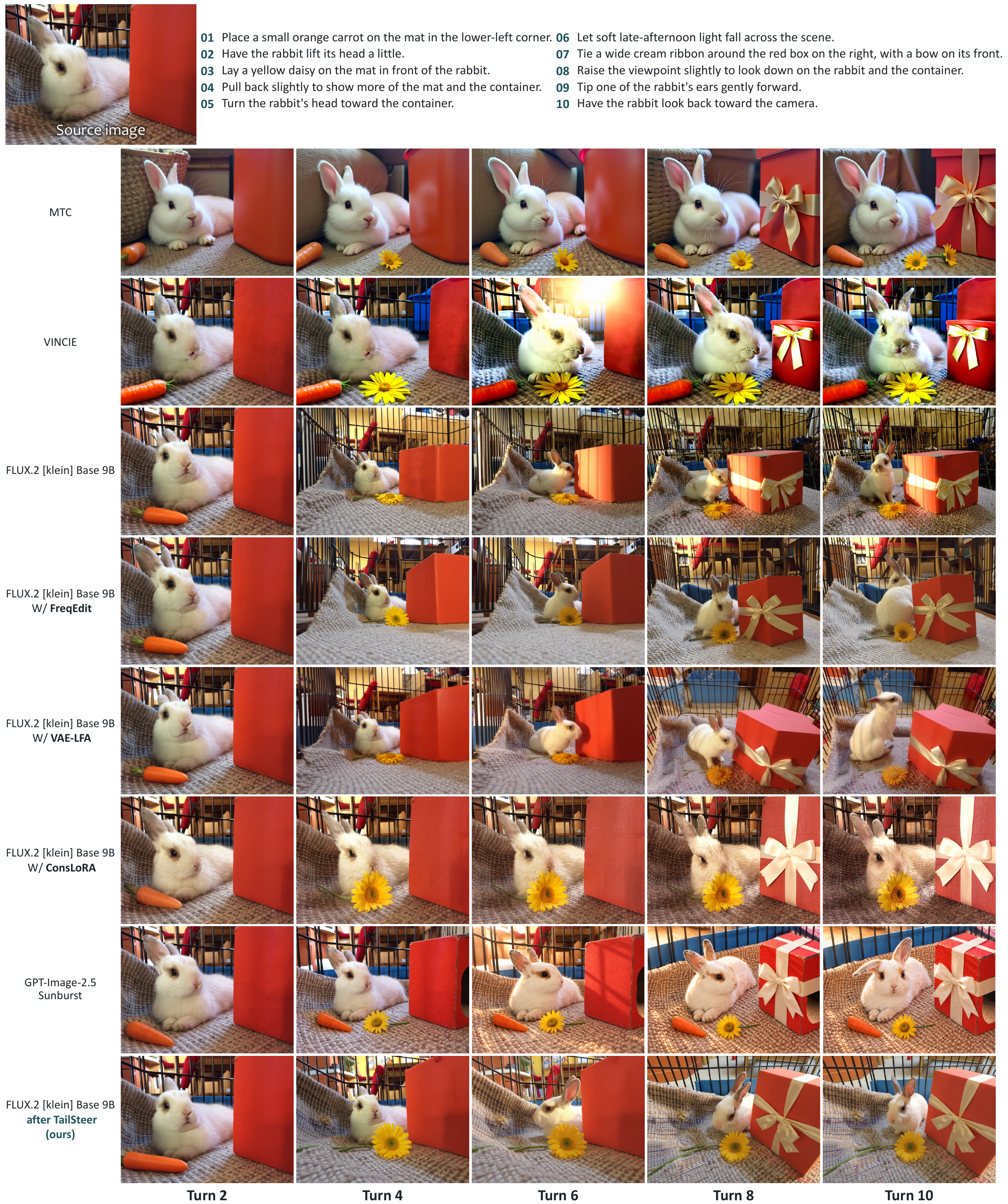}%
    }\par

    \caption{\textbf{Unlocking drift-resistant multi-turn editing after TailSteer: General edits.} Across ten challenging edits involving object additions, non-rigid pose changes, lighting adjustments, and camera movements, TailSteer retains the base model's editing capabilities while preserving the rabbit's distinctive appearance and the fine textures of the woven mat. In contrast, several baselines progressively distort the rabbit's features and introduce unnatural colors and textures into the surrounding scene. Even after substantial changes in pose and viewpoint, TailSteer maintains a natural photographic appearance and visual fidelity, with the previously added daisy and cream ribbon remaining clearly visible at turn~10.}

  \label{fig:supp-single-input-m2}
\end{figure}

\begin{figure}[p]
  \centering
    \vspace{-15mm}
    \makebox[\linewidth][c]{%
      \includegraphics[
        width=0.76\linewidth,
        keepaspectratio
      ]{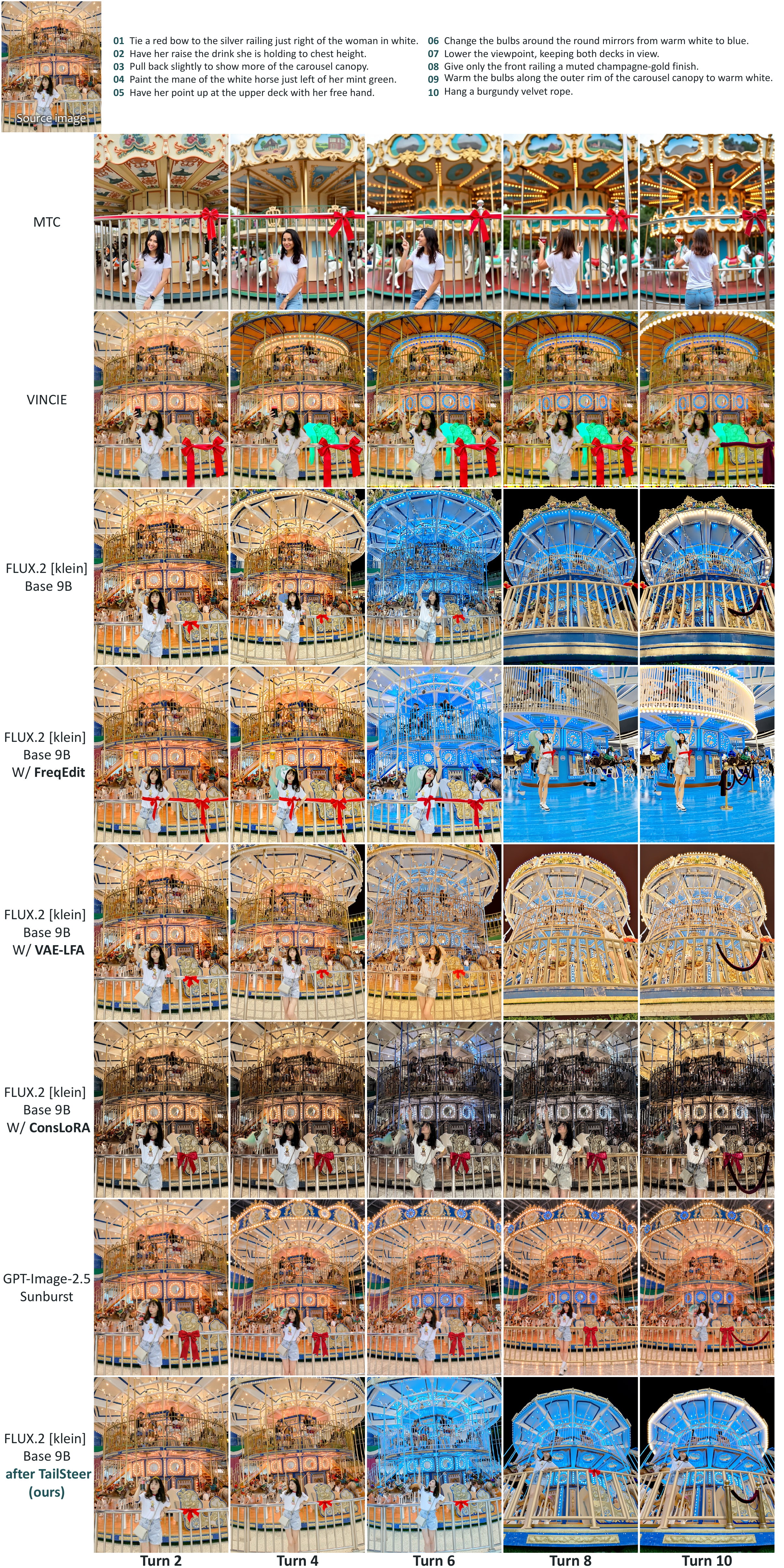}%
    }\par
    \vspace{-2mm}
    \caption{\textbf{Unlocking drift-resistant multi-turn editing after TailSteer: General edits.} Across ten challenging edits involving local modifications, pose changes, lighting, and camera movements, TailSteer retains the base model's editing capabilities while reducing cumulative drift. At turn~7, TailSteer successfully lowers the viewpoint while retaining the woman and both carousel decks, whereas other methods either fail to achieve the requested viewpoint or lose the subject altogether. By preserving the intricate canopy structure, TailSteer also successfully changes the outer ring of lights to warm white at turn~9 while retaining the blue interior lighting. Even at turn~10, the woman's facial appearance and clothing remain consistent, demonstrating strong preservation under multiple transformations.}

  \label{fig:supp-single-input-m3}
\end{figure}

We first compare FLUX.2 [klein] Base 9B before and after TailSteer with six baselines across eight ten-turn sequences, grouped into three types of editing tasks: (i) \emph{local edits}, which require precise modifications while preserving surrounding details (\cref{fig:supp-single-input-l1,fig:supp-single-input-l2,fig:supp-single-input-l3}); (ii) \emph{identity-preserving edits}, which test subject consistency across non-rigid pose changes and camera reframing (\cref{fig:supp-single-input-i2,fig:supp-single-input-i3}); and (iii) \emph{general edits}, which combine local and non-local transformations involving content, appearance, lighting, actions, and viewpoint (\cref{fig:supp-single-input-m1,fig:supp-single-input-m2,fig:supp-single-input-m3}). All methods follow the same sequence of instructions, using their own previous outputs as inputs. Across all three task types, TailSteer retains the base model's learned editing priors while substantially reducing cumulative drift, enabling successive edits with high visual fidelity even at turn~10.

\begin{figure}[p]
  \centering
    \vspace{-10mm}
    \makebox[\linewidth][c]{%
      \includegraphics[
        width=1.25\linewidth,
        keepaspectratio
      ]{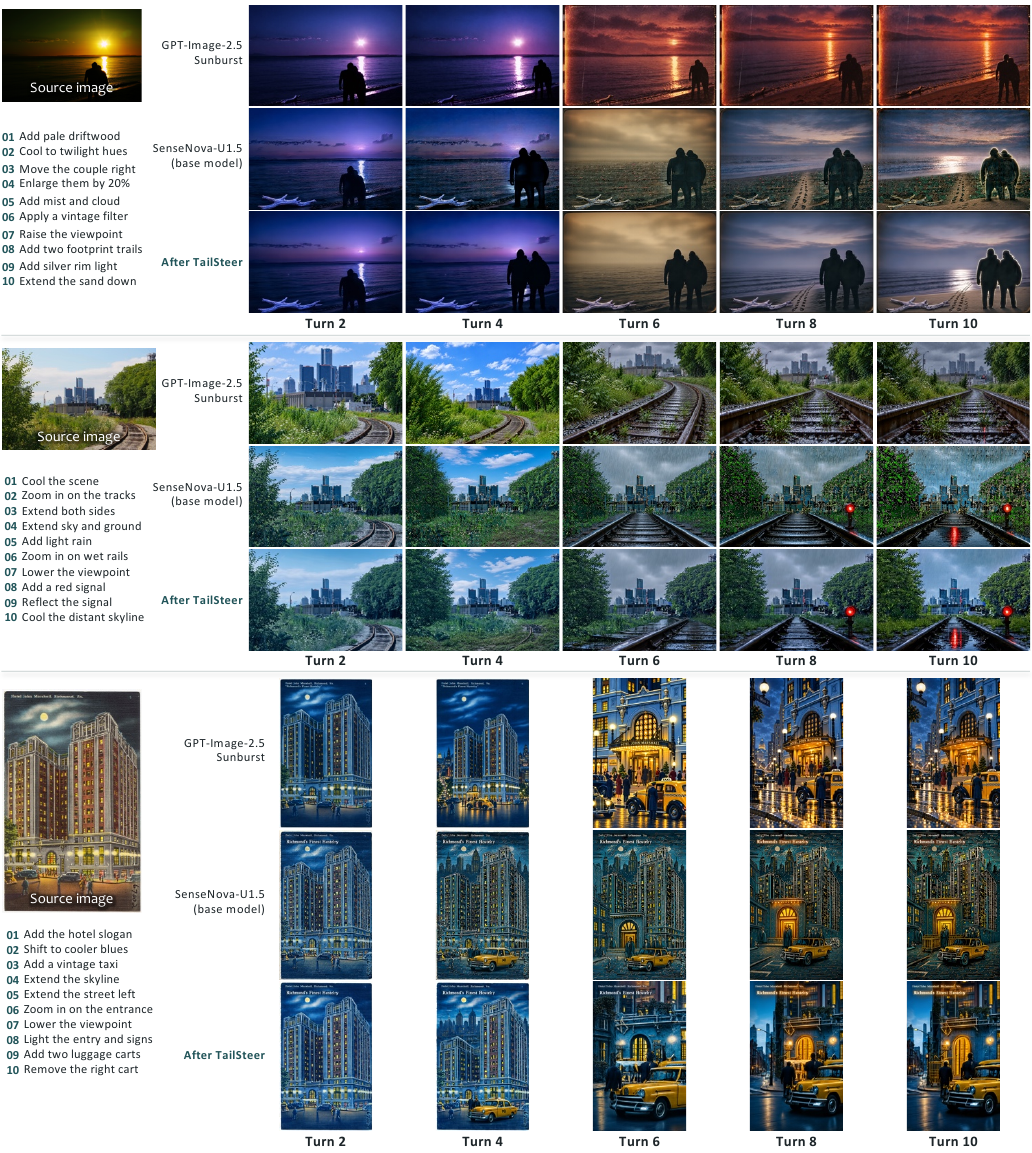}%
    }\par
    \vspace{-3mm}
    \caption{\textbf{Drift-resistant multi-turn editing on a VAE-free pixel-space model.} Ten-turn sequences compare GPT-Image-2.5 Sunburst, SenseNova-U1.5-8B-MoT, and TailSteer, with results shown every two turns. Despite successive changes in lighting, viewpoint, and scene content, TailSteer substantially reduces the texture degradation and structural collapse seen in the base model, preserving the beach scene, railway details, and hotel architecture. These examples demonstrate that TailSteer also improves long-term preservation in VAE-free models, achieving visual consistency comparable to Sunburst.}

  \label{fig:supp-single-input-sensenova}
\end{figure}

\begin{figure}[p]
  \centering
      \vspace{-10mm}
    \makebox[\linewidth][c]{%
      \includegraphics[
        width=1.25\linewidth,
        keepaspectratio
      ]{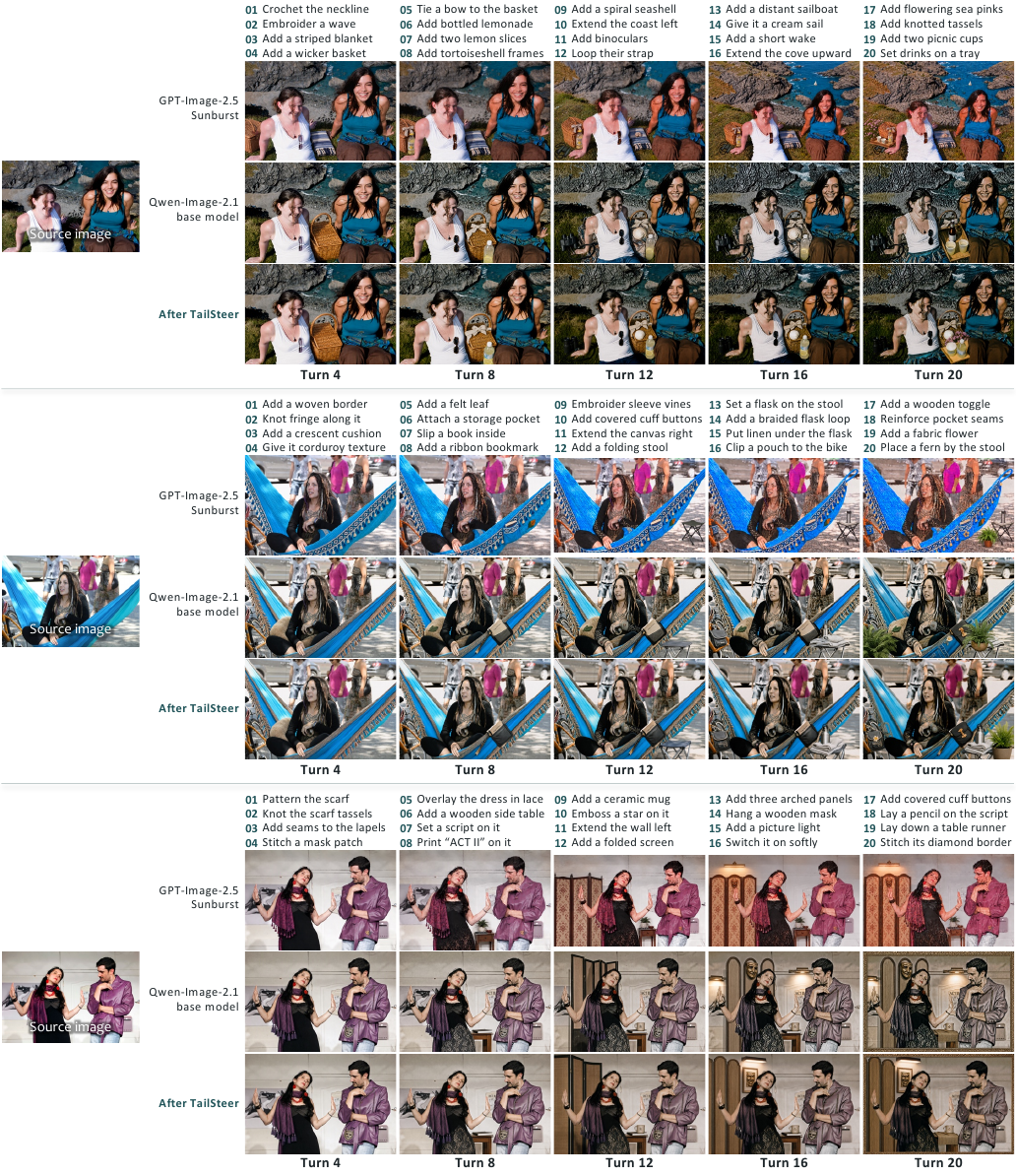}%
    }\par

    \caption{\textbf{Drift-resistant 20-turn editing with stronger identity preservation.} Across three diverse editing sequences, TailSteer ($k_s=30$) substantially reduces the cumulative drift of Qwen-Image-2.1 while retaining its learned editing capabilities. Even after 20 successive edits, facial identities, fine details, and surrounding scenes remain remarkably consistent with the sources, showing stronger preservation than both the base model and GPT-Image-2.5 Sunburst in these examples.}

  \label{fig:supp-single-input-qwen}
\end{figure}

We further compare SenseNova-U1.5-8B-MoT and Qwen-Image-2.1 before and after TailSteer with GPT-Image-2.5 Sunburst. In the SenseNova railway sequence, TailSteer preserves the photographic textures of the trees and skyline over ten turns (\cref{fig:supp-single-input-sensenova}). The Qwen examples extend to twenty turns, with TailSteer maintaining facial identity and fine details through changes to clothing, objects, and surroundings (\cref{fig:supp-single-input-qwen}). Across these scenarios, TailSteer substantially reduces cumulative drift, often matching or surpassing Sunburst in content preservation while retaining the base models' learned editing priors.

\subsection{Multi-Input Multi-Turn Editing}
\label{app:qualitative-multi-input}\label{app:multi-input-multi-turn-editing}

\begin{figure}[p]
  \centering
    \vspace{-10mm}
    \makebox[\linewidth][c]{%
      \includegraphics[
        width=1.25\linewidth,
        keepaspectratio
      ]{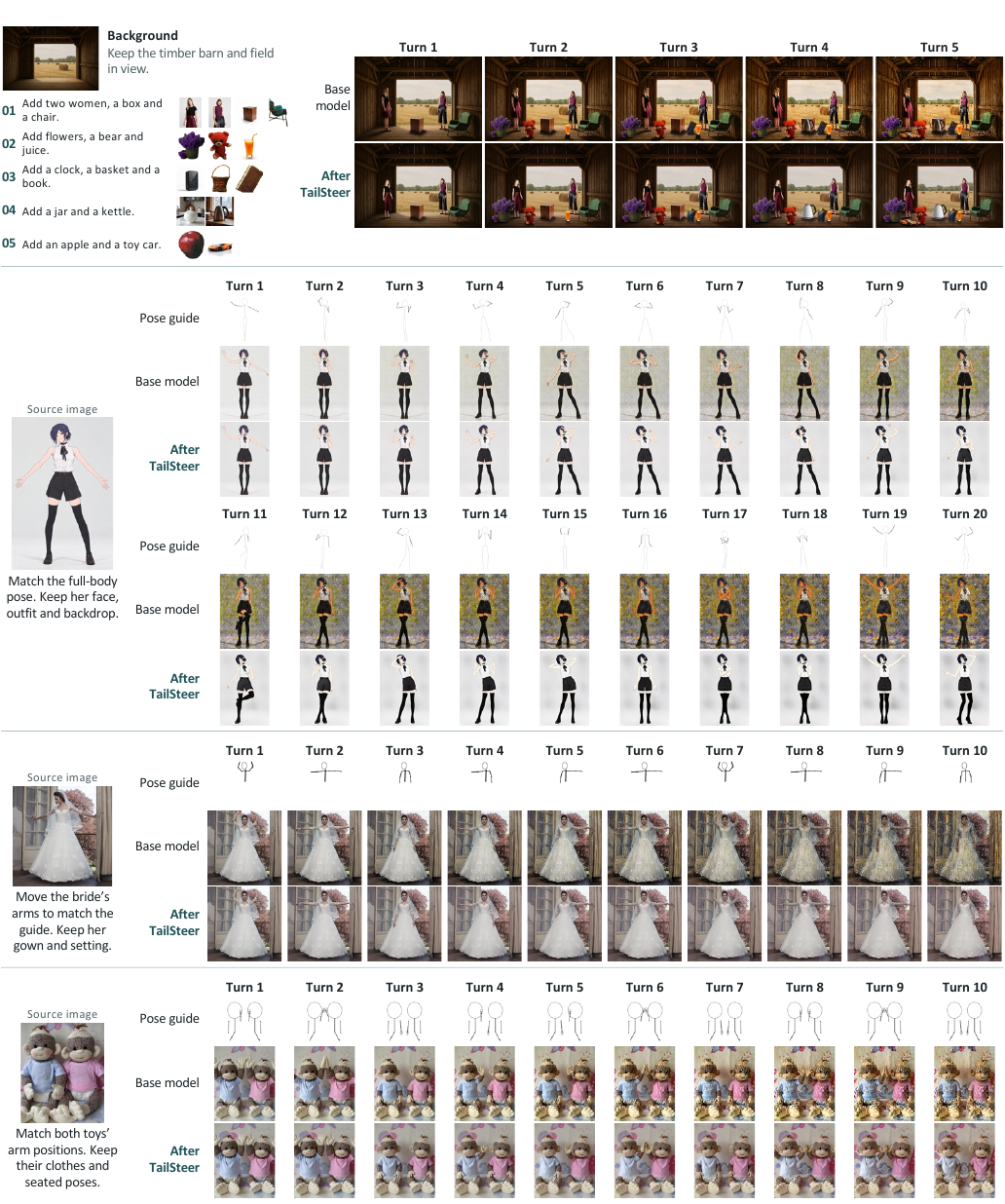}%
    }\par

    \caption{\textbf{Multi-input multi-turn editing with minimal cumulative drift.} FLUX.2 [klein] Base 9B before and after TailSteer is evaluated on five-turn multi-object compositing, twenty-turn dancer pose editing, and ten-turn skeleton-guided editing of a bride and two plush toys. TailSteer follows successive visual references while preserving previously added objects, subject identities, clothing, and background details. Compared with the base model, it substantially reduces the accumulation of unwanted textures and appearance distortions, particularly in the pose-guided sequences.}

  \label{fig:supp-multi-input-klein}
\end{figure}

\begin{figure}[p]
  \centering
    \vspace{-10mm}
    \makebox[\linewidth][c]{%
      \includegraphics[
        width=1.25\linewidth,
        keepaspectratio
      ]{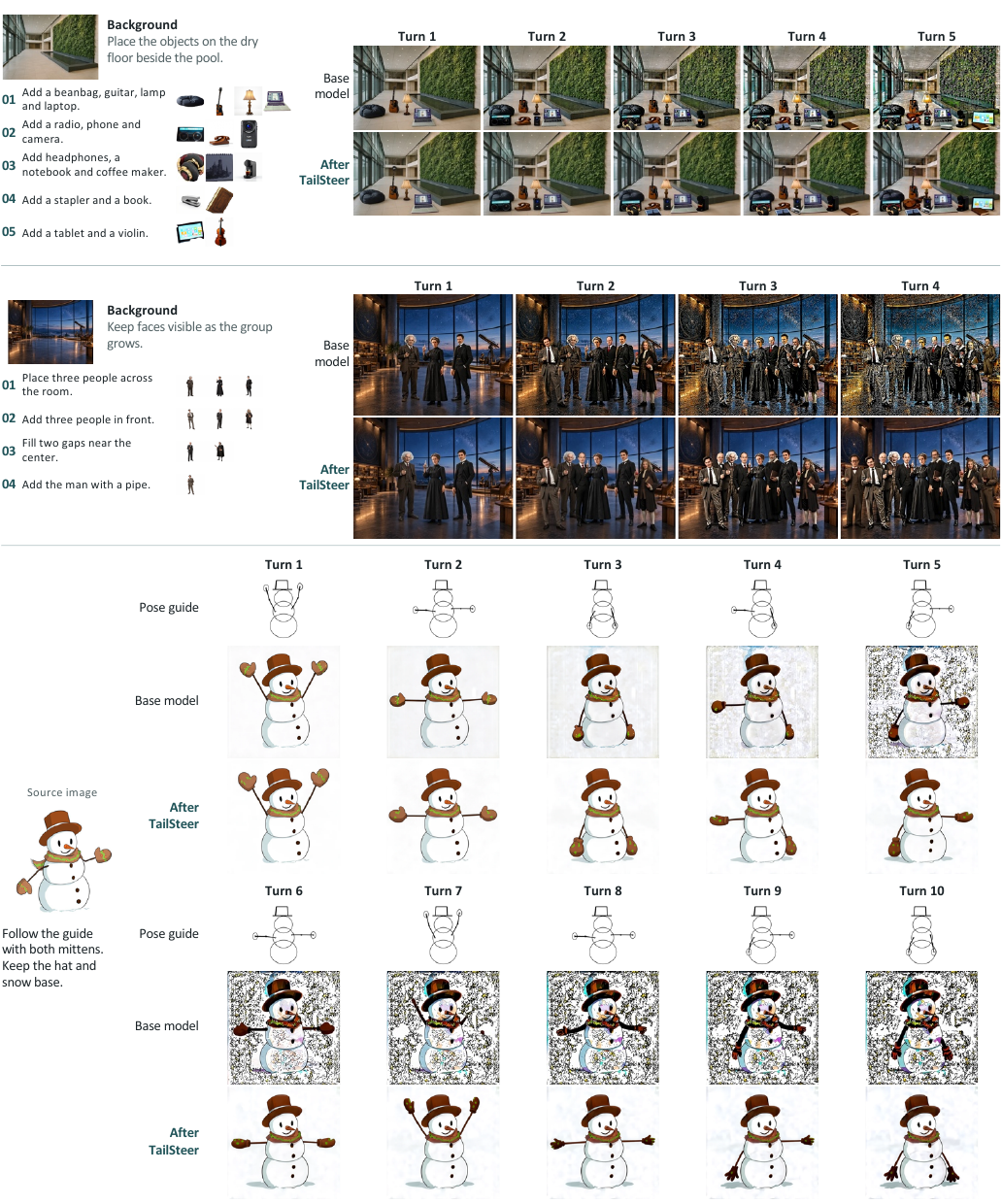}%
    }\par
    \caption{\textbf{Drift-resistant multi-input editing after TailSteer.} SenseNova-U1.5-8B-MoT is evaluated on five-turn object compositing, four-turn multi-person compositing, and ten-turn skeleton-guided pose editing. TailSteer faithfully integrates new visual references while preserving earlier additions, facial details, and background consistency. Compared with the base model, it substantially reduces accumulated texture distortions in the composited scenes and keeps the snowman's background clean throughout successive pose changes. These results suggest that trajectory-tail refinement also strengthens multi-input editing without compromising the base model's learned capabilities.}

  \label{fig:supp-multi-input-sensenova}
\end{figure}

\begin{figure}[p]
  \centering
    \vspace{-10mm}
    \makebox[\linewidth][c]{%
      \includegraphics[
        width=1.25\linewidth,
        keepaspectratio
      ]{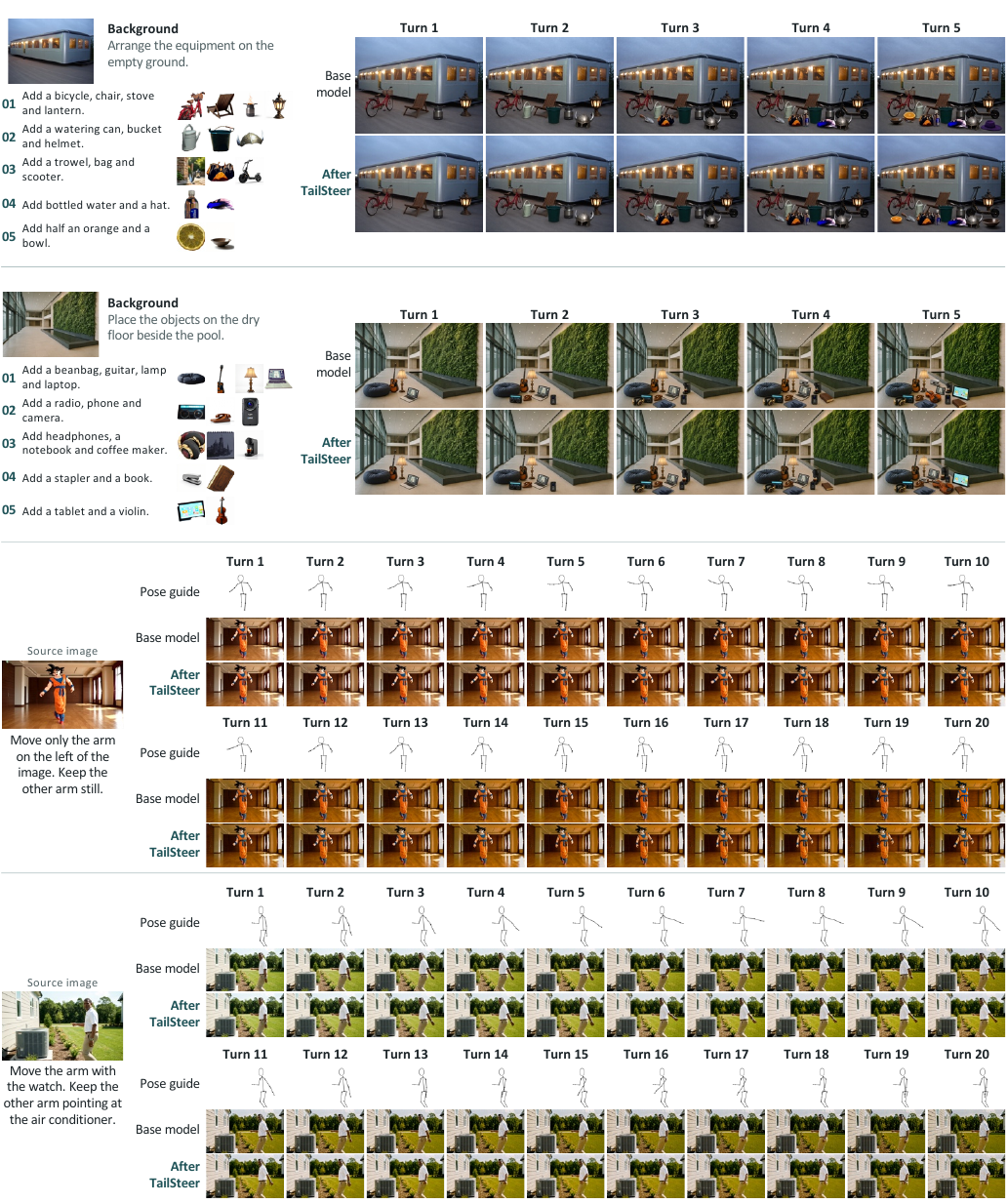}%
    }\par
    \caption{\textbf{Drift-resistant multi-input editing with Qwen-Image-2.1.} Five-turn object compositing with fourteen visual references, where TailSteer integrates new objects while preserving earlier additions and surrounding scene details. Twenty-turn skeleton-guided editing, where only the specified arm moves while the other remains unchanged. Across these settings, TailSteer retains the base model's multi-input editing capabilities while reducing cumulative drift in subject appearance, clothing, and background details.}

  \label{fig:supp-multi-input-qwen}
\end{figure}

Multi-input editing further tests whether models can incorporate new visual references while preserving content established in earlier turns. In \cref{fig:supp-multi-input-klein,fig:supp-multi-input-sensenova,fig:supp-multi-input-qwen}, each method receives the same instructions, visual references, and seeds, while editing its own previous output. For multi-object compositing, TailSteer retains previously added elements as new ones are introduced. For example, in the atrium sequence, the guitar and lamp added at turn~1 remain present as the tablet and violin are introduced at turn~5.

Pose-guided editing presents a related challenge: models must follow changing skeleton references while preserving the subject's appearance and surrounding scene. In the Klein dance and SenseNova snowman sequences, the base models gradually introduce unwanted background textures, whereas TailSteer keeps the backgrounds closer to the source while following the pose changes. The Qwen examples further test precise pose control by moving one arm while preserving the other. Together, these examples illustrate how TailSteer maintains the base models' ability to follow diverse visual guidance while reducing drift across successive edits.

\subsection{Simulated Creative Workflows}
\label{app:simulated-creative-workflows}

\begin{figure}[p]
  \centering
    \vspace{-10mm}
    \makebox[\linewidth][c]{%
      \includegraphics[
        width=1.25\linewidth,
        keepaspectratio
      ]{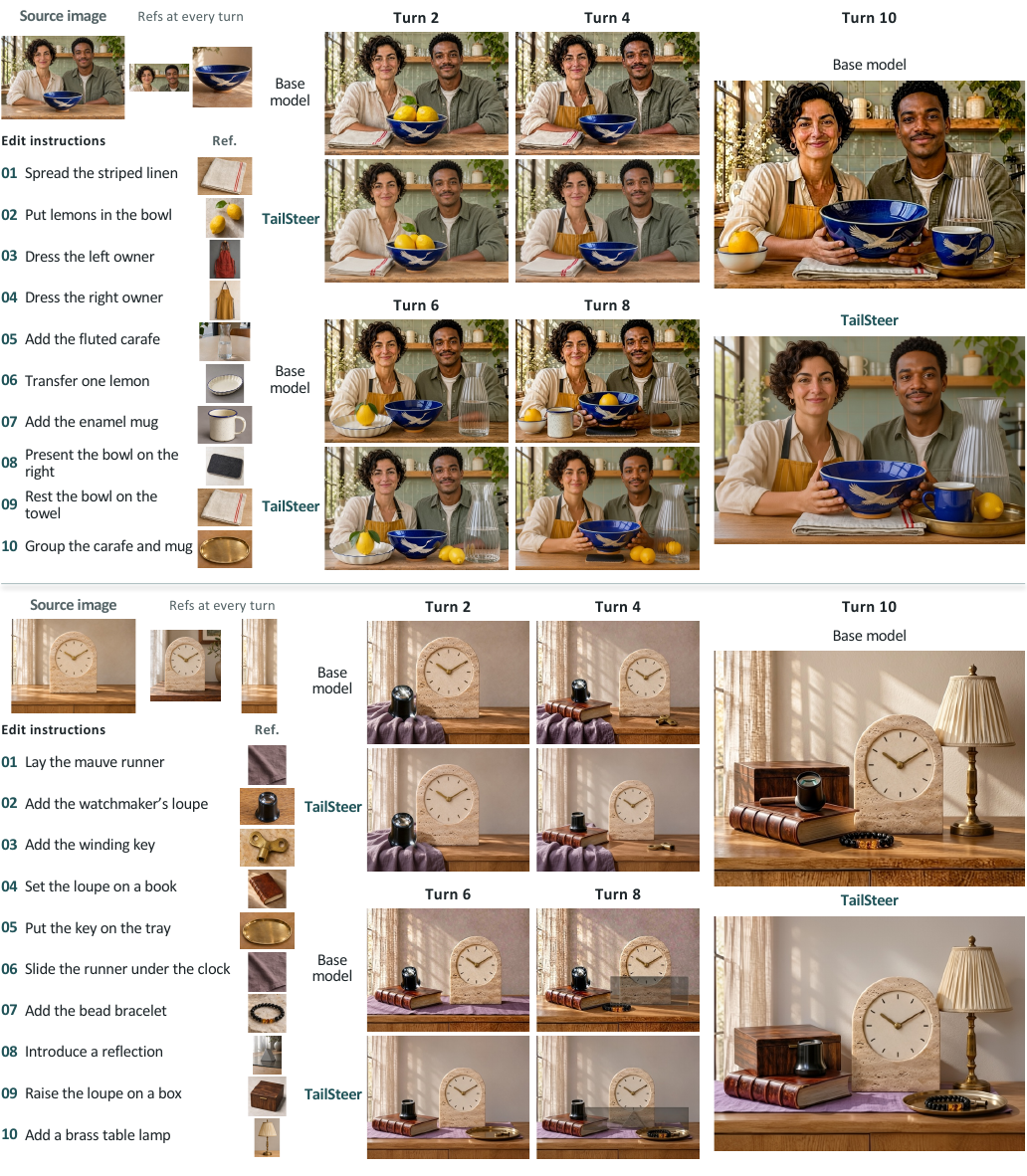}%
    }\par
    \vspace{-3mm}
    \caption{\textbf{Drift-resistant creative workflows after TailSteer.} We compare FLUX.2 [klein] Base 9B before and after TailSteer, using two fixed references and a turn-specific reference for each edit. In the ceramics shoot (top), TailSteer better preserves the owners' facial features, natural skin tones, and the bowl's distinctive design as the styling evolves, avoiding the exaggerated textures seen in the base model. In the clock shoot (bottom), TailSteer retains the original lighting and material details, along with the mauve runner and brass tray that disappear from the base model's final output. The enlarged turn-10 results highlight TailSteer's ability to maintain visual fidelity and earlier creative decisions throughout iterative creation.}
  \label{fig:supp-creative-workflows-klein}
\end{figure}

\begin{figure}[p]
  \centering
      \vspace{-10mm}
    \makebox[\linewidth][c]{%
      \includegraphics[
        width=1.25\linewidth,
        keepaspectratio
      ]{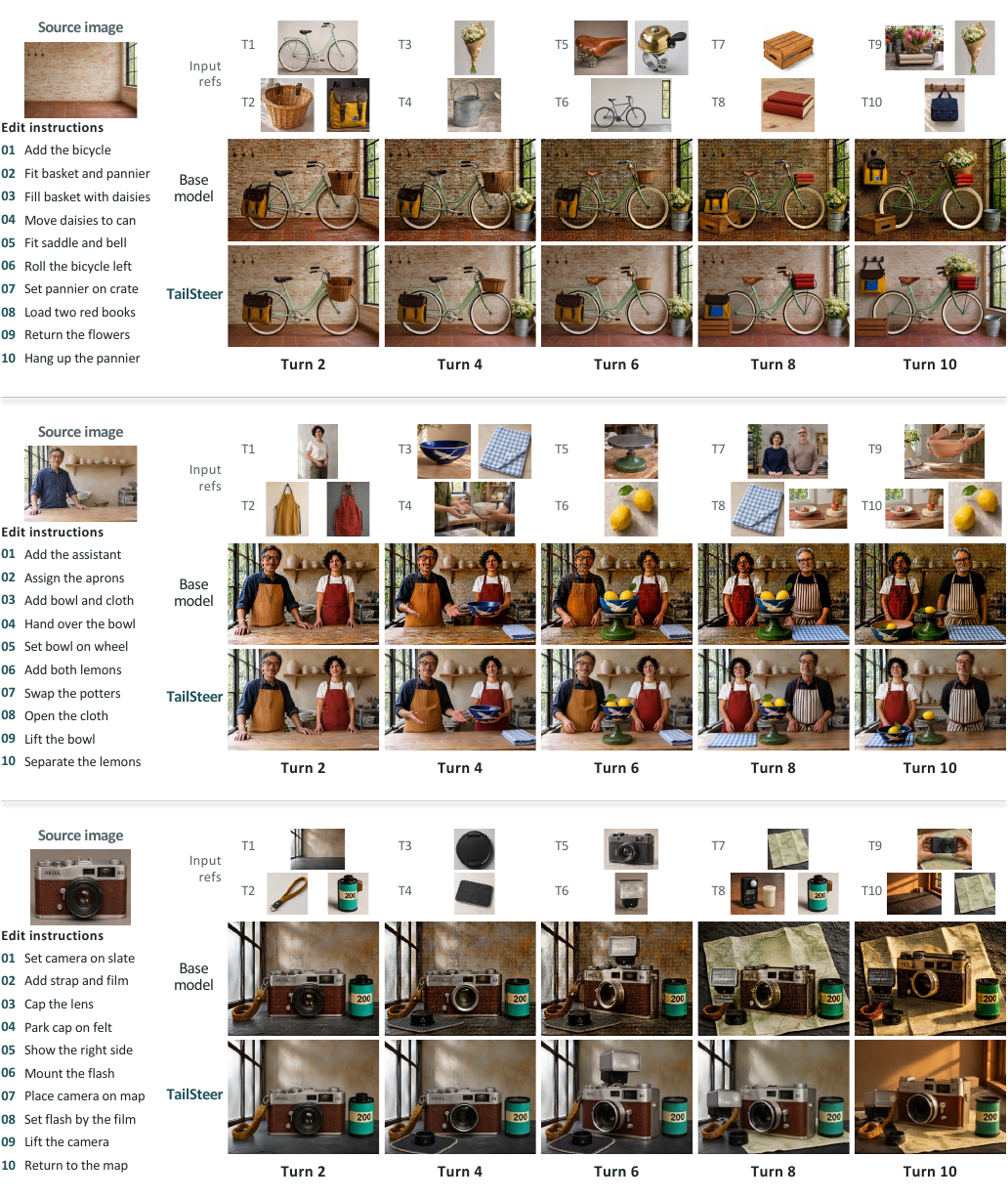}%
    }\par
    \vspace{-3mm}
    \caption{\textbf{Drift-resistant creative workflows after TailSteer.} We compare SenseNova-U1.5-8B-MoT before and after TailSteer across three ten-turn creative workflows, with visual references shown above each pair of edits. In the bicycle and pottery shoots, TailSteer preserves the original wall and floor textures, natural facial features, and scene lighting while incorporating new props and rearranging subjects. In the camera shoot, it retains the metallic finish, leather texture, and product details through successive styling changes. In contrast, the base model accumulates pronounced texture artifacts and color distortions, highlighting TailSteer's advantage in maintaining photographic fidelity throughout iterative creation.}
  \label{fig:supp-creative-workflows-sensenova}
\end{figure}

\begin{figure}[p]
  \centering
    \vspace{-10mm}
    \makebox[\linewidth][c]{%
      \includegraphics[
        width=1.25\linewidth,
        keepaspectratio
      ]{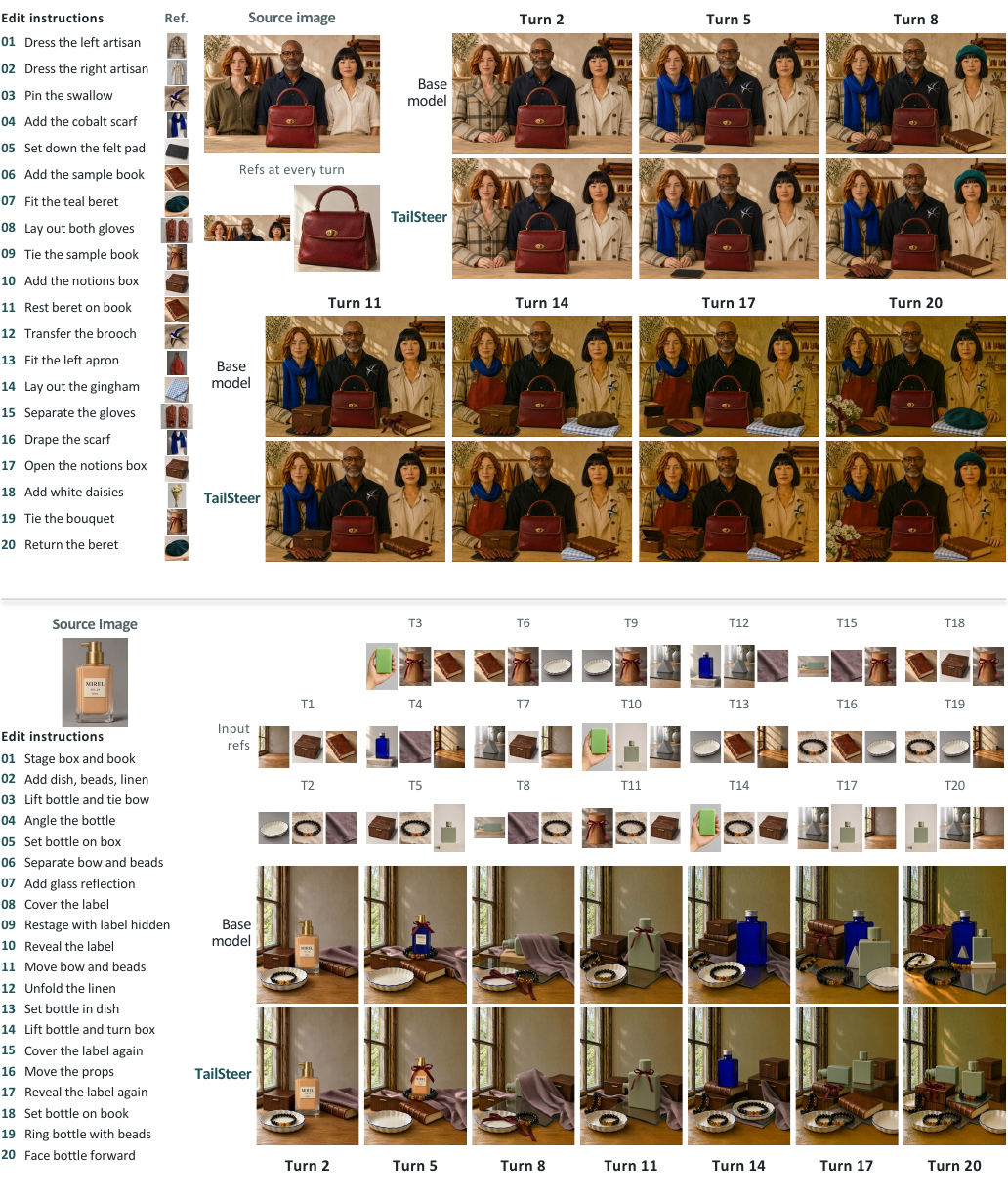}%
    }\par
    \vspace{-3mm}
    \caption{\textbf{Drift-resistant creative workflows after TailSteer.} We compare Qwen-Image-2.1 before and after TailSteer across two twenty-turn creative workflows. In the artisan shoot (top), TailSteer better preserves facial features, clothing details, and the handbag's distinctive appearance as the styling evolves. At turn~20, it correctly returns the beret to the woman, whereas the base model leaves it on the table. In the product shoot (bottom), TailSteer maintains a more coherent composition through repeated changes in object placement and visibility, avoiding the unintended bottle duplication seen in the base model's final output. These examples highlight TailSteer's ability to maintain visual consistency while accommodating evolving creative directions over extended editing sequences.}
  \label{fig:supp-creative-workflows-qwen}
\end{figure}

Building on the multi-input editing experiments, we explore how TailSteer supports iterative creation in more realistic settings. We script sevenphoto shoots featuring recurring people and products (\cref{fig:supp-creative-workflows-klein,fig:supp-creative-workflows-sensenova,fig:supp-creative-workflows-qwen}). At each turn, an art director's brief specifies changes to props or styling, sometimes revisiting earlier decisions, such as returning a beret to its owner or uncovering a label. These workflows require models to accommodate evolving creative directions while maintaining visual consistency across turns.

For each shoot, the base model and the refined model start from the same photograph and edit their own previous outputs, using identical prompts, seeds, and different references images at every turn. Klein and Qwen receive three external references per turn, while SenseNova receives one or two. 

\section{Limitations and Future Directions}
\label{app:limitations}

\subsection{Dependence on base model capabilities.}
\begin{figure}[p]
  \centering
      \vspace{-10mm}
    \makebox[\linewidth][c]{%
      \includegraphics[
        width=1.25\linewidth,
        keepaspectratio
      ]{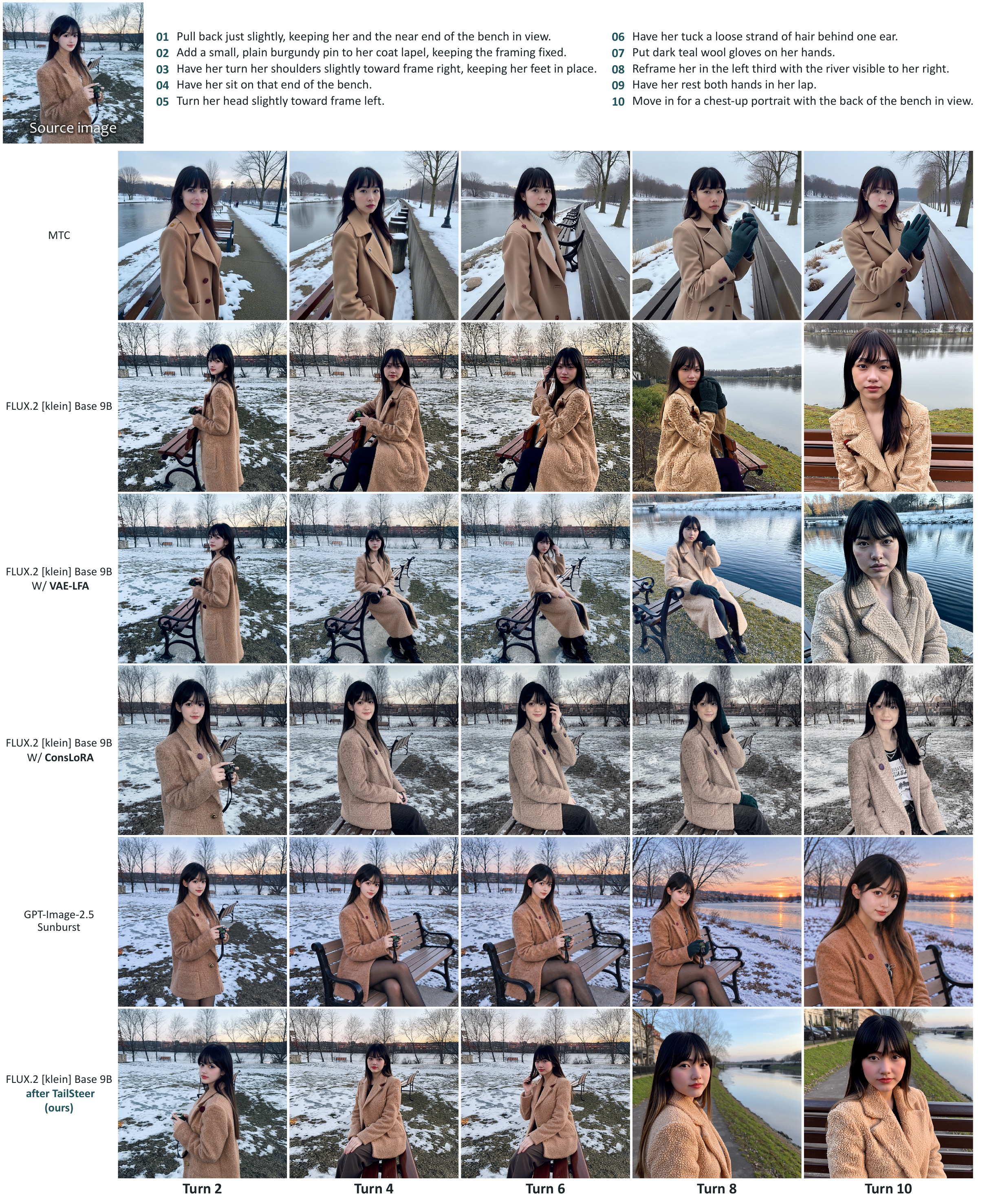}%
    }\par
    \caption{\textbf{Multi-turn editing inherits the limitations of the base model.} Across ten edits involving appearance, pose, and viewpoint changes, TailSteer better preserves the woman's makeup, particularly her lipstick color, as well as her bangs and coat texture compared with the base model and other baselines. However, GPT-Image-2.5 Sunburst more faithfully retains her intrinsic facial features, including the shapes of her eyes and nose. Despite improved appearance consistency, the final TailSteer portrait still exhibits noticeable identity drift, giving the impression of a different person. This example illustrates that reducing cumulative drift does not necessarily overcome limitations in the base model's learned identity-preserving capabilities.}
    \label{fig:supp-identity-limitations}

\end{figure}

TailSteer aims to unlock the multi-turn editing potential of a base model while retaining its learned generative and editing priors. Its effectiveness therefore depends in part on the capabilities already present in the base model, including preservation-related abilities such as identity consistency. As illustrated in \cref{fig:supp-identity-limitations}, TailSteer improves the facial consistency of FLUX.2 [klein] across successive edits, but still falls short of GPT-Image-2.5 Sunburst in this example, where base model itself struggles to preserve facial identity.

This limitation reflects a deliberate choice in our training philosophy. Although extensive multi-turn post-training may strengthen specific capabilities, training toward a narrow set of editing tasks risks compromising the general-purpose abilities acquired through pre-training, mid-training, and subsequent post-training stages, including RL and OPD. We believe that capabilities missing from the base model should primarily be developed during these earlier stages. Multi-turn robustness should then serve as a final training stage that makes the learned capabilities more persistent across successive edits. TailSteer follows this principle, focusing on reducing cumulative drift while retaining the versatility of the base model.

\subsection{Towards truly persistent editing.}

While TailSteer substantially reduces cumulative drift across different base models and compares favorably with existing methods, residual errors can still accumulate over successive edits. As shown in \cref{fig_8_preserve_plot}, TailSteer substantially slows the decline in PSNR and SSIM and the increase in LPIPS, but these curves have yet to stabilize. Ultimately, truly persistent editing should maintain high visual fidelity regardless of the number of editing turns, with preservation metrics approaching stable values.

One direction toward this goal is to improve representation fidelity between editing turns. Repeated VAE reconstruction introduces additional errors beyond those arising from the denoising trajectory itself. Our latent-feedback experiments demonstrate that addressing this source of degradation can substantially complement TailSteer's trajectory-tail refinement. Further improvements in VAE fidelity or latent reuse could therefore bring existing base models closer to persistent editing.

Another promising direction is to improve the long-term stability of trajectory-tail corrections. Although TailSteer substantially narrows the trajectory-level train--test gap, even small residual errors can grow exponentially if repeatedly amplified across 100 editing turns. An interesting extension is to learn an approximately idempotent preservation correction at the trajectory tail, such that repeated refinement of already preserved content introduces no further unintended changes. This could be explored through an iterative self-training process, in which the model learns from its own multi-turn editing trajectories and progressively refines its tail behavior. Such a process may bring us closer to persistent editing while retaining the base model's learned generative and editing priors.

\clearpage

\end{document}